\documentclass{article}

\makeatletter
\providecommand{\@trackname}{}
\makeatother
\usepackage[final]{neurips_2026}

\usepackage{amsmath}
\usepackage{placeins}
\usepackage{caption}
\usepackage{amsthm} 

\usepackage{algorithm}
\usepackage{subcaption}
\usepackage{algpseudocode}
\usepackage{xcolor}
\usepackage{float}

\theoremstyle{plain}
\newtheorem{theorem}{Theorem}[section]
\newtheorem{proposition}[theorem]{Proposition}

\theoremstyle{definition}
\newtheorem{definition}[theorem]{Definition}
\newtheorem{assumption}[theorem]{Assumption}

\theoremstyle{remark}

\usepackage[utf8]{inputenc} 
\usepackage[T1]{fontenc}    
\usepackage{hyperref}       
\usepackage{url}            
\usepackage{booktabs}       
\usepackage{amsfonts}       
\usepackage{nicefrac}       
\usepackage{microtype}      
\usepackage{xcolor}         
\usepackage{tcolorbox}
\usepackage{enumitem}
\usepackage{adjustbox}
\usepackage{tikz}
\usepackage{multirow}
\usepackage{float}
\usepackage{tikz}
\usetikzlibrary{calc}
\usetikzlibrary{matrix,arrows.meta,positioning,fit,backgrounds}
\usetikzlibrary{matrix,arrows.meta,positioning}

\title{Align \& Invert: Solving Inverse Problems with Diffusion and Flow-based Models via Representation Alignment}

\author{%
  Loukas Sfountouris \\
  Department of Computer Science \\
  University of Warwick \\
  Coventry, England \\
  \texttt{loukas.sfountouris@warwick.ac.uk}
  \And
  Giannis Daras \\
  Massachusetts Institute of Technology \\
  Massachusetts, USA \\
  \texttt{gdaras@mit.edu}
  \And
  Paris Giampouras \\
  Department of Computer Science \\
  University of Warwick \\
  Coventry, England \\
  \texttt{paris.giampouras@warwick.ac.uk}
}

\begin{document}

\maketitle

\begin{abstract}
   Enforcing alignment between the internal representations of diffusion or flow-based generative models and those of pretrained self-supervised encoders has recently been shown to provide a powerful inductive bias, improving both convergence and sample quality. In this work, we extend this idea to inverse problems, where pretrained generative models are employed as {\it priors}. We propose applying {\it representation alignment} ($\textsc{Repa}$) between diffusion or flow-based models and a DINOv2 visual encoder, to guide the reconstruction process at inference time. Although ground-truth signals are unavailable in inverse problems, we show that aligning model representations of approximate target features can substantially enhance reconstruction quality and perceptual realism.
We provide theoretical results showing (a) that  $\textsc{Repa}$ regularization can be viewed as a variational approach for minimizing a divergence measure in the DINOv2 space, and (b) how under certain regularity assumptions $\textsc{Repa}$ updates steer the latent diffusion states toward those of the clean image. 
We integrate \textsc{Repa}  into multiple state-of-the-art inverse problem solvers, and provide extensive experiments confirming that our method consistently improves reconstruction quality, while also reducing the number of discretization steps required to reach the same performance level as the underlying solver. 
Our code is publicly available at \url{https://github.com/Sfountouris/Align_And_Invert}.
\end{abstract}

\section{Introduction}

Pretrained diffusion and flow-based models, \citep{sohl2015deep, ho2020denoising,lipman2023flow}, have recently  been  at the heart of methods focusing on addressing various types of inverse problems. The crux of these approaches is to perform diffusion sampling,
incorporating measurement information during the reverse process, with the goal to reconstruct the clean image at the final step, \citep{patel2024steering, chung2023diffusion, thaker2025frequency}. Undoubtedly, diffusion and flow-based models have pushed the boundaries in addressing a wealth of inverse problems, \citep{daras2024survey}. However, they still struggle to produce highly detailed images particularly in cases of severe degradation and in challenging scenes, such as natural images with rich textures and complex backgrounds. These issues are further exacerbated when using latent diffusion models \citep{rombach2022high}, where the inclusion of encoder-decoder architectures introduces additional challenges during the reverse process due to nonlinearities. To mitigate these limitations, further inductive biases are needed during inference, \citep{rout2023solving, raphaeli2025silo}.

\begin{figure}[t]
\centering

\definecolor{panelbg}{RGB}{222,203,206}
\definecolor{sitgreen}{RGB}{186,216,186}


\newcommand{\squareimg}[2]{%
\begin{tikzpicture}[baseline]
  \clip (0,0) rectangle (#1,#1);
  \node[anchor=south west, inner sep=0pt] {\includegraphics[width=#1,height=#1]{#2}};
\end{tikzpicture}
}
\adjustbox{width=\textwidth}{%
\begin{tikzpicture}[
    every node/.style={transform shape},
    font=\ttfamily,
    >=stealth,
    imgbox/.style={
        draw,
        rounded corners=2pt,
        fill=white,
        minimum width=1.7cm,
        minimum height=1.5cm,
        align=center,
    },
    objective/.style={
        draw,
        rounded corners=5pt,
        minimum width=1.3cm,
        minimum height=1.1cm,
        align=center,
    },
    model/.style={
        draw,
        rounded corners=5pt,
        minimum width=2.2cm,
        minimum height=1.2cm,
        align=center,
        fill=gray!20,
    },
    transformer/.style={
        draw,
        rounded corners=4pt,
        minimum width=2.2cm,
        minimum height=0.85cm,
        align=center,
        fill=sitgreen,
    },
    panel/.style={
        draw=none,
        rounded corners=18pt,
        fill=panelbg,
        inner sep=0.7cm
    },
    mlp/.style={
        draw,
        rounded corners=4pt,
        minimum width=1cm,
        minimum height=0.6cm,
        fill=yellow!40,
        align=center,
    }
]

\begin{scope}[xshift = 2cm, yshift=-8mm, scale = 1.1, every node/.style={transform shape}]
\matrix (rightpanel) [
    matrix of nodes,
    nodes in empty cells,
    nodes={inner sep=0pt},
    column sep=0.30cm,
    row sep=0.22cm,
    anchor=north west
]{
    {\textbf{Measurement}} &
    {\textbf{Without $\textsc{REPA}$}} &
    {\textbf{With $\textsc{REPA}$}} &
    {\textbf{Ground Truth}} \\
    \node (repa1)   {\squareimg{3.3cm}{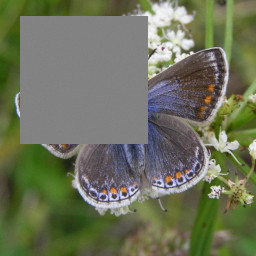}}; &
    \node (repa2)   {\squareimg{3.3cm}{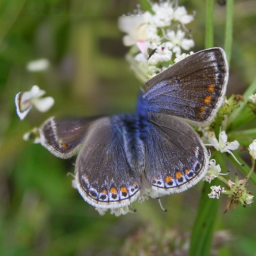}}; &
    \node (repa3)   {\squareimg{3.3cm}{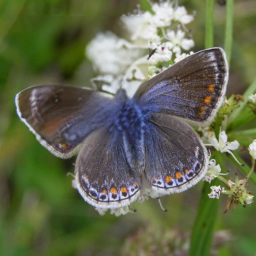}}; &
    \node (repa4)   {\squareimg{3.3cm}{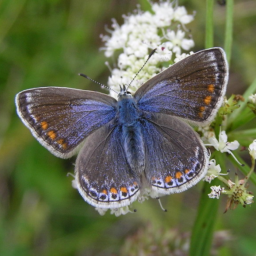}}; \\
    \node (worepa1) {\squareimg{3.3cm}{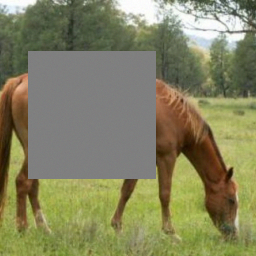}}; &
    \node (worepa2) {\squareimg{3.3cm}{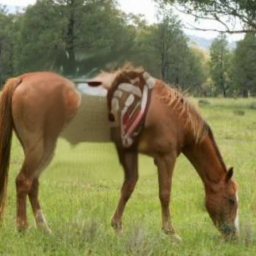}}; &
    \node (worepa3) {\squareimg{3.3cm}{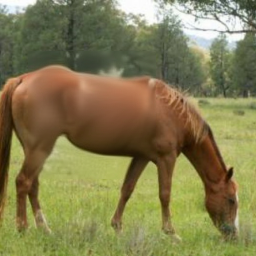}}; &
    \node (worepa4) {\squareimg{3.3cm}{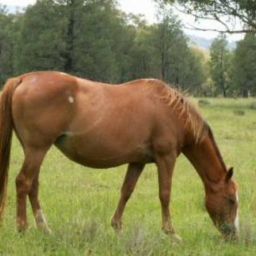}}; \\
};
\end{scope}

\begin{scope}[on background layer]
  \node[panel, draw=none,
        inner xsep=9pt,
        inner ysep=20pt,
        fit={(repa1.north west) (worepa4.south east)}] (bigpanel) {};
\end{scope}

\begin{scope}[xshift = 10.5cm, every node/.style={transform shape}]
{\large


\coordinate (alignanchor) at ($(bigpanel.east) + (3.6cm, 0)$);

\node[objective, above=3cm of alignanchor] (alignobj)
  {Representation Alignment\\Objective};

\node[model, draw=none, below=-0.7cm of alignanchor, xshift=-0.8cm] (dino)
  {DINOv2 \\ Encoder};

\node[inner sep=0pt, below=0.8cm of dino] (horse_kk)
  {\includegraphics[width=1.7cm]{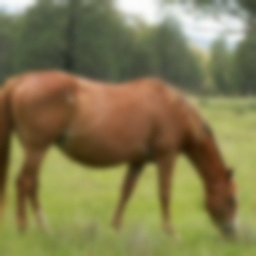}};

\node[below=2pt of horse_kk, align=center]
  {Approximate\\ reconstructed image};

\draw[->] (horse_kk.north) -- (dino.south);
\draw[->] (dino.north) -- ([xshift=-7.9mm]alignobj.south);
\node[objective, right=3.9cm of dino, yshift=3.48cm] (measobj)
  {Measurement Matching \\ Objective};

\node[transformer, draw=none, below=1.2cm of measobj, xshift=0.1cm] (t1)
  {SiT Transformer Block};
\node[transformer, draw=none, below=0.04cm of t1] (t2)
  {SiT Transformer Block};
\node[transformer, draw=none, below=0.04cm of t2] (t3)
  {SiT Transformer Block};
\node[transformer, draw=none, below=0.04cm of t3] (t4)
  {SiT Transformer Block};
\node[transformer, draw=none, below=0.04cm of t4] (t5)
  {SiT Transformer Block};
\node[transformer, draw=none, below=0.04cm of t5] (t6)
  {SiT Transformer Block};

\draw[->] (t1.north) -- (t1.north |- measobj.south);
\coordinate (p3) at ($(t4.west) + (0,0.2)$);
\node[mlp, draw=none, below right=-0.3cm and 0.5cm of dino] (mlp) {MLP};
\draw[<-] (mlp.south) |- (p3);


\draw[->] (mlp.north) -- ($(mlp.north |- alignobj.south)$);
\node[font=\bfseries\large]
  at ($(alignobj.east)!0.5!(measobj.west)$) {+};

}
\end{scope}

\end{tikzpicture}
}
\captionof{figure}{
Overview of our proposed framework. 
Left: Box inpainting results, where adding 
$\textsc{Repa}$ improves perceptual quality.
Right: Alignment between diffusion features and DINOv2 embeddings.
}
\label{fig:overview}

\end{figure}

Recent studies have shown that diffusion models learn semantic features in their hidden states,  \citep{li2023your}. The more expressive these features are the better the diffusion model performs on the generative task, \citep{xiang2023}. Building
on this insight, the seminal work of \cite{yu2024repa} introduced a regularizer
that aligns the internal representations of the diffusion model with those of a
pretrained visual encoder, \citep{oquab2023dinov2}. This framework, termed {\it representation alignment} (\textsc{Repa}), was shown to act as a strong semantic constraint leading to higher-fidelity generations and significantly faster convergence. 

The success of $\textsc{Repa}$  has attracted considerable interest in this direction, \citep{wang2025repa, tian2025u, yao2025reconstruction, leng2025repa, wang2025learning}. In particular, \citep{wang2025repa} observed that while $\textsc{Repa}$ enhances early training dynamics, its benefits diminish over time and can even degrade performance in later iterations. To address this issue, they introduced an early-stopping strategy which mitigated the late-stage degradation and
further improved overall convergence speed.  Other works have explored
applying alignment losses directly to VAE latent spaces
\citep{yao2025reconstruction, xu2025exploring}, finding that diffusion models
trained on these regularized spaces converge faster. Despite these advances,
most efforts focus on improving training efficiency, whereas relatively little
attention has been paid to leveraging representation alignment at inference time,
particularly for solving inverse problems. This motivates our central question: 
\begin{tcolorbox}[colback=gray!10, colframe=gray!50, boxrule=0.5pt, arc=2mm, left=4pt, right=4pt, top=1pt, bottom=1pt,before skip=3pt,
  after skip=5pt]
{\it Can we apply representation alignment to benefit existing algorithms for solving inverse problems using pretrained diffusion and flow-based models?}
\end{tcolorbox}
%
\textbf{Contributions.}
Our main contributions are as follows:
\begin{itemize}[topsep=2pt, itemsep=2pt, parsep=0pt, partopsep=0pt, leftmargin=*]
    \item We introduce a general framework for solving inverse problems with diffusion and flow-based models by enforcing alignment between internal diffusion representations and DINOv2 features (Fig. \ref{fig:overview}). Despite the absence of ground truth, alignment with proxy reconstructions remains effective thanks to the robustness/invariances of DINOv2 representations.
    \item We provide theoretical results that characterize the effect of the
    $\textsc{Repa}$ regularizer: (a) its connection to a divergence measure defined in the
    DINOv2 feature space, and (b) under suitable regularity assumptions, its ability to steer
    latent diffusion states toward those associated with the clean image.
    \item We demonstrate the versatility of our approach by integrating representational alignment into different existing state-of-the-art inverse problem solvers.
    \item Through extensive experiments on super-resolution, Gaussian deblurring, motion deblurring, and box inpainting, we show consistent improvements over prior methods across tasks and metrics, and further demonstrate that applying \textsc{Repa} to existing solvers allows them to match their original performance using significantly fewer discretization steps.
\end{itemize}

\section{Related Work}
\subsection{Diffusion  \& Flow-based Models}

Generative models based on denoising are a class of deep generative models that aim to learn a data distribution \( p_0(x) \) by training on progressively 
corrupted versions of real samples \citep{sohl2015deep, ho2020denoising, song2021denoising, 
liu2023flow}. These models define a) a forward (noising) process, which gradually interpolates between the data distribution and a reference distribution $p_1(x)$ (typically a Gaussian), and b) a reverse (denoising) process, which is trained to invert this corruption and iteratively reconstruct data from noise. Following the formulation of stochastic interpolants \citep{albergo2023stochastic}, the time marginals of this process can be expressed as
\begin{equation}
    x_t = \alpha_t x^* + \sigma_t \varepsilon, \qquad 
    x^* \sim p_0(x), \; \varepsilon \sim \mathcal{N}(0, I),
    \label{eq:time_dependent_process}
\end{equation}
where \( \alpha_t \) and \( \sigma_t \) correspond to decreasing and increasing functions of \( t \), respectively, with boundary conditions \( \alpha_0 = \sigma_1 = 1 \) and \( \alpha_1 = \sigma_0 = 0 \). Throughout, we use the notation \(\dot{\alpha}_t := \tfrac{\mathrm{d}\alpha_t}{\mathrm{d}t}\) and \(\dot{\sigma}_t := \tfrac{\mathrm{d}\sigma_t}{\mathrm{d}t}\) to denote time derivatives. It was shown in \cite{lipman2023flow} that there exists a probability flow ODE, \(\dot{x}_t = v(x_t,t)\), whose solution follows the same time marginals as the process in eq.~(\ref{eq:time_dependent_process}), where \(v(x_t,t)\) denotes the true drift field transporting \(p_0(x)\) to \(p_1(x)\).

To enable sampling from the data distribution, one has to train a neural network \( v_\theta(x_t, t) \) to approximate  \( v(x_t, t) \). 
This is done by regressing the network output toward the optimal vector field using \emph{conditional flow matching} \citep{lipman2023flow, hyvarinen2005estimation}, minimizing
\begin{equation}\label{eq:velocity}
\mathcal{L}_{\mathrm{velocity}}(\theta) := 
\mathbb{E}_{x^{\ast}, \varepsilon, t} 
\bigl[\| v_\theta(x_t, t) - \dot{\alpha}_t x^{\ast} - \dot{\sigma}_t \varepsilon \|^2 \bigr],
\end{equation}

Beyond the ODE view, there also exists a reverse \emph{stochastic differential equation} (SDE) with the same time marginals:
\begin{equation}\label{eq:sde}
\mathrm{d}x_t = v(x_t,t)\,\mathrm{d}t - g^2(t)\, s(x_t,t)\,\mathrm{d}t + g(t)\,\mathrm{d}\bar{w}_t,
\end{equation}
where \(s(x_t,t)=\nabla_{x_t}\log p_t(x_t)\) is the score, \(g(t)\) is the diffusion coefficient, and \(\bar{w}_t\) is a standard Wiener process running backward in time. Notably, the score can be expressed directly in terms of the velocity field:
\begin{equation}
s(x_t, t) = \sigma_t^{-1} \cdot
\frac{\alpha_t v(x_t, t) - \dot{\alpha}_t x_t}
{\dot{\alpha}_t \sigma_t - \alpha_t \dot{\sigma}_t}.
\label{eq:score}
\end{equation}

Eq. (\ref{eq:score}) hints that the score parameterization used in diffusion models and 
the velocity parameterization used in flow-based models can lead to equivalent 
formulations of the sampling process. In practice, diffusion models correspond to a 
stochastic discretization of the probability flow ODE, while flow-based models correspond 
to its deterministic counterpart \cite{lai2025principles}.

\subsection{Diffusion Models with Representation Guidance}\label{sec:rep_align}

Diffusion models learn rich, discriminative features in their hidden states, which are
crucial for their generative performance \citep{xiang2023}. Nevertheless, their learned
representations still fall behind on downstream tasks
when compared with state-of-the-art self-supervised visual encoders,
\citep{assran2023self, simeoni2025dinov3}. This representation gap has been identified
as a key bottleneck for improving the generative performance of diffusion models, as 
richer and more semantically structured representations have been shown to enhance 
sample fidelity \citep{yu2024repa}. To address this, a growing line of work introduces representation guidance, which conditions diffusion models on external representations from pretrained self-supervised encoders \citep{li2024return}. Typically these approaches fall into two categories: 
(i) conditioning the diffusion model on external representations, either learned,
\citep{li2024return}, or retrieved from a database of samples, 
\citep{blattmann2022retrieval, hu2023self, sheynin2022knn}; and 
(ii) explicitly aligning the model’s internal representations with those of large 
pretrained encoders, thereby encouraging more semantically meaningful feature-spaces, 
\citep{tian2025u, hu2023self, leng2025repa}.


Within the \textsc{Repa} framework, feature-level alignment is applied directly to intermediate representations of the diffusion model. In particular, let the diffusion model consist of \(L\) transformer blocks, and \(\ell \in \{1,\dots,L\}\) denote the layer at which features are extracted. The quantity \(h_t \in \mathbb{R}^{N\times D_2}\) denotes the matrix of patch tokens obtained after the first \(\ell\) transformer blocks when the model processes an input image \(x\) at timestep \(t\). Here, \(N\) is the number of patch tokens and \(D_2\) the embedding dimension, with \(h_t^{[n]} \in \mathbb{R}^{D_2}\) representing the \(n\)-th patch token. For convenience, this is written as
$
h_t = \textsc{DiffEnc}(x_t,t),
$
where $x_t$ denotes the noisy image and \textsc{DiffEnc} returns the patch-level hidden representation after the first \(\ell\) blocks of the diffusion model’s transformer encoder. \textsc{Repa} enforces patch-wise similarity between the \(D_1\)-dimensional DINOv2 representation of the target image \(x\), i.e., \(f^{[n]}_{\text{DINOv2}}(x) \in \mathbb{R}^{D_1}\) for patches \(n=1,\dots,N\), and the intermediate hidden states \(h^{[n]}_t\), by maximizing the following objective during training:
\begin{equation}\label{eq:6}
    \textsc{Repa}(x, h_t)
    = \frac{1}{N} \sum_{n=1}^{N} 
    \cos\!\left(f^{[n]}_{\text{DINOv2}}(x),\, g_{\phi}\!\big(h^{[n]}_{t}\big)\right),
\end{equation}
where \(g_{\phi} : \mathbb{R}^{D_2} \to \mathbb{R}^{D_1}\) is a learnable multi-layer perceptron (MLP) that projects hidden states into the DINOv2 space. By jointly optimizing \(\mathcal{L}_{\text{REPA}}\) with the standard diffusion loss, the model learns internal representations that are semantically richer and better aligned with high-level visual features, ultimately yielding higher-quality image generation and faster convergence during training.

\subsection{Inverse Problems with Diffusion and Flow Models}
\label{sec:inverse}

When solving inverse problems the goal is to recover the clean signal $x_0$ from noisy or degraded measurements $y$.  
In the flow- and diffusion-based framework, this is accomplished by replacing the unconditional score function in eq.~(\ref{eq:sde}) with the conditional score $\nabla_{x_t} \log p(x_t \mid y)$. Applying Bayes’ rule, the conditional score can be decomposed as
\begin{equation*}
\nabla_{x_t} \log p(x_t \mid y)
=
\nabla_{x_t} \log p_t(x_t)
+
\nabla_{x_t} \log p(y \mid x_t),
\label{eq:conditional_score}
\end{equation*}
Using this decomposition, the reverse-time SDE becomes
\begin{equation*}
dx_t
=
\bigl(
v(x_t,t)
- g^2(t)\, \nabla_{x_t} \log p(x_t \mid y)
\bigr)\, dt
+ g(t)\, d\bar{w}_t .
\label{eq:reverse_sde_conditional}
\end{equation*}
The unconditional score $\nabla_{x_t} \log p_t(x_t)$ is provided by a pretrained diffusion or
flow-based model, while the likelihood term $\nabla_{x_t} \log p(y \mid x_t)$ enforces
measurement consistency but is generally intractable, as it requires marginalizing over all possible clean signals $x_0$.
One of the most prominent approaches is {\it Diffusion Posterior Sampling (DPS)}, \citep{chung2023diffusion}, which uses the following approximation:
\begin{equation*}\label{eq:4}
    p(y \mid x_t) \approx p\left(y \mid \hat{x}_0 = \mathbb{E}[x_0 \mid x_t]\right),
\end{equation*}
where $\mathbb{E}[x_0 \mid x_t]$ denotes the conditional expectation of $x_0$ given $x_t$.
In many practical scenarios, performing the diffusion process directly in pixel space is computationally prohibitive. 
Latent diffusion models (LDMs) address this limitation by operating in a compressed latent space. 
Let $\mathcal{E}: \mathbb{R}^{k} \rightarrow \mathbb{R}^{d}$ be an encoder mapping an image $x_0$ to its latent representation $z_0 = \mathcal{E}(x_0)$, 
and let $\mathcal{D}$ denote the corresponding decoder such that $\hat{x}_0 = \mathcal{D}(z_0) \approx x_0$. Building on this formulation, recent work has proposed latent-space counterparts of pixel-based inverse problem solvers. 
For example, DPS has been formulated in latent space by redefining the measurement likelihood as
\begin{equation*}\label{eq:5}
    p(y \mid z_t) \approx p\!\left(y \,\middle|\, \hat{x}_0 = \mathcal{D}\!\big(\mathbb{E}[z_0 \mid z_t]\big)\right).
\end{equation*}

This straightforward formulation, commonly referred to as Latent DPS, has been observed to introduce artifacts, largely due to the nonlinearity of the decoder. 
To mitigate this effect, \cite{rout2023solving} proposed a regularization term encouraging the latent variables to remain close to a fixed point of the encoder–decoder mapping, 
while \cite{song2024solving} enforced data consistency during optimization followed by a projection of the latent variables back onto the noisy manifold.

\section{Proposed Approach}\label{sec:approach}
Given a noisy observation \(y \in \mathbb{R}^m\) of an unknown signal 
\(x \in \mathbb{R}^k\), our goal is to sample from \(p_\theta(x_0 \mid y)\) using a pretrained diffusion or flow-based models. 
We aim to apply the $\textsc{Repa}$ framework to inverse problems. 
However, a  key challenge is that representation alignment requires access to the ground-truth 
signal, which is not available in this setting. To overcome this, we must choose an alternative 
input to the DINOv2 encoder that can stand in for the missing ground truth. In particular, we 
need an approximation $\bar{x}_0$ of the unknown $x_0$ that produces a \emph{proxy 
representation} $c_{\text{proxy}} \in \mathbb{R}^{N \times D_1}$, where \(N\) denotes the number of patch 
tokens and \(D_1\) is their feature dimension. This representation should satisfy
\begin{equation}
    c_{\text{proxy}} \equiv f_{\text{DINOv2}}(\bar{x}_0) 
    \approx f_{\text{DINOv2}}(x_0).
    \label{eq:dino_rep}
\end{equation}

We then introduce a \emph{tilted distribution}, \cite{pachebat2025iterative}, that incorporates representation
alignment as reward:
\begin{equation}
\begin{aligned}
\widetilde p(x_t \mid y)
&\propto
p(y \mid x_t)\, p_t(x_t)
\exp\!\bigl(\lambda\, \textsc{Repa}(\bar{x}_0, h_t)\bigr),
\end{aligned}
\label{eq:post_tilted}
\end{equation}
where \(h_t = \textsc{DiffEnc}(x_t,t)\) denotes the intermediate diffusion representation
and \(\lambda > 0\) controls the strength of the alignment term.
The likelihood \(p(y \mid x_t)\) enforces measurement consistency, while
\(p_t(x_t)\) corresponds to the unconditional prior induced by the generative model. The \textsc{Repa} term defined in Eq.~\eqref{eq:6} encourages alignment
between the diffusion representation of \(x_t\) and the proxy DINOv2 features, thereby biasing the sampling process toward semantically consistent
reconstructions.

\textbf{On the selection of $c_{\text{proxy}}$.}
To address the fact that representation alignment~\citep{yu2024repa} is not directly
applicable to inverse problems where the ground-truth image is unavailable at inference
time, we construct $c_{\text{proxy}}$ from an approximate reconstruction. We initialize
the proxy using the DINOv2 features of the available observation and, as the reverse
diffusion process progresses and the reconstruction becomes more informative, we
gradually replace this proxy with the features of the model’s current denoised estimate,
namely \(f_{\text{DINOv2}}(\mathbb{E}[x_0 \mid x_t])\). In both cases, we rely on the
robustness of pretrained DINOv2 features, which, as shown in our experiments, remain
stable under the degradations present in super-resolution and deblurring tasks. An ablation in Appendix~\ref{sec:dino_rob} quantifies the reliability of this proxy across
tasks. We further compare against alternative encoders, such as CLIP~\cite{radford2021learning},
in Appendix~\ref{subseq:CLIP}, where DINOv2 consistently yields improved performance.
We also examine the behavior of the method across a broader range of degradation levels
in Appendix~\ref{subsec:severe_deg}, where the proxy-based alignment remains effective
even in more challenging settings.

\paragraph{On the availability of the alignment mapping.}
The regularizer in \eqref{eq:post_tilted} relies on a projection head $g_\phi$
mapping diffusion representations into the \text{DINOv2} feature space. For
models trained with representation alignment, this mapping is available
from pretraining. More generally, our framework can also be applied to arbitrary
pretrained diffusion models by learning a post-hoc mapping between
diffusion representations and \text{DINOv2} features while keeping the diffusion
model itself frozen. Since only the projection head $g_\phi$ is optimized, this
additional training is lightweight and inexpensive, compared to diffusion-model pretraining. We refer the
reader to Appendix~\ref{app:mapping} for further details. As we show in Appendix~\ref{sec:independently_pretrained_results}, this post-hoc adaptation preserves the
benefits of representation alignment for inverse problems and enables \textsc{repa} regularization to be used
with existing generative models not previously pretrained with \textsc{repa}.

\noindent\textbf{Extension to latent diffusion models.}
The proposed representation-alignment strategy applies naturally to latent diffusion
models.
Let $z_t$ denote the latent state at timestep $t$.  
In this setting, the tilted distribution becomes
\begin{equation*}
\widetilde p(z_t \mid y)
\propto 
 p(y \mid z_t)p_t(z_t)\exp\left(\lambda\textsc{Repa}(\bar{x}_0, h_t)\right),
\end{equation*}
With a slight abuse of notation, we use \textsc{DiffEnc} to denote the diffusion encoder in
both pixel-space and latent-space models, and write
\(h_t = \textsc{DiffEnc}(z_t,t)\) for latent diffusion. 

Our general algorithm for a latent-space velocity-based sampler is summarized in
Algorithm~\ref{alg:algorithm}.  
Step~6 is intentionally kept abstract; as discussed in Section~\ref{sec:inverse}, 
different inverse-problem solvers approximate this term in different ways.  
The proposed regularizer is compatible with standard diffusion-based reconstruction
pipelines.  
In our experiments, we instantiate the framework using two state-of-the-art latent
diffusion methods, with further implementation details provided in
Appendix~\ref{sec:latent_dps} and Appendix~\ref{sec:resample}. We further provide an ablation study in Appendix~\ref{subsec:ablations} analyzing the effect of all hyperparameters of our framework, including the cutoff timestep.
\begin{algorithm}[H]
\caption{$\textsc{Repa}$-regularized Inverse Algorithm}
\label{alg:algorithm}
\begin{algorithmic}[1]
\Require Flow or diffusion model $u_\theta$, measurement $y$, 
learning rate $\eta$, regularizer strength $\lambda$, proxy representation $c_{\text{proxy}}$, timestep $t_\text{cutoff}$, Decoder $\mathcal{D}$, Initialize $z_T \sim \mathcal{N}(0,I)$, 
\For{$t = T, \dots, 1$}
    \State $z_{t-1} \gets z_t - \frac{1}{T} \cdot u_\theta(z_t,t) + \eta\, \nabla_{z_t} \log p(y \mid z_t)$
    \State \textcolor{blue}{$z_{t-1} \gets z_{t-1}
    + \lambda\, \nabla_{z_t}
    \sum_{n=1}^{N}
    \cos\!\bigl(
    c^{[n]}_{\text{proxy}},$}
    \textcolor{blue}{$
    g_{\phi}\!\big(\mathrm{DiffEnc}^{[n]}(z_t,t))
    \bigr)$}
    \If{$t < t_{\text{cutoff}}$}
        \State $c_{\text{proxy}} \gets
        f_{\mathrm{DINOv2}}\!\bigl(
        \mathcal{D}(\mathbb{E}[z_0 \mid z_t])
        \bigr)$
    \EndIf
\EndFor
\State \Return $z_0$
\end{algorithmic}
\end{algorithm}

\section{Theoretical Results}

Recent advances in generative modeling have demonstrated that objectives defined in deep
feature spaces, rather than purely pixel-based discrepancies, produce reconstructions that
better capture perceptual similarity~\citep{dosovitskiy2016generating, zhang2018unreasonable, fu2023dreamsim}.
Motivated by this observation, our goal in this section is to clarify how the $\textsc{Repa}$ regularizer
fits within this broader class of feature-based objectives. In particular, we present two
theoretical results that characterize how $\textsc{Repa}$ acts on (i) the DINOv2-space representations
of reconstructed images and (ii) the latent diffusion state itself.

First, we show that maximizing the \textsc{Repa} objective encourages alignment between the DINOv2 feature distributions of reconstructed and ground-truth images, yielding an interpretation in terms of maximum mean discrepancy (MMD). Second, we study how $\textsc{Repa}$
affects the latent states of the diffusion model during the reverse process. Under suitable
assumptions, we show that a $\textsc{Repa}$ update moves the current diffusion state toward the
latent representation of the clean image. This provides a complementary
view: $\textsc{Repa}$ influences not only the DINOv2 representation of the reconstruction but also the
latent state that guides the sampling trajectory.  

Before presenting the main results, we introduce two quantities that capture, respectively, the alignment between feature spaces and the accuracy of the proxy representations used in our analysis.

\begin{definition}[Misalignment between DINOv2 and diffusion representations]
\label{def:misrepa}
For an image $x$ and diffusion timestep $t$, we define the representation misalignment as
\[
\textstyle
\mathrm{MisREPA}(x,t)
:= \mathbb{E}_{n}\,
\|f_{\!\mathrm{DINOv2}}^{[n]}(x)
- g_\phi(\mathrm{DiffEnc}^{[n]}(x,t))\|_2^2,
\]
where the expectation is taken uniformly over patch indices
$n \in \{1,\dots,N\}$.
Here $f_{\mathrm{DINOv2}}^{[n]}(x)$ and $\mathrm{DiffEnc}^{[n]}(x,t)$ denote the $n$-th patch features
from the DINOv2 encoder and the diffusion model encoder, respectively. The mapping $g_\phi$
is the projection head introduced in Section~\ref{sec:rep_align}\footnote{When the timestep argument is omitted, all $t$-dependent quantities are evaluated at $t=0$.
In particular, $\mathrm{MisREPA}(x) \equiv \mathrm{MisREPA}(x,0)$ and
$\mathrm{DiffEnc}^{[n]}(x) \equiv \mathrm{DiffEnc}^{[n]}(x,0)$.}.
\end{definition}
\begin{definition}[Proxy approximation error]
For a ground-truth image $x$ and its proxy approximation $\bar{x}$ (defined in
Section~\ref{sec:approach}), we define the proxy approximation error as
\[
\mathrm{ApproxErr}(x,\bar{x})
:=
\mathbb{E}_{n}\,
\bigl\|
    f_{\mathrm{DINOv2}}^{[n]}(x)
    -
    f_{\mathrm{DINOv2}}^{[n]}(\bar{x})
\bigr\|_2^2,
\]
where the expectation is taken uniformly over patch indices
$n \in \{1,\dots,N\}$.
\end{definition}

In practice, both quantities defined above are typically small. The feature misalignment
$\mathrm{MisREPA}(x,t)$ is mitigated through the explicit alignment of DINOv2 and diffusion
encoder representations during training~\citep{yu2024repa}, while the approximation error
$\mathrm{ApproxErr}(x,\bar{x})$ is reduced by constructing proxy features $\bar{x}$ that
serves as reliable substitutes for the unavailable ground-truth features
(see Section~\ref{sec:approach}). Robustness of DINOv2 features to common image
degradations~\citep{oquab2023dinov2}, ensures that the resulting approximation gap remains small across the inverse problems we study.

\textbf{Feature-Space Alignment: $\textsc{Repa}$ as an MMD Surrogate.}
Building on the definitions introduced above, we now establish a connection between the
$\textsc{Repa}$ regularizer and the statistical discrepancy between DINOv2 features of clean and
reconstructed images.  

\begin{proposition}
\label{lem:repa_mmd_mine}
Assume $x \sim p_X$ is a ground-truth image and $\hat{x} \sim p_{\hat{X}\mid Y}$ is its
corresponding reconstruction, where $p_{\hat{X}\mid Y}$ is the distribution induced
by any reconstruction method. Assume further that all DINOv2 feature embeddings are
$\ell_2$--normalized. Let $\bar{x}$ be a proxy approximation of $x$.
Define the empirical mean DINOv2 feature embedding as
$
\mu_f(x)
:=
\mathbb{E}_{n}\!\left[f_{\mathrm{DINOv2}}^{[n]}(x)\right].
$
Then the expected REPA alignment satisfies
\begin{equation*}
\mathbb{E}_{\hat{x},\bar{x}}\!\left[\text{REPA}(\bar{x},\hat{x})\right]
\le
1 - \tfrac{1}{8}\,\textsc{MMD}_{\mathrm{DINOv2}}\!\left(p_X,p_{\hat X}\right)
+ \tfrac{1}{2}\,\mathbb{E}_{x,\bar x}\!\left[\mathrm{ApproxErr}(x,\bar{x})\right]
+ \tfrac{1}{4}\,\mathbb{E}_{\hat x}\!\left[\mathrm{MisREPA}(\hat{x})\right].
\end{equation*}
where
$
\textsc{MMD}_{\mathrm{DINOv2}}\!\left(p_X,\,p_{\hat X}\right)
=
\bigl\|
    \mathbb{E}_{x}[\mu_f(x)]
    -
    \mathbb{E}_{\hat x}[\mu_f(\hat{x})]
\bigr\|_2^2
$
is the maximum mean discrepancy between $p_X$ and $p_{\hat X}$ in the DINOv2 feature space.
\end{proposition}
The proof is provided in the Appendix \ref{sec:proofs}. Proposition~\ref{lem:repa_mmd_mine} establishes a connection between the representation alignment term and the maximum mean discrepancy (MMD)~\citep{arbel2019maximum}, which quantifies perceptual divergence between feature distributions.
This relationship indicates that \emph{maximizing} the $\textsc{Repa}$ term encourages the \emph{minimization} of a divergence measure captured by the MMD together with additional penalties accounting for feature misalignment between diffusion and DINOv2 representations, as well as an approximation error arising from the use of proxy rather than ground-truth features for alignment.

When these terms are sufficiently small, Proposition~\ref{lem:repa_mmd_mine} implies that maximizing the $\textsc{Repa}$ term effectively corresponds to minimizing the maximum mean discrepancy (MMD) between the DINOv2 feature distributions of reconstructed and ground-truth images.
This interpretation reveals that $\textsc{Repa}$ regularization offers a principled, representation-based mechanism for enforcing perceptual consistency.

\textbf{Effect of REPA on Diffusion Representations.} 
We now study how a
single \textsc{Repa} update influences the latent diffusion state relative to that of the
clean image.

\begin{proposition}[informal]
\label{prop:zspace_contraction_repa}
Consider an inverse problem with observation \(y \in \mathbb{R}^m\) generated from a
clean image \(x \sim p_X\), and let \(\bar{x}\) denote the proxy used by REPA.
Let \(z_t \in \mathbb{R}^d\) be the diffusion state at time \(t\),
\(z_t^{(\mathrm{REPA})}\) its REPA-regularized update. Let $z^*\in\mathbb{R}^d$ denote the latent variable associated with the clean image $x$. Under regularity   assumptions on the Jacobian
of the diffusion encoder \(J_{\mathrm{DiffEnc}}\) and the mapping \(g_\phi\) (see
Appendix~\ref{assump:repa}), there exists a radius \(r>0\) such that,
whenever \(\|z_t - z^*\|_2 < r\), the REPA update induces a contraction in the
diffusion state space. In particular,
\begin{equation*}
\begin{aligned}
&\|z_t^{(\mathrm{REPA})}-z^*\|_2
\le
C_1\,\|z_t-z^*\|_2
+ e(x),
\quad e(x)
:=
C_2\left(
\sqrt{\mathrm{ApproxErr}(x,\bar x)}
+
\sqrt{\mathrm{MisREPA}(x)}
\right),
\end{aligned}
\end{equation*}
where \(C_1 \in (0,1)\) and \(C_2 > 0\) are constants that depend on the conditioning of \(J_{\mathrm{DiffEnc}}\) and \(g_\phi\).
\end{proposition}

\begin{figure*}[t]
\centering


\begin{subfigure}[t]{0.48\textwidth}
  \centering
  \begin{tabular}{@{}c@{\hspace{2pt}}c@{\hspace{2pt}}c@{\hspace{2pt}}c@{}}
    {\tiny \textbf{Measurement}} &
    {\tiny \textbf{Without REPA}} &
    {\tiny \textbf{With REPA}} &
    {\tiny \textbf{Reference}} \\[0.3ex]

    \includegraphics[width=0.25\linewidth]{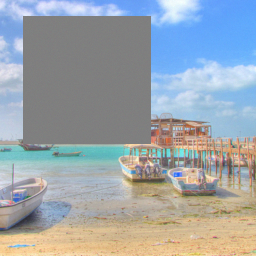} &
    \includegraphics[width=0.25\linewidth]{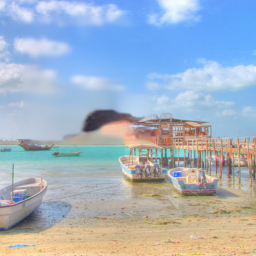} &
    \includegraphics[width=0.25\linewidth]{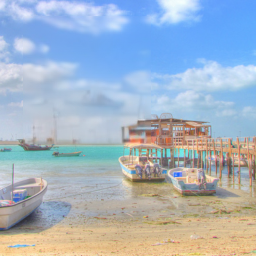} &
    \includegraphics[width=0.25\linewidth]{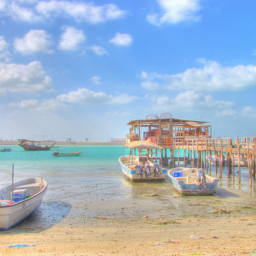} \\[-0.2ex]

    \includegraphics[width=0.25\linewidth]{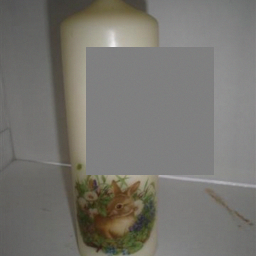} &
    \includegraphics[width=0.25\linewidth]{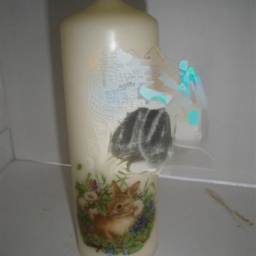} &
    \includegraphics[width=0.25\linewidth]{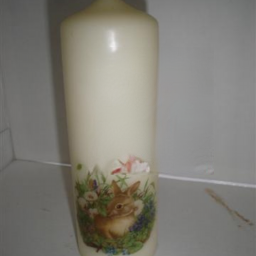} &
    \includegraphics[width=0.25\linewidth]{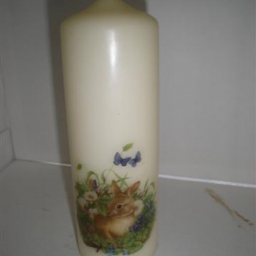} \\
  \end{tabular}
\end{subfigure}
\hspace{4pt}
\begin{subfigure}[t]{0.48\textwidth}
  \centering
  \begin{tabular}{@{}c@{\hspace{2pt}}c@{\hspace{2pt}}c@{\hspace{2pt}}c@{}}
    {\tiny \textbf{Measurement}} &
    {\tiny \textbf{Without REPA}} &
    {\tiny \textbf{With REPA}} &
    {\tiny \textbf{Reference}} \\[0.3ex]

    \includegraphics[width=0.25\linewidth]{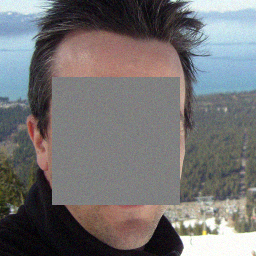} &
    \includegraphics[width=0.25\linewidth]{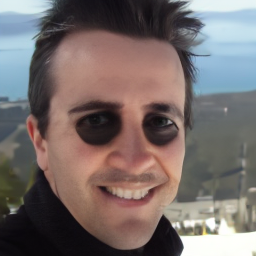} &
    \includegraphics[width=0.25\linewidth]{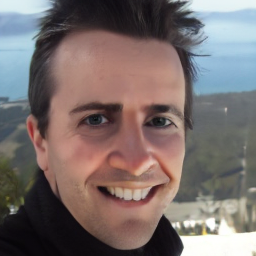} &
    \includegraphics[width=0.25\linewidth]{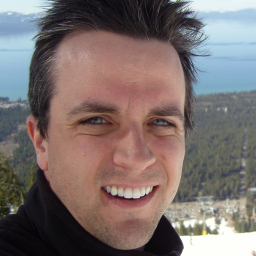} \\[-0.2ex]

    \includegraphics[width=0.25\linewidth]{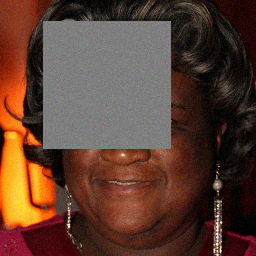} &
    \includegraphics[width=0.25\linewidth]{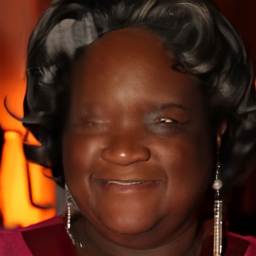} &
    \includegraphics[width=0.25\linewidth]{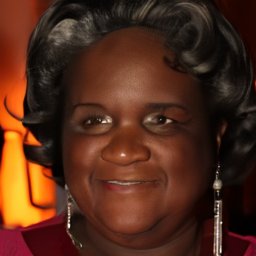} &
    \includegraphics[width=0.25\linewidth]{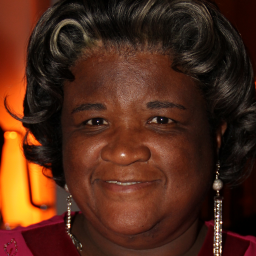} \\
  \end{tabular}
\end{subfigure}

\vspace{6pt}


\begin{subfigure}[t]{0.48\textwidth}
  \centering
  \begin{tabular}{@{}c@{\hspace{2pt}}c@{\hspace{2pt}}c@{\hspace{2pt}}c@{}}
    {\tiny \textbf{Measurement}} &
    {\tiny \textbf{Without REPA}} &
    {\tiny \textbf{With REPA}} &
    {\tiny \textbf{Reference}} \\[0.3ex]

    \includegraphics[width=0.25\linewidth]{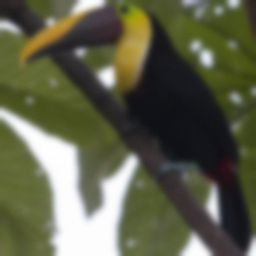} &
    \includegraphics[width=0.25\linewidth]{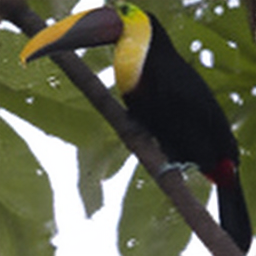} &
    \includegraphics[width=0.25\linewidth]{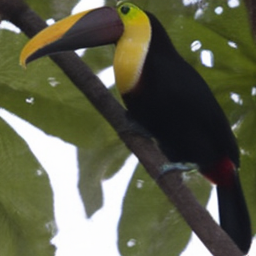} &
    \includegraphics[width=0.25\linewidth]{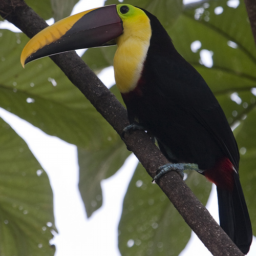} \\

    \includegraphics[width=0.25\linewidth]{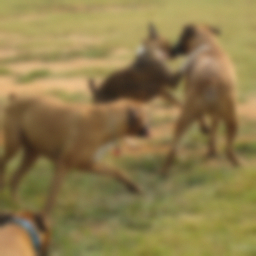} &
    \includegraphics[width=0.25\linewidth]{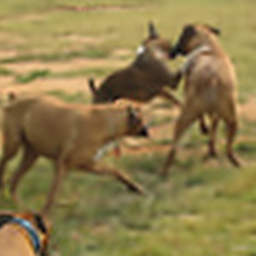} &
    \includegraphics[width=0.25\linewidth]{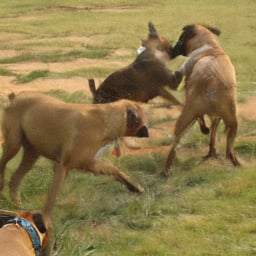} &
    \includegraphics[width=0.25\linewidth]{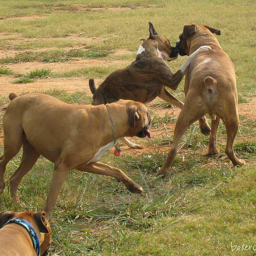} \\[-0.2ex]

  \end{tabular}
\end{subfigure}
\hspace{4pt}
\begin{subfigure}[t]{0.48\textwidth}
  \centering
  \begin{tabular}{@{}c@{\hspace{2pt}}c@{\hspace{2pt}}c@{\hspace{2pt}}c@{}}
    {\tiny \textbf{Measurement}} &
    {\tiny \textbf{Without REPA}} &
    {\tiny \textbf{With REPA}} &
    {\tiny \textbf{Reference}} \\[0.3ex]

    \includegraphics[width=0.25\linewidth]{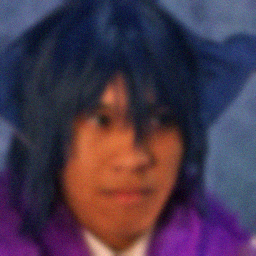} &
    \includegraphics[width=0.25\linewidth]{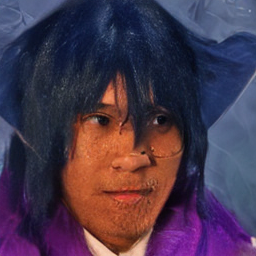} &
    \includegraphics[width=0.25\linewidth]{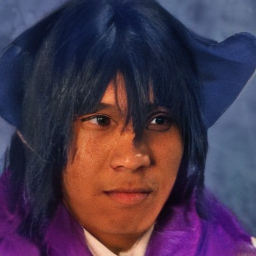} &
    \includegraphics[width=0.25\linewidth]{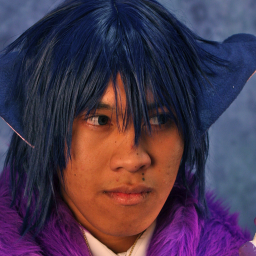} \\[-0.2ex]

    \includegraphics[width=0.25\linewidth]{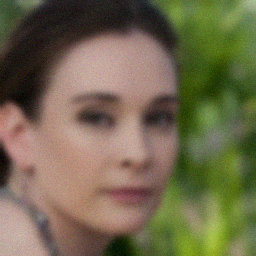} &
    \includegraphics[width=0.25\linewidth]{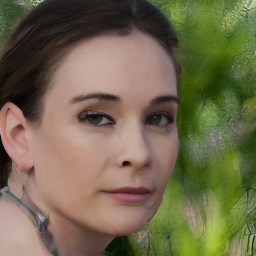} &
    \includegraphics[width=0.25\linewidth]{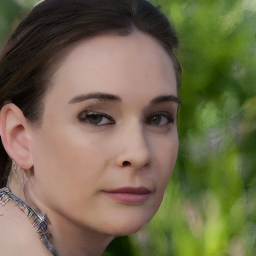} &
    \includegraphics[width=0.25\linewidth]{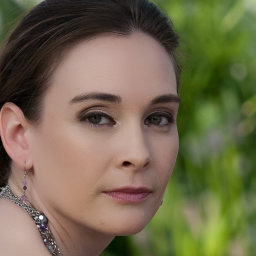} \\
  \end{tabular}
\end{subfigure}
\caption{
Qualitative comparison of inverse problem reconstruction with and without \textsc{Repa}.
The top block corresponds to box inpainting and the bottom block to Gaussian deblurring;
Within each panel, the first row corresponds to Latent DPS and the second row to ReSample.
}
\label{fig:repa_qualitative_tasks}
\end{figure*}

The proof is provided in Appendix~\ref{sec:proofs}.
Proposition~\ref{prop:zspace_contraction_repa} characterizes how the \textsc{Repa} update
modifies the latent diffusion state.
The inequality shows that, when the current state \(z_t\) is reasonably close to the
clean latent code \(z^*\), a single \textsc{Repa} step contracts their distance up to two
additive terms: the proxy approximation error and the
representation misalignment, both of which are expected to be
small in practice.
This contraction therefore holds in a local regime around \(z^*\), which corresponds in
practice to stages of the diffusion process where the global structure of the
reconstruction is already correct and the remaining errors are primarily fine-scale
details and artifacts.
In this regime, \textsc{Repa} acts as a perceptual regularizer, refining local image
details and correcting artifacts that are not effectively addressed by the baseline
update alone.
Qualitative examples illustrating this behavior are provided in Figure \ref{fig:repa_qualitative_tasks} and the
Appendix \ref{sec:qualitative}.

%

\section{Experimental Results}\label{sec:experiments}

In this section, we present experimental results demonstrating the effectiveness of our alignment regularizer. 
We evaluate reconstruction quality using four widely adopted metrics: PSNR (peak signal-to-noise ratio), SSIM (structural similarity index)~\citep{wang2004image}, LPIPS (learned perceptual image patch similarity)~\citep{zhang2018unreasonable}, and FID (Fréchet Inception Distance)~\citep{heusel2017gans}. 
Experiments are conducted on both the ImageNet and FFHQ datasets~\citep{deng2009imagenet, karras2019style}, with FFHQ images resized to $256 \times 256$. 
All metrics for both datasets are averaged over 100 images from their corresponding validation splits.

We consider four inverse problems: super-resolution, box inpainting, Gaussian deblurring, 
and motion deblurring. Following the standard corruption operators used in state-of-the-art 
diffusion-based inverse problem methods \citep{chung2023diffusion, zhang2025improving} we adopt the same degradation models in our 
experiments. For super-resolution, images are downsampled by a factor of $4$. For box 
inpainting, we mask out a $128 \times 128$ square region. Gaussian deblurring is performed 
using a $61 \times 61$ kernel with standard deviation $3.0$. Motion blur is simulated using 
a kernel with size $61 \times 61$ and intensity $0.5$. For ImageNet we use additive Gaussian noise with standard deviation $0.01$. 
For FFHQ we increase the noise level to $0.05$ in order to obtain a more challenging 
reconstruction setting. We also include results about two non-linear inverse problems for the ImageNet dataset in Appendix \ref{subsec:non_linear}.

\begin{table*}[t]
\centering
\small
\setlength{\tabcolsep}{3pt}
\caption{Performance comparison on ImageNet and FFHQ across inverse tasks.}
\label{tab:inverse_results}
\begin{tabular}{llcccc|cccc}
\toprule
\textbf{Task} & \textbf{Method}
& \multicolumn{4}{c}{\textbf{ImageNet}}
& \multicolumn{4}{c}{\textbf{FFHQ}} \\
\cmidrule(lr){3-6}\cmidrule(lr){7-10}
& & LPIPS$\downarrow$ & FID$\downarrow$ & PSNR$\uparrow$ & SSIM$\uparrow$
& LPIPS$\downarrow$ & FID$\downarrow$ & PSNR$\uparrow$ & SSIM$\uparrow$ \\
\midrule
\multirow{6}{*}{$4\times$ SR}
& Latent DPS & 0.238 & 123.82 & 26.88 & 0.732 & 0.188 & 56.69 & 28.99 & 0.814 \\
& Latent DPS + REPA & 0.217 & 86.88 & 26.82 & 0.731 & 0.176 & 51.27 & 29.12 & 0.819 \\
& Resample & 0.208 & 97.02 & 26.70 & 0.736 & 0.178 & 52.82 & 29.06 & 0.814 \\
& Resample + REPA & \textbf{0.197} & \textbf{74.70} & 26.99 & 0.740 & \textbf{0.167} & \textbf{46.36} & \textbf{29.17} & 0.818 \\
& DPS & 0.244 & 92.13 & 24.42 & 0.681 & 0.182 & 54.19 & 27.67 & 0.804 \\
& Latent DAPS & 0.252 & 120.57 & \textbf{27.13} & \textbf{0.744} & 0.206 & 74.82 & 28.95 & \textbf{0.820} \\
\midrule
\multirow{6}{*}{Inpainting}
& Latent DPS & 0.151 & 116.53 & 20.53 & 0.822 & 0.192 & 65.89 & 23.55 & 0.785 \\
& Latent DPS + REPA & \textbf{0.139} & 88.69 & 20.45 & 0.824 & \textbf{0.178} & \textbf{57.04} & 23.83 & 0.784 \\
& Resample & 0.153 & 121.98 & 20.53 & 0.812 & 0.192 & 67.99 & 23.86 & 0.792 \\
& Resample + REPA & 0.143 & \textbf{87.15} & 20.79 & \textbf{0.827} & \textbf{0.178} & 61.97 & 24.01 & 0.793 \\
& DPS & 0.191 & 97.95 & 19.11 & 0.769 & 0.190 & 104.53 & 20.02 & 0.786 \\
& Latent DAPS & 0.318 & 188.44 & \textbf{21.21} & 0.718 & 0.210 & 89.21 & \textbf{24.39} & \textbf{0.817} \\
\midrule
\multirow{6}{*}{Gaussian Deblur}
& Latent DPS & 0.288 & 152.96 & 25.76 & 0.648 & 0.192 & 60.65 & 28.15 & 0.783 \\
& Latent DPS + REPA & 0.256 & 102.99 & 25.66 & 0.669 & 0.186 & 55.53 & 28.21 & 0.787 \\
& Resample & 0.259 & 115.47 & 26.19 & 0.699 & 0.172 & 56.41 & 27.01 & 0.753 \\
& Resample + REPA & \textbf{0.223} & \textbf{89.67} & \textbf{26.49} & \textbf{0.707} & \textbf{0.168} & \textbf{52.73} & 27.17 & 0.756 \\
& DPS & 0.366 & 157.73 & 19.55 & 0.461 & 0.177 & 53.18 & 27.19 & 0.789 \\
& Latent DAPS & 0.291 & 159.67 & 26.02 & 0.692 & 0.229 & 104.43 & \textbf{28.73} & \textbf{0.809} \\

\midrule

\multirow{6}{*}{Motion Deblur}
& Latent DPS & 0.249 & 129.08 & 27.19 & 0.738 & 0.170 & 52.14 & 27.20 & 0.773 \\
& Latent DPS + REPA & 0.225 & 90.23 & 27.01 & 0.735 & 0.165 & 47.18 & 27.16 & 0.772 \\
& Resample & 0.210 & 89.95 & 27.57 & 0.739 & 0.157 & 52.19 & 28.36 & 0.791 \\
& Resample + REPA & \textbf{0.192} & \textbf{75.11} & \textbf{27.80} & \textbf{0.766} & 0.151 & 50.02 & 28.41 & 0.794 \\
& DPS & 0.242 & 89.06 & 24.17 & 0.678 & \textbf{0.146} & \textbf{43.32} & 27.28 & 0.793 \\
& Latent DAPS & 0.264 & 125.18 & 27.25 & 0.744 & 0.191 & 76.21 & \textbf{29.77} & \textbf{0.836} \\

\bottomrule
\end{tabular}
\end{table*}

\begin{figure*}[!t]
    \centering

    \begin{subfigure}[t]{0.24\textwidth}
        \includegraphics[width=\linewidth]{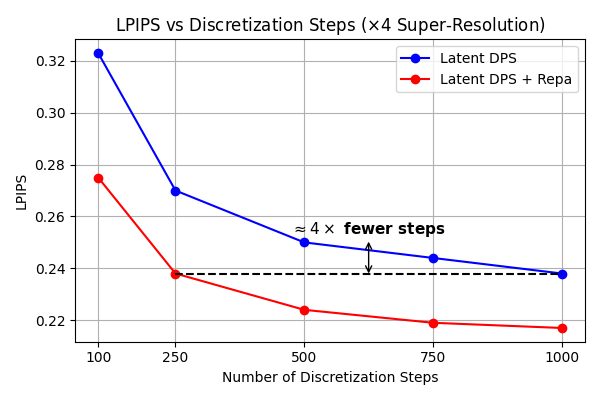}
        \caption{Super Resolution}
    \end{subfigure}
    \hfill
    \begin{subfigure}[t]{0.24\textwidth}
        \includegraphics[width=\linewidth]{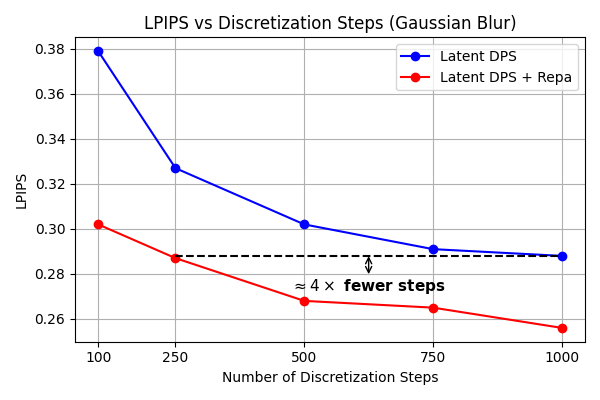}
        \caption{Gaussian Deblurring}
    \end{subfigure}
    \hfill
    \begin{subfigure}[t]{0.24\textwidth}
        \includegraphics[width=\linewidth]{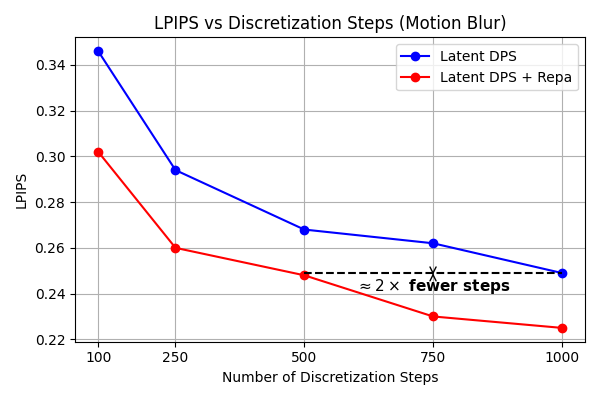}
        \caption{Motion Deblurring}
    \end{subfigure}
    \hfill
    \begin{subfigure}[t]{0.24\textwidth}
        \includegraphics[width=\linewidth]{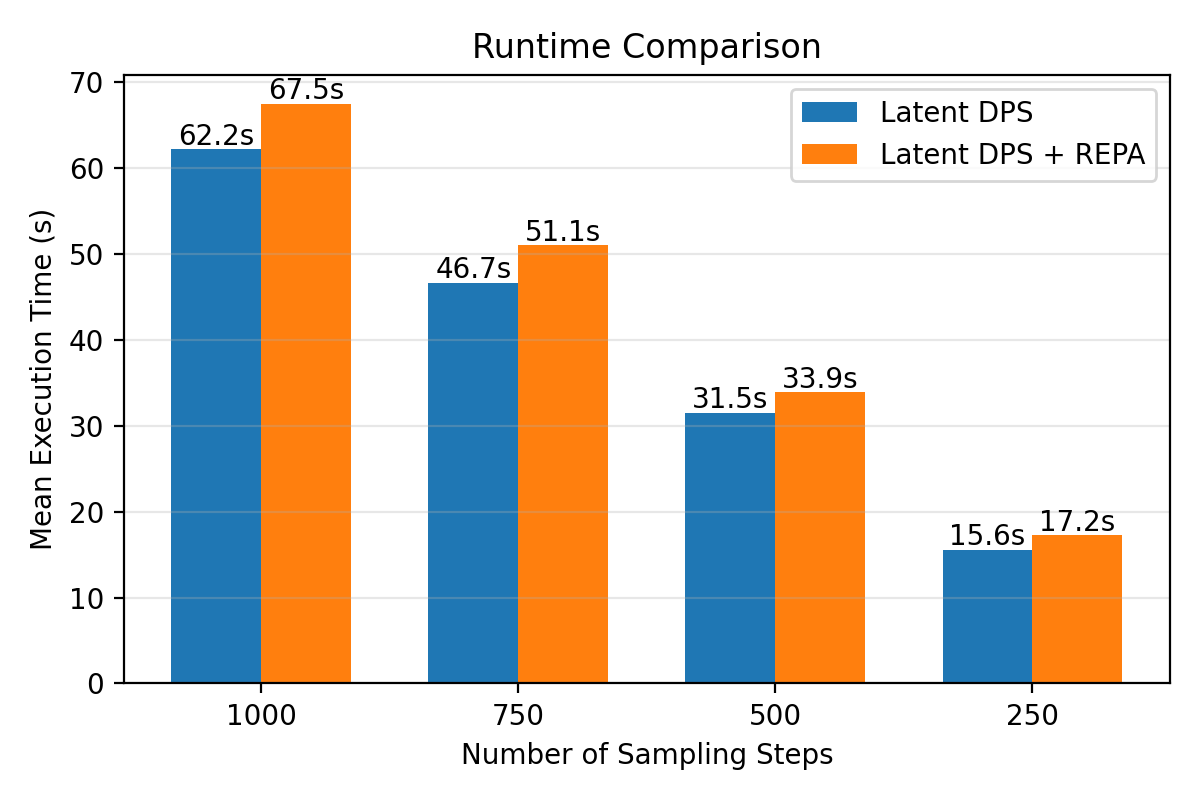}
        \caption{Runtime Comparison}
    \end{subfigure}

    \caption{
    Comparison of LPIPS as a function of sampling steps, together with runtime comparison.
    }
    \label{fig:similarity_corruption_levels_discretization}
\end{figure*}

\subsection{Effectiveness of the $\textsc{Repa}$ Regularizer}

This subsection evaluates how the $\textsc{Repa}$ regularizer influences reconstruction quality by 
applying it to two representative latent-space solvers. To this end, we instantiate 
Algorithm~\ref{alg:algorithm} with Latent DPS and ReSample \citep{song2024solving}. A detailed description of these methods and their 
integration with $\textsc{Repa}$ is provided in Appendix~\ref{sec:latent_dps}.  For ImageNet experiments, we use the latent diffusion model trained with representation alignment from \cite{yu2024repa}, together with its pretrained MLP that projects diffusion representations into the DINOv2 space. For FFHQ, we train a representation-aligned \textsc{SiT-Base} model from scratch following the same setup as \cite{yu2024repa}. In ImageNet experiments, the regularizer is applied to representations extracted after the eighth transformer block, while for FFHQ it is applied after the fourth block, we include an ablation study about this choice in Appendix \ref{subsec:effect_block}. Our approach can be extended to independently pretrained diffusion models by learning a mapping between the diffusion representations and
the \text{DINOv2} feature space. We study this setting in
Appendix~\ref{sec:independently_pretrained_results}. In this case, incorporating the \textsc{Repa}
regularizer continues to provide perceptual improvements for inverse problems.
In line with our theory, the method remains effective as long as the learned
mapping is sufficiently accurate, corresponding to a low $\mathrm{MisREPA}$ error.


Table~\ref{tab:inverse_results} reports quantitative results across all four inverse problems. Incorporating $\textsc{Repa}$ yields consistent improvements in perceptual metrics, notably reducing LPIPS and FID for both Latent DPS and ReSample while maintaining comparable PSNR and SSIM values.  
PSNR and SSIM remain comparable to the baseline, while perceptual metrics consistently improve. This behaviour is in line with the perception–distortion trade-off, whereby improvements in perceptual realism do not necessarily translate to gains in pixel-wise metrics. We include qualitative results in Appendix~\ref{sec:perception_distortion} illustrating this phenomenon. These results suggest that representation alignment guides the diffusion trajectory toward semantically consistent and perceptually realistic reconstructions that better match the target image.  
Figure~\ref{fig:repa_qualitative_tasks} presents qualitative comparisons for box inpainting and Gaussian deblurring, where reconstructions obtained with $\textsc{Repa}$ are visually sharper and more faithful to the ground truth. In addition to improving these latent solvers, we also compare our approach with other state-of-the-art reconstruction algorithms.  
For pixel-space methods, we consider DPS~\citep{chung2023diffusion}, evaluated using the pretrained diffusion model from~\cite{dhariwal2021diffusion}.  
For latent diffusion baselines, we include Latent DAPS, implemented with the conditional ImageNet latent diffusion model from~\cite{rombach2022high}.  
As shown in Table~\ref{tab:inverse_results}, the proposed alignment regularizer consistently achieves state-of-the-art performance in perception-based metrics across the considered inverse problems.


\textbf{Effect of REPA with Varying Discretization Steps.} 
Figure \ref{fig:similarity_corruption_levels_discretization} illustrates how LPIPS varies with the number of discretization steps across three inverse problems, together with the corresponding runtime comparison. Incorporating the REPA regularizer consistently reduces perceptual error compared to Latent DPS alone. More importantly, REPA enables comparable or better perceptual quality with substantially fewer sampling steps: for example, in super-resolution and Gaussian deblurring, REPA achieves the same LPIPS with roughly $4\times$ fewer steps, while in motion deblurring it achieves the same level with about $2\times$ fewer steps.  The fourth plot in Figure \ref{fig:similarity_corruption_levels_discretization} shows the runtime comparison for different numbers of discretization steps. The reported runtime is averaged over 100 images for the super resolution problem. While REPA introduces additional computation per step, the reduced number of required sampling steps allows the method to reach the same perceptual quality in less overall runtime. All runtime experiments were run on a single NVIDIA L40 GPU (48GB).

\noindent\textbf{\textsc{Repa} vs Guidance in the DINO feature space.}
To further showcase the benefits of representation alignment, Appendix~\ref{sec:feature_vs_repa}
compares \textsc{Repa} to a variant that enforces consistency directly in the DINOv2
feature space of the reconstructed image.
Concretely, the feature-only variant applies the following guidance at each timestep:
\begin{equation*}
z_{t-1} \gets z_{t-1}
+ \lambda \nabla_{z_t} \sum_{n=1}^{N}
\cos\!\Bigl(
c^{[n]}_{\text{proxy}},
f_{\mathrm{DINOv2}}^{[n]}\!\bigl(\mathcal{D}(\mathbb{E}[z_0 \mid z_t])\bigr)
\Bigr),
\end{equation*}
This approach relies solely on the DINOv2 features of the decoded image. In contrast, \textsc{Repa} aligns the model’s internal representations with DINOv2 features.
As observed in Appendix~\ref{sec:feature_vs_repa}, this feature-only guidance performs
worse in practice.
A natural explanation through the lens of Proposition \ref{prop:zspace_contraction_repa} is that the composite mapping
$z_t \mapsto \mathcal{D}(\mathbb{E}[z_0 \mid z_t]) \mapsto
f_{\mathrm{DINOv2}}(\cdot)$
is significantly more ill-conditioned than the lightweight projection used in
\textsc{Repa}, resulting in weaker and less stable gradients with respect to the diffusion
state.

\section{Conclusions}
\vspace{-0.2cm}

In this work, we introduced \emph{representation alignment} for solving inverse problems with diffusion- and flow-based generative models. By aligning internal model representations with features from a pretrained DINOv2 encoder, our method provides a strong semantic prior that guides the reconstruction process toward perceptually meaningful solutions. We further offered theoretical insights into this mechanism, showing that REPA regularization corresponds to minimizing a divergence in the DINOv2 feature space and reduces the discrepancy between the current and the clean latent state. These results elucidate why alignment can enhance perceptual fidelity in inverse problem settings. Overall, our study demonstrates that representation alignment constitutes an effective tool for improving generative inverse problem solvers, yielding benefits at inference time.

\bibliographystyle{plain}  
\bibliography{references}

@article{ho2020denoising,
  title={Denoising diffusion probabilistic models},
  author={Ho, Jonathan and Jain, Ajay and Abbeel, Pieter},
  journal={Advances in neural information processing systems},
  volume={33},
  pages={6840--6851},
  year={2020}
}

@article{pachebat2025iterative,
  title={Iterative Tilting for Diffusion Fine-Tuning},
  author={Pachebat, Jean and Conforti, Giovanni and Durmus, Alain and Janati, Yazid},
  journal={arXiv preprint arXiv:2512.03234},
  year={2025}
}

@article{fu2023dreamsim,
  title={Dreamsim: Learning new dimensions of human visual similarity using synthetic data},
  author={Fu, Stephanie and Tamir, Netanel and Sundaram, Shobhita and Chai, Lucy and Zhang, Richard and Dekel, Tali and Isola, Phillip},
  journal={arXiv preprint arXiv:2306.09344},
  year={2023}
}

@inproceedings{karras2019style,
  title={A style-based generator architecture for generative adversarial networks},
  author={Karras, Tero and Laine, Samuli and Aila, Timo},
  booktitle={Proceedings of the IEEE/CVF conference on computer vision and pattern recognition},
  pages={4401--4410},
  year={2019}
}

@article{lai2025principles,
  title={The principles of diffusion models},
  author={Lai, Chieh-Hsin and Song, Yang and Kim, Dongjun and Mitsufuji, Yuki and Ermon, Stefano},
  journal={arXiv preprint arXiv:2510.21890},
  year={2025}
}

@inproceedings{thaker2025frequency,
  title={Frequency-guided posterior sampling for diffusion-based image restoration},
  author={Thaker, Darshan and Goyal, Abhishek and Vidal, Ren{\'e}},
  booktitle={Proceedings of the IEEE/CVF International Conference on Computer Vision},
  pages={12873--12882},
  year={2025}
}

@inproceedings{raphaeli2025silo,
  title={Silo: Solving inverse problems with latent operators},
  author={Raphaeli, Ron and Man, Sean and Elad, Michael},
  booktitle={Proceedings of the IEEE/CVF International Conference on Computer Vision},
  pages={10570--10580},
  year={2025}
}

@inproceedings{sohl2015deep,
  title={Deep unsupervised learning using nonequilibrium thermodynamics},
  author={Sohl-Dickstein, Jascha and Weiss, Eric and Maheswaranathan, Niru and Ganguli, Surya},
  booktitle={International conference on machine learning},
  pages={2256--2265},
  year={2015},
  organization={PMLR}
}

@article{hyvarinen2005estimation,
  title={Estimation of non-normalized statistical models by score matching.},
  author={Hyv{\"a}rinen, Aapo and Dayan, Peter},
  journal={Journal of Machine Learning Research},
  volume={6},
  number={4},
  year={2005}
}

@inproceedings{rombach2022high,
  title={High-resolution image synthesis with latent diffusion models},
  author={Rombach, Robin and Blattmann, Andreas and Lorenz, Dominik and Esser, Patrick and Ommer, Bj{\"o}rn},
  booktitle={Proceedings of the IEEE/CVF conference on computer vision and pattern recognition},
  pages={10684--10695},
  year={2022}
}

@article{arbel2019maximum,
  title={Maximum mean discrepancy gradient flow},
  author={Arbel, Michael and Korba, Anna and Salim, Adil and Gretton, Arthur},
  journal={Advances in Neural Information Processing Systems},
  volume={32},
  year={2019}
}

@article{daras2024survey,
  title={A survey on diffusion models for inverse problems},
  author={Daras, Giannis and Chung, Hyungjin and Lai, Chieh-Hsin and Mitsufuji, Yuki and Ye, Jong Chul and Milanfar, Peyman and Dimakis, Alexandros G and Delbracio, Mauricio},
  journal={arXiv preprint arXiv:2410.00083},
  year={2024}
}

@inproceedings{
chung2023diffusion,
title={Diffusion Posterior Sampling for General Noisy Inverse Problems},
author={Hyungjin Chung and Jeongsol Kim and Michael Thompson Mccann and Marc Louis Klasky and Jong Chul Ye},
booktitle={The Eleventh International Conference on Learning Representations },
year={2023},
}

@inproceedings{
rout2023solving,
title={Solving Linear Inverse Problems Provably via Posterior Sampling with Latent Diffusion Models},
author={Litu Rout and Negin Raoof and Giannis Daras and Constantine Caramanis and Alex Dimakis and Sanjay Shakkottai},
booktitle={Thirty-seventh Conference on Neural Information Processing Systems},
year={2023},
}

@inproceedings{
song2024solving,
title={Solving Inverse Problems with Latent Diffusion Models via Hard Data Consistency},
author={Bowen Song and Soo Min Kwon and Zecheng Zhang and Xinyu Hu and Qing Qu and Liyue Shen},
booktitle={The Twelfth International Conference on Learning Representations},
year={2024},
}

@article{yu2024repa,
  title={Representation Alignment for Generation: Training Diffusion Transformers Is Easier Than You Think},
  author={Sihyun Yu and Sangkyung Kwak and Huiwon Jang and Jongheon Jeong and Jonathan Huang and Jinwoo Shin and Saining Xie},
  year={2024},
  journal={arXiv preprint arXiv:2410.06940},
}

@inproceedings{xiang2023,
  title={Denoising Diffusion Autoencoders are Unified Self-supervised Learners},
  author={Xiang, Weilai and Yang, Hongyu and Huang, Di and Wang, Yunhong},
  booktitle={Proceedings of the IEEE/CVF International Conference on Computer Vision},
  year={2023}
}

@inproceedings{li2023your,
  title={Your diffusion model is secretly a zero-shot classifier},
  author={Li, Alexander C and Prabhudesai, Mihir and Duggal, Shivam and Brown, Ellis and Pathak, Deepak},
  booktitle={Proceedings of the IEEE/CVF International Conference on Computer Vision},
  pages={2206--2217},
  year={2023}
}

@inproceedings{deng2009imagenet,
  title={Imagenet: A large-scale hierarchical image database},
  author={Deng, Jia and Dong, Wei and Socher, Richard and Li, Li-Jia and Li, Kai and Fei-Fei, Li},
  booktitle={2009 IEEE conference on computer vision and pattern recognition},
  pages={248--255},
  year={2009},
  organization={Ieee}
}

@article{oquab2023dinov2,
  title={Dinov2: Learning robust visual features without supervision},
  author={Oquab, Maxime and Darcet, Timoth{\'e}e and Moutakanni, Th{\'e}o and Vo, Huy and Szafraniec, Marc and Khalidov, Vasil and Fernandez, Pierre and Haziza, Daniel and Massa, Francisco and El-Nouby, Alaaeldin and others},
  journal={arXiv preprint arXiv:2304.07193},
  year={2023}
}

@inproceedings{zhang2018unreasonable,
  title={The unreasonable effectiveness of deep features as a perceptual metric},
  author={Zhang, Richard and Isola, Phillip and Efros, Alexei A and Shechtman, Eli and Wang, Oliver},
  booktitle={Proceedings of the IEEE conference on computer vision and pattern recognition},
  pages={586--595},
  year={2018}
}

@inproceedings{radford2021learning,
  title={Learning transferable visual models from natural language supervision},
  author={Radford, Alec and Kim, Jong Wook and Hallacy, Chris and Ramesh, Aditya and Goh, Gabriel and Agarwal, Sandhini and Sastry, Girish and Askell, Amanda and Mishkin, Pamela and Clark, Jack and others},
  booktitle={International conference on machine learning},
  pages={8748--8763},
  year={2021},
  organization={PMLR}
}

@article{dhariwal2021diffusion,
  title={Diffusion models beat gans on image synthesis},
  author={Dhariwal, Prafulla and Nichol, Alexander},
  journal={Advances in neural information processing systems},
  volume={34},
  pages={8780--8794},
  year={2021}
}

@inproceedings{
song2021denoising,
title={Denoising Diffusion Implicit Models},
author={Jiaming Song and Chenlin Meng and Stefano Ermon},
booktitle={International Conference on Learning Representations},
year={2021},
}

@article{wang2004image,
  title={Image quality assessment: from error visibility to structural similarity},
  author={Wang, Zhou and Bovik, Alan C and Sheikh, Hamid R and Simoncelli, Eero P},
  journal={IEEE transactions on image processing},
  volume={13},
  number={4},
  pages={600--612},
  year={2004},
  publisher={IEEE}
}

@article{heusel2017gans,
  title={Gans trained by a two time-scale update rule converge to a local nash equilibrium},
  author={Heusel, Martin and Ramsauer, Hubert and Unterthiner, Thomas and Nessler, Bernhard and Hochreiter, Sepp},
  journal={Advances in neural information processing systems},
  volume={30},
  year={2017}
}

@inproceedings{zhang2025improving,
  title={Improving diffusion inverse problem solving with decoupled noise annealing},
  author={Zhang, Bingliang and Chu, Wenda and Berner, Julius and Meng, Chenlin and Anandkumar, Anima and Song, Yang},
  booktitle={Proceedings of the Computer Vision and Pattern Recognition Conference},
  pages={20895--20905},
  year={2025}
}

@article{wang2025repa,
  title={REPA Works Until It Doesn't: Early-Stopped, Holistic Alignment Supercharges Diffusion Training},
  author={Wang, Ziqiao and Zhao, Wangbo and Zhou, Yuhao and Li, Zekai and Liang, Zhiyuan and Shi, Mingjia and Zhao, Xuanlei and Zhou, Pengfei and Zhang, Kaipeng and Wang, Zhangyang and others},
  journal={arXiv preprint arXiv:2505.16792},
  year={2025}
}

@article{wang2025learning,
  title={Learning Diffusion Models with Flexible Representation Guidance},
  author={Wang, Chenyu and Zhou, Cai and Gupta, Sharut and Lin, Zongyu and Jegelka, Stefanie and Bates, Stephen and Jaakkola, Tommi},
  journal={arXiv preprint arXiv:2507.08980},
  year={2025}
}

@article{tian2025u,
  title={U-repa: Aligning diffusion u-nets to vits},
  author={Tian, Yuchuan and Chen, Hanting and Zheng, Mengyu and Liang, Yuchen and Xu, Chao and Wang, Yunhe},
  journal={arXiv preprint arXiv:2503.18414},
  year={2025}
}

@inproceedings{yao2025reconstruction,
  title={Reconstruction vs. generation: Taming optimization dilemma in latent diffusion models},
  author={Yao, Jingfeng and Yang, Bin and Wang, Xinggang},
  booktitle={Proceedings of the Computer Vision and Pattern Recognition Conference},
  pages={15703--15712},
  year={2025}
}

@article{leng2025repa,
  title={Repa-e: Unlocking vae for end-to-end tuning with latent diffusion transformers},
  author={Leng, Xingjian and Singh, Jaskirat and Hou, Yunzhong and Xing, Zhenchang and Xie, Saining and Zheng, Liang},
  journal={arXiv preprint arXiv:2504.10483},
  year={2025}
}

@article{xu2025exploring,
  title={Exploring representation-aligned latent space for better generation},
  author={Xu, Wanghan and Yue, Xiaoyu and Wang, Zidong and Teng, Yao and Zhang, Wenlong and Liu, Xihui and Zhou, Luping and Ouyang, Wanli and Bai, Lei},
  journal={arXiv preprint arXiv:2502.00359},
  year={2025}
}

@inproceedings{assran2023self,
  title={Self-supervised learning from images with a joint-embedding predictive architecture},
  author={Assran, Mahmoud and Duval, Quentin and Misra, Ishan and Bojanowski, Piotr and Vincent, Pascal and Rabbat, Michael and LeCun, Yann and Ballas, Nicolas},
  booktitle={Proceedings of the IEEE/CVF Conference on Computer Vision and Pattern Recognition},
  pages={15619--15629},
  year={2023}
}

@article{simeoni2025dinov3,
  title={Dinov3},
  author={Sim{\'e}oni, Oriane and Vo, Huy V and Seitzer, Maximilian and Baldassarre, Federico and Oquab, Maxime and Jose, Cijo and Khalidov, Vasil and Szafraniec, Marc and Yi, Seungeun and Ramamonjisoa, Micha{\"e}l and others},
  journal={arXiv preprint arXiv:2508.10104},
  year={2025}
}

@article{li2024return,
  title={Return of unconditional generation: A self-supervised representation generation method},
  author={Li, Tianhong and Katabi, Dina and He, Kaiming},
  journal={Advances in Neural Information Processing Systems},
  volume={37},
  pages={125441--125468},
  year={2024}
}

@article{blattmann2022retrieval,
  title={Retrieval-augmented diffusion models},
  author={Blattmann, Andreas and Rombach, Robin and Oktay, Kaan and M{\"u}ller, Jonas and Ommer, Bj{\"o}rn},
  journal={Advances in Neural Information Processing Systems},
  volume={35},
  pages={15309--15324},
  year={2022}
}

@inproceedings{hu2023self,
  title={Self-guided diffusion models},
  author={Hu, Vincent Tao and Zhang, David W and Asano, Yuki M and Burghouts, Gertjan J and Snoek, Cees GM},
  booktitle={Proceedings of the IEEE/CVF Conference on Computer Vision and Pattern Recognition},
  pages={18413--18422},
  year={2023}
}

@article{sheynin2022knn,
  title={Knn-diffusion: Image generation via large-scale retrieval},
  author={Sheynin, Shelly and Ashual, Oron and Polyak, Adam and Singer, Uriel and Gafni, Oran and Nachmani, Eliya and Taigman, Yaniv},
  journal={arXiv preprint arXiv:2204.02849},
  year={2022}
}

@inproceedings{lipman2023flow,
  title={Flow Matching for Generative Modeling},
  author={Lipman, Yaron and Chen, Ricky TQ and Ben-Hamu, Heli and Nickel, Maximilian and Le, Matt},
  booktitle={11th International Conference on Learning Representations, ICLR 2023},
  year={2023}
}

@article{patel2024steering,
  title={Steering rectified flow models in the vector field for controlled image generation},
  author={Patel, Maitreya and Wen, Song and Metaxas, Dimitris N and Yang, Yezhou},
  journal={arXiv preprint arXiv:2412.00100},
  year={2024}
}

@article{albergo2023stochastic,
  title={Stochastic interpolants: A unifying framework for flows and diffusions},
  author={Albergo, Michael S and Boffi, Nicholas M and Vanden-Eijnden, Eric},
  journal={arXiv preprint arXiv:2303.08797},
  year={2023}
}

@inproceedings{liu2023flow,
  title={Flow Straight and Fast: Learning to Generate and Transfer Data with Rectified Flow},
  author={Liu, Xingchao and Gong, Chengyue and Liu, Qiang},
  booktitle={The Eleventh International Conference on Learning Representations (ICLR)},
  year={2023}
}

@article{dosovitskiy2016generating,
  title={Generating Images with Perceptual Similarity Metrics based on Deep Networks},
  author={Dosovitskiy, Alexey and Brox, Thomas},
  journal={Advances in Neural Information Processing Systems},
  volume={29},
  year={2016}
}








\appendix

\section{Appendix}

\subsection{Proofs}\label{sec:proofs}

\begin{proposition}
\label{lem:repa_mmd_mine_app}
Assume $x \sim p_X$ is a ground-truth image and $\hat{x} \sim p_{\hat{X}\mid Y}$ is its
corresponding reconstruction, where $p_{\hat{X}\mid Y}$ is the distribution induced
by any reconstruction method. Assume further that all DINOv2 feature embeddings are
$\ell_2$--normalized. Let $\bar{x}$ be a proxy approximation of $x$
Define the empirical mean DINOv2 feature embedding as
\[
\mu_f(x)
:=
\mathbb{E}_{n}\!\left[f_{\mathrm{DINOv2}}^{[n]}(x)\right].
\]

Then the expected REPA alignment satisfies
\begin{equation*}
\begin{aligned}
&\mathbb{E}_{\hat{x},\bar{x}}
\!\left[\text{REPA}(\bar{x},\hat{x})\right]
\;\le\;
1
-\frac{1}{8}\,
\textsc{MMD}_{\mathrm{DINOv2}}\!\left(p_X,\,p_{\hat X}\right)
\\
&\quad
+\frac{1}{2}\,
\mathbb{E}_{x,\bar x}\!\big[\mathrm{ApproxErr}(x,\bar{x})\big]
+\frac{1}{4}\,
\mathbb{E}_{\hat x}\!\big[\mathrm{MisREPA}(\hat{x})\big].
\end{aligned}
\label{eq:bound_mmd_approx_clean}
\end{equation*}
where
\[
\textsc{MMD}_{\mathrm{DINOv2}}\!\left(p_X,\,p_{\hat X}\right)
=
\bigl\|
    \mathbb{E}_{x}[\mu_f(x)]
    -
    \mathbb{E}_{\hat x}[\mu_f(\hat{x})]
\bigr\|_2^2
\]
is the maximum mean discrepancy between $p_X$ and $p_{\hat X}$ in the DINOv2 feature space.
\end{proposition}
\begin{proof}
\[
\text{REPA}(\bar{x},\hat{x})
=
\frac{1}{N}\sum_{n=1}^N 
\cos\!\left(
f_{\text{DINOv2}}^{[n]}(\bar{x}),\,
g_\phi\!\big(\textsc{DiffEnc}^{[n]}(\hat{x})\big)
\right),
\]
Assume the features are $\ell_2$–normalized. 
Then for each $n$,
\[
\cos\!\left(
f_{\text{DINOv2}}^{[n]}(\bar{x}),\,
g_\phi\!\big(\textsc{DiffEnc}^{[n]}(\hat{x})\big)
\right)
= 
1 - \tfrac{1}{2}\,\big\|
f_{\text{DINOv2}}^{[n]}(\bar{x})
- 
g_\phi(\textsc{DiffEnc}^{[n]}(\hat{x}))
\big\|_2^2,
\]
and hence
\begin{align*}
\text{REPA}(\bar{x},\hat{x})
&= 
1 
\;-\; 
\frac{1}{2N}\sum_{n=1}^N
\big\|
f_{\text{DINOv2}}^{[n]}(\bar{x})
- 
g_\phi(\textsc{DiffEnc}^{[n]}(\hat{x}))
\big\|_2^2.
\end{align*}

Let the empirical (mean) feature embeddings be
\begin{align*}
\mu_f(x)      &:= \frac{1}{N}\sum_{n=1}^N f_{\text{DINOv2}}^{[n]}(x),
& \mu_f(\bar x) &:= \frac{1}{N}\sum_{n=1}^N f_{\text{DINOv2}}^{[n]}(\bar x), \\[4pt]
\mu_f(\hat x) &:= \frac{1}{N}\sum_{n=1}^N f_{\text{DINOv2}}^{[n]}(\hat x),
& \mu_g(\hat x) &:= \frac{1}{N}\sum_{n=1}^N g_\phi(\textsc{DiffEnc}^{[n]}(\hat x)).
\end{align*}
From Jensen's inequality for the squared norm : 
\[
\big\|\mu_f(x)-\mu_f(\hat{x})\big\|_2^2
\le
2\,\big\|\mu_f(x)-\mu_g(\hat{x})\big\|_2^2
+2\,\big\|\mu_f(\hat{x})-\mu_g(\hat{x})\big\|_2^2.
\]
Moreover,
\[
\big\|\mu_f(x)-\mu_g(\hat{x})\big\|_2^2
=\Big\|\frac{1}{N}\sum_{n=1}^N\!\big(f_{\text{DINOv2}}^{[n]}(x)-g_\phi(\textsc{DiffEnc}^{[n]}(\hat{x}))\big)\Big\|_2^2
\;\le\;
\frac{1}{N}\sum_{n=1}^N\!\big\|f_{\text{DINOv2}}^{[n]}(x)-g_\phi(\textsc{DiffEnc}^{[n]}(\hat{x}))\big\|_2^2,
\]
and similarly
\[
\big\|\mu_f(\hat{x})-\mu_g(\hat{x})\big\|_2^2
\;\le\;
\frac{1}{N}\sum_{n=1}^N\!\big\|f_{\text{DINOv2}}^{[n]}(\hat{x})-g_\phi(\textsc{DiffEnc}^{[n]}(\hat{x}))\big\|_2^2
\;= \text{MisREPA}(\hat{x}).
\]
Combining the two yields
\begin{align}
&\big\|\mu_f(x)-\mu_f(\hat{x})\big\|_2^2
\le
\frac{2}{N}\sum_{n=1}^N\!\big\|f_{\text{DINOv2}}^{[n]}(x)
    - g_\phi(\textsc{DiffEnc}^{[n]}(\hat{x}))\big\|_2^2
+ 2 \,\text{MisREPA}(\hat{x}) \nonumber\\[6pt]
&\le
\frac{4}{N}\sum_{n=1}^N\!\big\|f_{\text{DINOv2}}^{[n]}(\bar{x})
    - g_\phi(\textsc{DiffEnc}^{[n]}(\hat{x}))\big\|_2^2
+ \frac{4}{N}\sum_{n=1}^N\!\big\|f_{\text{DINOv2}}^{[n]}(x)
    - f_{\text{DINOv2}}^{[n]}(\bar{x})\big\|_2^2
+ 2\,\text{MisREPA}(\hat{x}). \\
& = \frac{4}{N}\sum_{n=1}^N\!\big\|f_{\text{DINOv2}}^{[n]}(\bar{x})
    - g_\phi(\textsc{DiffEnc}^{[n]}(\hat{x}))\big\|_2^2
+  4\text{ApproxErr}(x,\bar{x})
+ 2\text{MisREPA}(\hat{x}).
\label{eq:mean-bound}
\end{align}

Using the cosine–squared identity for unit–norm features,
\[
\sum_{n=1}^N
\big\|f_{\text{DINOv2}}^{[n]}(\bar{x})
- g_\phi(\textsc{DiffEnc}^{[n]}(\hat{x}))\big\|_2^2
= 2N\big(1-\text{REPA}(\bar{x},\hat{x})\big),
\]
\[
\big\|\mu_f(x)-\mu_f(\hat{x})\big\|_2^2
\;\le\;
8\Big(1-\text{REPA}(\bar{x},\hat{x})\Big)
\;+\; 4\text{ApproxErr}(x,\bar{x})
\;+\;
2\,\text{MisREPA}(\hat{x}).
\]

Rearranging gives the bound
\[
\text{REPA}(\bar{x},\hat{x})
\;\le\;
1 
-\frac{1}{8}\,\big\|\mu_f(x)-\mu_f(\hat{x})\big\|_2^2
+\frac{1}{2}\,\text{ApproxErr}(x,\bar{x})
+\frac{1}{4}\,\text{MisREPA}(\hat{x}).
\]

Taking expectations over \(x\sim p_X\), and \(\hat{x}\sim p_{\hat{X}}\), and using the convexity of the squared norm,
\[
\mathbb{E}\!\big[\|\mu_f(x)-\mu_f(\hat{x})\|_2^2\big]
\;\ge\;
\big\|\mathbb{E}_{x}[\mu_f(x)]
-\mathbb{E}_{\hat{x}}[\mu_f(\hat{x})]\big\|_2^2
= \textsc{MMD}_{\text{DINOv2}}(p_X,p_{\hat X}),
\]
we obtain
\begin{equation}
\mathbb{E}_{\hat{x}, \bar{x}}\!\left[\text{REPA}(\bar{x},\hat{x})\right]
\;\le\;
1 
- \frac{1}{8}\,\textsc{MMD}_{\textsc{DINOv2}}\!\left(p_X,\,p_{\hat{X}}\right)
+ \frac{1}{2}\,\mathbb{E}_{x, \bar x}\!\big[\text{ApproxErr}(x,\bar{x})\big]
+ \frac{1}{4}\,\mathbb{E}_{\hat{x}}\!\big[\text{MisREPA}(\hat{x})\big],
\label{eq:bound_mmd_approx}
\end{equation}
\end{proof}

\begin{assumption}[Regularity of the diffusion encoder and REPA projection]
\label{assump:repa}
We operate under the following regularity and conditioning assumptions on the diffusion
feature extractor and the REPA projection. For notational simplicity in the analysis, we
introduce an operator corresponding to the vectorized diffusion encoder. Let
$
\widetilde G(\cdot)
:= \mathrm{vec}(\textsc{DiffEnc}(\cdot,t))
:\mathbb{R}^d \to \mathbb{R}^{ND}
$
denote the vectorized diffusion feature extractor at timestep \(t\), and let
$
\widetilde g_{\phi} : \mathbb{R}^{ND} \to \mathbb{R}^{ND}
$
denote the patchwise lift of the projection head
\(g_{\phi} : \mathbb{R}^D \to \mathbb{R}^D\).
Fix two diffusion states \(z_t, z^* \in \mathbb{R}^d\) and define
\(u_t := \widetilde G(z_t)\) and \(u^* := \widetilde G(z^*)\).
Let
\[
B := \{ z^* + s (z_t - z^*) : s \in [0,1] \}
\quad\text{and}\quad
\mathcal U := \{ \widetilde G(z) : z \in B \}.
\]

\begin{enumerate}[label=\textbf{A.2.\arabic*}, leftmargin=2.6em]

\item
The mapping \(\widetilde G\) is differentiable, and its Jacobian
\(J_{\widetilde G}(z) \in \mathbb{R}^{(ND)\times d}\) is Lipschitz continuous on \(B\)
with Lipschitz constant \(L_G\).

\item
There exist constants \(m_G > 0\) and \(M_G < \infty\) such that
\[
J_{\widetilde G}(z_t)^\top J_{\widetilde G}(z_t) \succeq m_G^2 I_d,
\qquad
\| J_{\widetilde G}(z) \|_2 \le M_G
\quad \text{for all } z \in B.
\]

\item
The mapping \(\widetilde g_{\phi}\) is differentiable, and its Jacobian
\(J_{\widetilde g_{\phi}}(u) \in \mathbb{R}^{(ND)\times (ND)}\) is uniformly
well-conditioned at \(u_t\): there exists a constant \(m_\phi > 0\) such that
\[
J_{\widetilde g_{\phi}}(u_t)^\top J_{\widetilde g_{\phi}}(u_t)
\succeq m_\phi^2 I_{ND},
\qquad
\| J_{\widetilde g_{\phi}}(u) \|_2 \le M_\phi
\quad \text{for all } u \in \mathcal U.
\]

\item
The Jacobian \(J_{\widetilde g_{\phi}}\) is Lipschitz continuous on \(\mathcal U\) with
Lipschitz constant \(L_\phi\).

\end{enumerate}
\end{assumption}

\begin{proposition}[Local contraction of REPA in the diffusion state space]
\label{prop:zspace_contraction_repa_app_nonlinear}
Consider an inverse problem in which an observation $y\in\mathbb{R}^m$ is generated
from a ground-truth image $x\sim p_X$, and let $\bar x\sim p_{\bar X}$ denote the proxy
approximation used during REPA.
Let $z_t\in\mathbb{R}^d$ be the diffusion state at timestep $t$ produced by
Algorithm~\ref{alg:algorithm}, and let $z_t^{(\mathrm{REPA})}$ denote the diffusion state
after applying the REPA regularization update.
Let $z^*\in\mathbb{R}^d$ denote the latent variable associated with the clean image $x$. Under Assumptions~\textbf{\(A_1\)}–\textbf{\(A_4\)}, define
\[
r
:=
\frac{m^2_G\,m^2_\phi}{
L_{\phi}M^3_GM_\phi
+
L_GM^2_\phi M_G.
}
\]
If $\|z_t - z^*\|_2 < r$, then there exists a stepsize
\[
0 < \lambda \le \frac{1}{2M_G^2\,M_\phi^2}
\]
such that the REPA update induces a local contraction in the diffusion state space. In
particular,
\[
\|z_t^{(\mathrm{REPA})} - z^*\|_2
\;\le\;
C_1\,\|z_t - z^*\|_2
+
C_2\left(
\sqrt{\mathrm{ApproxErr}(x,\bar x)}
+
\sqrt{\mathrm{MisREPA}(x)}
\right),
\]
where
\[
C_1 := 1 - \lambda m^2_G m^2_\phi \in (0,1),
\qquad
C_2 := 2\,\lambda\,M_G\,M_\phi\sqrt{N}.
\]
\end{proposition}

\begin{proof}
We first rewrite all quantities in vectorized form. Let $\mathrm{vec}(\cdot)$ denote the
operator that stacks the rows of a matrix into a single vector. For any matrix
$H\in\mathbb{R}^{k_1\times k_2}$, define
\[
\widetilde H := \mathrm{vec}(H)\in\mathbb{R}^{k_1k_2}.
\]

For the DINOv2 features $f_{\mathrm{DINOv2}}(x)\in\mathbb{R}^{N\times D}$ we write
\[
\widetilde f(x):=\mathrm{vec}\!\bigl(f_{\mathrm{DINOv2}}(x)\bigr)\in\mathbb{R}^{ND}.
\]

For the diffusion-model feature extractor
\(\textsc{DiffEnc} : \mathbb{R}^d \times \mathbb{R} \to \mathbb{R}^{N \times D}\),
we define its vectorized form by
\[
\widetilde G(z_t)
:= \mathrm{vec}\!\bigl(\textsc{DiffEnc}(z_t,t)\bigr)
\in \mathbb{R}^{ND}.
\]
Recall that $\phi:\mathbb{R}^{D}\to\mathbb{R}^{D}$ is differentiable. Its lifted Jacobian
acts patchwise and satisfies
\[
J_{\widetilde g_{\phi}}(\widetilde G(z_t))
=
I_N \otimes J_\phi(G_2(z_t)).
\]
The REPA update is given by
\[
z_t^{(\mathrm{REPA})}
=
z_t
-
\lambda\nabla_{z_t}
\bigl\|
\widetilde f(\bar x)
-
\widetilde g_{\phi}(\widetilde G(z_t))
\bigr\|_2^2.
\]
By the chain rule,
\[
z_t^{(\mathrm{REPA})}
=
z_t
+
2\lambda\,J_{\widetilde G}(z_t)^\top
J_{\widetilde g_{\phi}}(\widetilde G(z_t))^\top
\bigl(
\widetilde f(\bar x)
-
\widetilde g_{\phi}(\widetilde G(z_t))
\bigr).
\]
Subtracting $z^*$ from both sides yields
\[
z_t^{(\mathrm{REPA})} - z^*
=
z_t - z^*
+
2\lambda\,J_{\widetilde G}(z_t)^\top
J_{\widetilde g_{\phi}}(\widetilde G(z_t))^\top
\bigl(
\widetilde f(\bar x)
-
\widetilde g_{\phi}(\widetilde G(z_t))
\bigr).
\]
We now add and subtract $\widetilde f(x)$ and $\widetilde g_{\phi}(\widetilde G(z^*))$ inside the parentheses:
\begin{align}
z_t^{(\mathrm{REPA})}-z^*
&=
(z_t-z^*)
-
2\lambda\,J_{\widetilde G}(z_t)^\top
J_{\widetilde g_{\phi}}(\widetilde G(z_t))^\top
\bigl(
\widetilde g_{\phi}(\widetilde G(z_t))
-
\widetilde g_{\phi}(\widetilde G(z^*))
\bigr)
\nonumber\\
&\quad
+
2\lambda\,J_{\widetilde G}(z_t)^\top
J_{\widetilde g_{\phi}}(\widetilde G(z_t))^\top
\bigl(
\widetilde f(\bar x)-\widetilde f(x)
\bigr)
\nonumber\\
&\quad
+
2\lambda\,J_{\widetilde G}(z_t)^\top
J_{\widetilde g_{\phi}}(\widetilde G(z_t))^\top
\bigl(
\widetilde f(x)-\widetilde g_{\phi}(\widetilde G(z^*))
\bigr).
\label{eq:z_split_three_terms_nonlinear_phi}
\end{align}

We define the \emph{contractive part} as
\[
\mathcal{T}_t
:=
(z_t-z^*)
-
2\lambda\,J_{\widetilde G}(z_t)^\top
J_{\widetilde g_{\phi}}(\widetilde G(z_t))^\top
\bigl(
\widetilde g_{\phi}(\widetilde G(z_t))
-
\widetilde g_{\phi}(\widetilde G(z^*))
\bigr).
\]
Applying the fundamental theorem of calculus to
$u\mapsto \widetilde g_{\phi}(u)$ along
$u(s)=\widetilde G(z(s))$ with $z(s)=z^*+s(z_t-z^*)$, $s\in[0,1]$, yields
\[
\widetilde g_{\phi}(\widetilde G(z_t))
-
\widetilde g_{\phi}(\widetilde G(z^*))
=
\int_0^1
J_{\widetilde g_{\phi}}\!\bigl(\widetilde G(z(s))\bigr)\,
J_{\widetilde G}\!\bigl(z(s)\bigr)\,(z_t-z^*)\,ds.
\]
Substituting into the definition of $\mathcal{T}_t$ gives
\begin{align}
\mathcal{T}_t
&=
\Bigl[
I
-
2\lambda
\int_0^1
J_{\widetilde G}(z_t)^\top
J_{\widetilde g_{\phi}}\!\bigl(\widetilde G(z_t)\bigr)^\top
J_{\widetilde g_{\phi}}\!\bigl(\widetilde G(z(s))\bigr)
J_{\widetilde G}\!\bigl(z(s)\bigr)
\,ds
\Bigr]
(z_t - z^*).
\label{eq:Tt_final_clean_nonlinear_phi}
\end{align}
Add and subtract
$J_{\widetilde g_{\phi}}(\widetilde G(z_t))\,J_{\widetilde G}(z_t)$
inside the integral to obtain
\begin{align}
\mathcal{T}_t
&=
\Bigl[
I
-
2\lambda
\,J_{\widetilde G}(z_t)^\top
J_{\widetilde g_{\phi}}(\widetilde G(z_t))^\top
J_{\widetilde g_{\phi}}(\widetilde G(z_t))
J_{\widetilde G}(z_t)
\Bigr]
(z_t-z^*)
\nonumber\\
&\quad
-
2\lambda
\Bigl[
\int_0^1
J_{\widetilde G}(z_t)^\top
J_{\widetilde g_{\phi}}(\widetilde G(z_t))^\top
\Bigl(
J_{\widetilde g_{\phi}}(\widetilde G(z(s)))\,J_{\widetilde G}(z(s))
-
J_{\widetilde g_{\phi}}(\widetilde G(z_t))\,J_{\widetilde G}(z_t)
\Bigr)\,ds
\Bigr]
(z_t-z^*).
\label{eq:Tt_two_term_decomp_nonlinear_phi}
\end{align}
Taking Euclidean norms in \eqref{eq:Tt_two_term_decomp_nonlinear_phi} and using the
triangle inequality gives
\begin{align}
\|\mathcal{T}_t\|_2
&\le
\Bigl\|
I
-
2\lambda
\,J_{\widetilde G}(z_t)^\top
J_{\widetilde g_{\phi}}(\widetilde G(z_t))^\top
J_{\widetilde g_{\phi}}(\widetilde G(z_t))
J_{\widetilde G}(z_t)
\Bigr\|_2\;
\|z_t-z^*\|_2
\nonumber\\
&\quad
+
2\lambda
\Bigl\|
\int_0^1
J_{\widetilde G}(z_t)^\top
J_{\widetilde g_{\phi}}(\widetilde G(z_t))^\top
\Bigl(
J_{\widetilde g_{\phi}}(\widetilde G(z(s)))\,J_{\widetilde G}(z(s))
-
J_{\widetilde g_{\phi}}(\widetilde G(z_t))\,J_{\widetilde G}(z_t)
\Bigr)\,ds
\Bigr\|_2\;
\|z_t-z^*\|_2 .
\label{eq:Tt_two_term_norm_nonlinear_phi}
\end{align}
Moreover, by submultiplicativity of the spectral norm and
$\bigl\|\int_0^1 A(s)\,ds\bigr\|_2 \le \int_0^1 \|A(s)\|_2\,ds$, the second term can be
bounded as
\begin{align}
\Bigl\|
&\int_0^1
J_{\widetilde G}(z_t)^\top
J_{\widetilde g_{\phi}}(\widetilde G(z_t))^\top
\Bigl(
J_{\widetilde g_{\phi}}(\widetilde G(z(s)))\,J_{\widetilde G}(z(s))
-
J_{\widetilde g_{\phi}}(\widetilde G(z_t))\,J_{\widetilde G}(z_t)
\Bigr)\,ds
\Bigr\|_2 \leq \\
&
\int_0^1
\|J_{\widetilde G}(z_t)\|_2\;
\|J_{\widetilde g_{\phi}}(\widetilde G(z_t))\|_2
\Bigl\|
J_{\widetilde g_{\phi}}(\widetilde G(z(s)))\,J_{\widetilde G}(z(s))
-
J_{\widetilde g_{\phi}}(\widetilde G(z_t))\,J_{\widetilde G}(z_t)
\Bigr\|_2
\,ds .
\label{eq:integral_norm_bound_nonlinear_phi}
\end{align}
Combining \eqref{eq:Tt_two_term_norm_nonlinear_phi}--\eqref{eq:integral_norm_bound_nonlinear_phi}
yields
\begin{align}
&\|\mathcal{T}_t\|_2
\le
\Bigl\|
I
-
2\lambda
\,J_{\widetilde G}(z_t)^\top
J_{\widetilde g_{\phi}}(\widetilde G(z_t))^\top
J_{\widetilde g_{\phi}}(\widetilde G(z_t))
J_{\widetilde G}(z_t)
\Bigr\|_2\;
\|z_t-z^*\|_2
\nonumber\\
&\quad
+
2\lambda
\Bigl(
\int_0^1
\|J_{\widetilde G}(z_t)\|_2\;
\|J_{\widetilde g_{\phi}}(\widetilde G(z_t))\|_2
\Bigl\|
J_{\widetilde g_{\phi}}(\widetilde G(z(s)))\,J_{\widetilde G}(z(s))
-
J_{\widetilde g_{\phi}}(\widetilde G(z_t))\,J_{\widetilde G}(z_t)
\Bigr\|_2
\,ds
\Bigr)
\|z_t-z^*\|_2 .
\label{eq:Tt_final_norm_bound_nonlinear_phi}
\end{align}
We bound the difference inside the integral by writing
\begin{align}
&\Bigl\|
J_{\widetilde g_{\phi}}(\widetilde G(z(s)))\,J_{\widetilde G}(z(s))
-
J_{\widetilde g_{\phi}}(\widetilde G(z_t))\,J_{\widetilde G}(z_t)
\Bigr\|_2
\nonumber\\
&\le
\Bigl\|
\bigl(
J_{\widetilde g_{\phi}}(\widetilde G(z(s)))
-
J_{\widetilde g_{\phi}}(\widetilde G(z_t))
\bigr)
J_{\widetilde G}(z(s))
\Bigr\|_2
+
\Bigl\|
J_{\widetilde g_{\phi}}(\widetilde G(z_t))
\bigl(
J_{\widetilde G}(z(s))
-
J_{\widetilde G}(z_t)
\bigr)
\Bigr\|_2
\nonumber\\
&\le
\|J_{\widetilde g_{\phi}}(\widetilde G(z(s)))
-
J_{\widetilde g_{\phi}}(\widetilde G(z_t))\|_2\;
\|J_{\widetilde G}(z(s))\|_2
+
\|J_{\widetilde g_{\phi}}(\widetilde G(z_t))\|_2\;
\|J_{\widetilde G}(z(s))
-
J_{\widetilde G}(z_t)\|_2 .
\end{align}
Using the Lipschitz property of $J_{\widetilde G}$,
\[
\|J_{\widetilde G}(z(s))-J_{\widetilde G}(z_t)\|_2
\le
L_G\,\|z(s)-z_t\|_2
=
L_G(1-s)\,\|z_t-z^*\|_2,
\]
together with the Lipschitz continuity of $J_{\widetilde g_{\phi}}$,
\[
\|J_{\widetilde g_{\phi}}(\widetilde G(z(s)))
-
J_{\widetilde g_{\phi}}(\widetilde G(z_t))\|_2
\le
L_{\phi}M_G(1-s)\,\|z_t-z^*\|_2,
\]
and the uniform bounds
\[
\|J_{\widetilde G}(z(s))\|_2 \le M_G,
\qquad
\|J_{\widetilde g_{\phi}}(\widetilde G(z_t))\|_2 \le M_\phi,
\]
This yields
\[
\Bigl\|
J_{\widetilde g_{\phi}}(\widetilde G(z(s)))\,J_{\widetilde G}(z(s))
-
J_{\widetilde g_{\phi}}(\widetilde G(z_t))\,J_{\widetilde G}(z_t)
\Bigr\|_2
\le
(1-s)
\Bigl(
L_{\phi}M^2_G
+
L_GM_\phi
\Bigr)
\|z_t-z^*\|_2.
\]
Substituting this bound into \eqref{eq:Tt_final_norm_bound_nonlinear_phi} and
integrating over $s\in[0,1]$ yields
\begin{align}
\|\mathcal{T}_t\|_2
&\le
\Bigl\|
I
-
2\lambda
\,J_{\widetilde G}(z_t)^\top
J_{\widetilde g_{\phi}}(\widetilde G(z_t))^\top
J_{\widetilde g_{\phi}}(\widetilde G(z_t))
J_{\widetilde G}(z_t)
\Bigr\|_2\;
\|z_t-z^*\|_2
\nonumber\\
&\quad
+
\lambda
\Bigl(
L_{\phi}M^3_GM_\phi
+
L_GM^2_\phi M_G
\Bigr)
\|z_t-z^*\|_2^2 .
\label{eq:Tt_norm_bound_local_nonlinear_phi}
\end{align}
Since the symmetric matrix
\[
S_t
:=
J_{\widetilde G}(z_t)^\top
J_{\widetilde g_{\phi}}(\widetilde G(z_t))^\top
J_{\widetilde g_{\phi}}(\widetilde G(z_t))
J_{\widetilde G}(z_t)
\]
has eigenvalues contained in $[m^2_G m^2_\phi,\, M^2_G M^2_\phi]$, its spectral norm satisfies
\[
\bigl\| I - 2\lambda S_t \bigr\|_2
=
\max_{\sigma\in [m^2_G m^2_\phi,\, M^2_G M^2_\phi]} |1-2\lambda\sigma|.
\]
For any $0<\lambda\le 1/(2M^2_G M^2_\phi)$, the maximum is attained at $\sigma=m^2_G m^2_\phi$, and hence
\[
\bigl\| I - 2\lambda S_t \bigr\|_2
\le
1 - 2\lambda m^2_G m^2_\phi.
\]
Substituting this bound into \eqref{eq:Tt_norm_bound_local_nonlinear_phi} yields
\begin{align}
\|\mathcal{T}_t\|_2
&\le
\Bigl(
1-2\lambda m^2_G m^2_\phi
+
\lambda
\Bigl(
L_{\phi}M^3_GM_\phi
+
L_GM^2_\phi M_G
\Bigr)
\|z_t-z^*\|_2
\Bigr)
\|z_t-z^*\|_2 .
\label{eq:Tt_contraction_factor_nonlinear_phi}
\end{align}
Therefore, if
\[
\|z_t-z^*\|_2
<
\frac{m^2_G\,m^2_\phi}{
L_{\phi}M^3_GM_\phi
+
L_GM^2_\phi M_G
\
}
\qquad\text{and}\qquad
0<\lambda\le \frac{1}{2M_G^2 M_\phi^2},
\]
then the contractive part satisfies the strict contraction
\[
\|\mathcal{T}_t\|_2
\le
\bigl(1-\lambda m_G^2 m_\phi^2\bigr)\,\|z_t-z^*\|_2,
\]
with contraction factor strictly smaller than one.
By the definitions of the approximation and misalignment errors,
\[
\|\widetilde f(\bar x)-\widetilde f(x)\|_2
=
\sqrt{N}\,\sqrt{\mathrm{ApproxErr}(x,\bar x)},
\qquad
\|\widetilde f(x)-\widetilde g_{\phi}\,(\widetilde G(z^*))\|_2
=
\sqrt{N}\,\sqrt{\mathrm{MisREPA}(x)}.
\]
Returning to the decomposition we combine the
contraction bound for $\mathcal{T}_t$ with the remaining two terms. Using the triangle
inequality together with the uniform bounds
$\|J_{\widetilde G}(z_t)\|_2\le M_G$ and
$\|J_{\widetilde g_{\phi}}(\widetilde G(z_t))\|_2 \le M_\phi$, we obtain
\begin{align}
\|z_t^{(\mathrm{repa})}-z^*\|_2
&\le
\|\mathcal{T}_t\|_2
+
2\lambda M_GM_\phi\,
\|\widetilde f(\bar x)-\widetilde f(x)\|_2
\nonumber\\
&\quad
+
2\lambda M_GM_\phi\,
\|\widetilde f(x)-\widetilde g_{\phi}(\,\widetilde G(z^*))\|_2 .
\label{eq:z_full_bound}
\end{align}
Substituting the contraction estimate together with the above identities yields
\begin{align}
\|z_t^{(\mathrm{REPA})}-z^*\|_2
&\le
(1-\lambda m^2_Gm^2_{\phi})\,\|z_t-z^*\|_2
+
C
\Bigl(
\sqrt{\mathrm{ApproxErr}(x,\bar x)}
+
\sqrt{\mathrm{MisREPA}(x)}
\Bigr),
\end{align}
where the constant $C$ is defined explicitly as
\[
C := 2\lambda M_GM_\phi \sqrt{N}.
\]
This concludes the proof.
\end{proof}

\subsection{Details About Latent DPS + \textsc{Repa} Implementation}
\label{sec:latent_dps}

We implement Latent DPS + \textsc{Repa} following Algorithm~\ref{alg:latent_dps_repa} and use
1000 sampling steps throughout.
We employ an SNR-based learning rate schedule of the form
\[
\eta(t) = \frac{\kappa}{\max\!\left(\tfrac{t}{1 - t},\, 1 \right)},
\]
where $\kappa$ is a tunable scaling factor.
This schedule accounts for the varying signal-to-noise ratio across diffusion timesteps and
provides stable performance in practice.
The parameters $\kappa$ and the representation-alignment strength $\lambda$ are tuned on a
small validation set with present the best parameters found in Table \ref{table:best_hparams}.

\begin{table}[H]
\centering
\small
\caption{
Hyperparameters used for Latent DPS + \textsc{Repa}.
}
\begin{tabular}{lcccc}
\toprule
& \multicolumn{2}{c}{\textbf{ImageNet}} & \multicolumn{2}{c}{\textbf{FFHQ}} \\
\cmidrule(lr){2-3} \cmidrule(lr){4-5}
\textbf{Task}
& $\boldsymbol{\kappa}$ & $\boldsymbol{\lambda}$
& $\boldsymbol{\kappa}$ & $\boldsymbol{\lambda}$ \\
\midrule
Super-resolution      & 2.0  & 0.01 & \emph{1.25} & \emph{0.025} \\
Gaussian deblurring   & 0.25 & 0.05 & \emph{0.5}  & \emph{0.05} \\
Motion deblurring     & 0.5  & 0.01 & \emph{0.5}  & \emph{0.05} \\
Box inpainting        & 0.5  & 0.01 & \emph{0.75} & \emph{0.025} \\
\bottomrule
\end{tabular}
\label{table:best_hparams}
\end{table}

\begin{algorithm*}[t]
\caption{Latent DPS + REPA Algorithm}
\label{alg:latent_dps_repa}
\begin{algorithmic}[1]
\Require flow model $u_\theta$, measurement $y$, pretrained encoder $f$, proxy representation $c_{\text{proxy}}$, timestep $t_\text{cutoff}$, initialize $z_T \sim \mathcal{N}(0, I)$
\For{$t \in \{T, \dots, 0\}$}
    \State $v \gets u_\theta(z_t, t)$
    \State $\hat{z}_0 \gets \mathbb{E}[z_0 \mid z_t]$
    \State $z_{t-1} \gets z_t - \frac{1}{T} \cdot v$
    \State $z_{t-1} \gets z_{t-1} - \eta \nabla_{z_t} \|y - \mathcal{A}(\mathcal{D}(\hat{z}_0))\|_2^2$
    \State \textcolor{blue}{$z_{t-1} \gets z_{t-1} + \lambda \nabla_{z_t} \sum_{n=1}^{N} \cos(c_{\text{proxy}}^{[n]}, g_{\phi}(\mathrm{DiffEnc}^{[n]}(z_t,t))$}
    \If{$t < t_{\text{cutoff}}$}
        \State $c_{\text{proxy}} \gets
        f_{\mathrm{DINOv2}}\!\bigl(
        \mathcal{D}(\mathbb{E}[z_0 \mid z_t])
        \bigr)$
    \EndIf
\EndFor
\State \Return $x_0$
\end{algorithmic}
\end{algorithm*}

\subsection{Details About Resample + REPA Implementation}\label{sec:resample}

In Table~\ref{alg:latent_resample}, we present an adaptation of the algorithm proposed in \cite{song2024solving} to the flow-based setting, augmented with the REPA regularizer as described in the methodology section. We note that the original ReSample algorithm employs a three-stage procedure for enforcing data consistency: (i) gradient steps in latent space, similar to Latent DPS; (ii) pixel-space optimization, which is computationally efficient and captures high-level semantics but often leads to blurrier reconstructions; and (iii) latent space optimization as outlined to line 7 of Algorithm \ref{alg:latent_resample}. In contrast, our adaptation omits the pixel-space stage, as our focus is on maximizing perceptual quality. We found that this modification together with the inclusion of the REPA reguralizer yields sharper and more visually convincing results, while still benefiting from the refinement effect of the final latent-space consistency updates. The best hyperparameters used for this variant are reported in
Table~\ref{table:best_hparams_resample}.
In addition to the parameters $\kappa$ and $\lambda$, which play analogous roles to those
in Latent DPS, we also tune the maximum number of inner-loop optimization steps and the
corresponding inner learning rate of the Resample procedure
(see Algorithm~\ref{alg:latent_resample}).

\begin{table}[t]
\centering
\small
\caption{
Hyperparameters used for Resampling + \textsc{Repa}.
}
\begin{tabular}{lcccccccc}
\toprule
& \multicolumn{4}{c}{\textbf{ImageNet}} & \multicolumn{4}{c}{\textbf{FFHQ}} \\
\cmidrule(lr){2-5} \cmidrule(lr){6-9}
\textbf{Task}
& $\boldsymbol{\kappa}$ & $\boldsymbol{\lambda}$ & \textbf{max iters} & \textbf{inner lr}
& $\boldsymbol{\kappa}$ & $\boldsymbol{\lambda}$ & \textbf{max iters} & \textbf{inner lr} \\
\midrule
Super-resolution      & 0.05 & 3.25 & 150 & 0.005 & 0.01 & 2.25 & 100 & 0.0001 \\
Gaussian deblurring   & 0.075 & 0.5 & 300 & 0.005 & 0.05 & 0.75 & 300 & 0.00075 \\
Motion deblurring     & 0.5 & 0.75 & 300 & 0.005 & 0.05 & 0.75 & 100 & 0.005 \\
Box inpainting        & 0.025 & 100 & 200 & 0.0005 & 0.05 & 0.75 & 200 & 0.0005 \\
\bottomrule
\end{tabular}
\label{table:best_hparams_resample}
\end{table}



\begin{algorithm*}[t]
\caption{Resample + REPA Algorithm}
\label{alg:latent_resample}
\begin{algorithmic}[1]
\Require flow model $u_\theta $, measurement $y$, pretrained encoder $f$, resample steps $C$, parameter $\gamma$, proxy representation $c_{\text{proxy}}$, timestep $t_\text{cutoff}$, initialize $z_T \sim \mathcal{N}(0, I)$
\For{$t \in \{T, \dots, 0\}$}
    \State $\hat{z}_0 \gets \mathbb{E}[z_0 \mid z_t]$
    \State $z_{t-1} \gets z_t - \frac{1}{T} \cdot u_\theta(z_t, t)$
    \If{$t \in C$}
        \State $\tilde{z}_0(y) \gets \arg\min_{z}\tfrac{1}{2}\|y-\mathcal{A}(\mathcal{D}(z))\|_2^2$ \Comment{initialize at $\hat{z}_0$}
        \State $z_{t-1} \gets \textsc{StochasticResample}\!\left(\tilde{z}_0(y),\, z_{t-1},\, \gamma\right)$
    \EndIf
    \State \textcolor{blue}{$z_{t-1} \gets z_{t-1} + \lambda \nabla_{z_t} \sum_{n=1}^{N} \cos\!\big(c_{\text{proxy}}^{[n]},\, g_{\phi}(\mathrm{DiffEnc}^{[n]}(z_t,t))\big)$}
    \If{$t < t_{\text{cutoff}}$}
        \State $c_{\text{proxy}} \gets
        f_{\mathrm{DINOv2}}\!\bigl(
        \mathcal{D}(\mathbb{E}[z_0 \mid z_t])
        \bigr)$
    \EndIf
\EndFor
\State $x_0 \gets \mathcal{D}(z_0)$
\State \Return $x_0$
\end{algorithmic}
\end{algorithm*}

\subsection{Robustness of DINOv2 Representations to Corruptions}\label{sec:dino_rob}
To evaluate the robustness of DINOv2 representations under various corruptions, we conduct experiments on a fixed set of 100 images on the ImageNet dataset (Similar results hold for the FFHQ dataset as illustrated in Appendix \ref{subseq:CLIP}). For each image, we compute the average patch similarity between its ground truth representation and that of its corrupted version. We assess this similarity across different corruption types specifically super-resolution and Gaussian deblurring at varying levels of severity. We report the average similarity across the selected subset of 100 images.

\begin{figure}[ht]
    \centering

    \begin{subfigure}[t]{0.41\textwidth}
        \includegraphics[width=\linewidth]{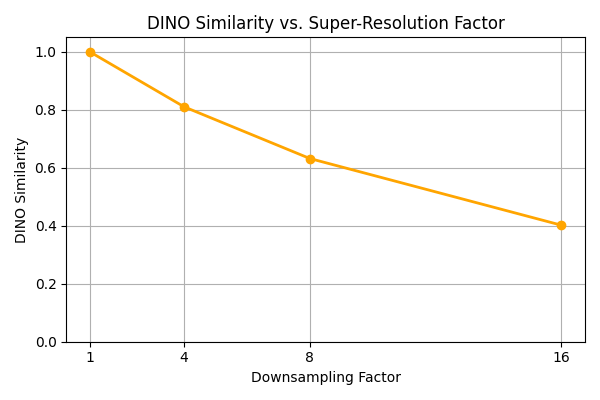}
        \caption{Super-Resolution}
    \end{subfigure}
    \hfill
    \begin{subfigure}[t]{0.41\textwidth}
        \includegraphics[width=\linewidth]{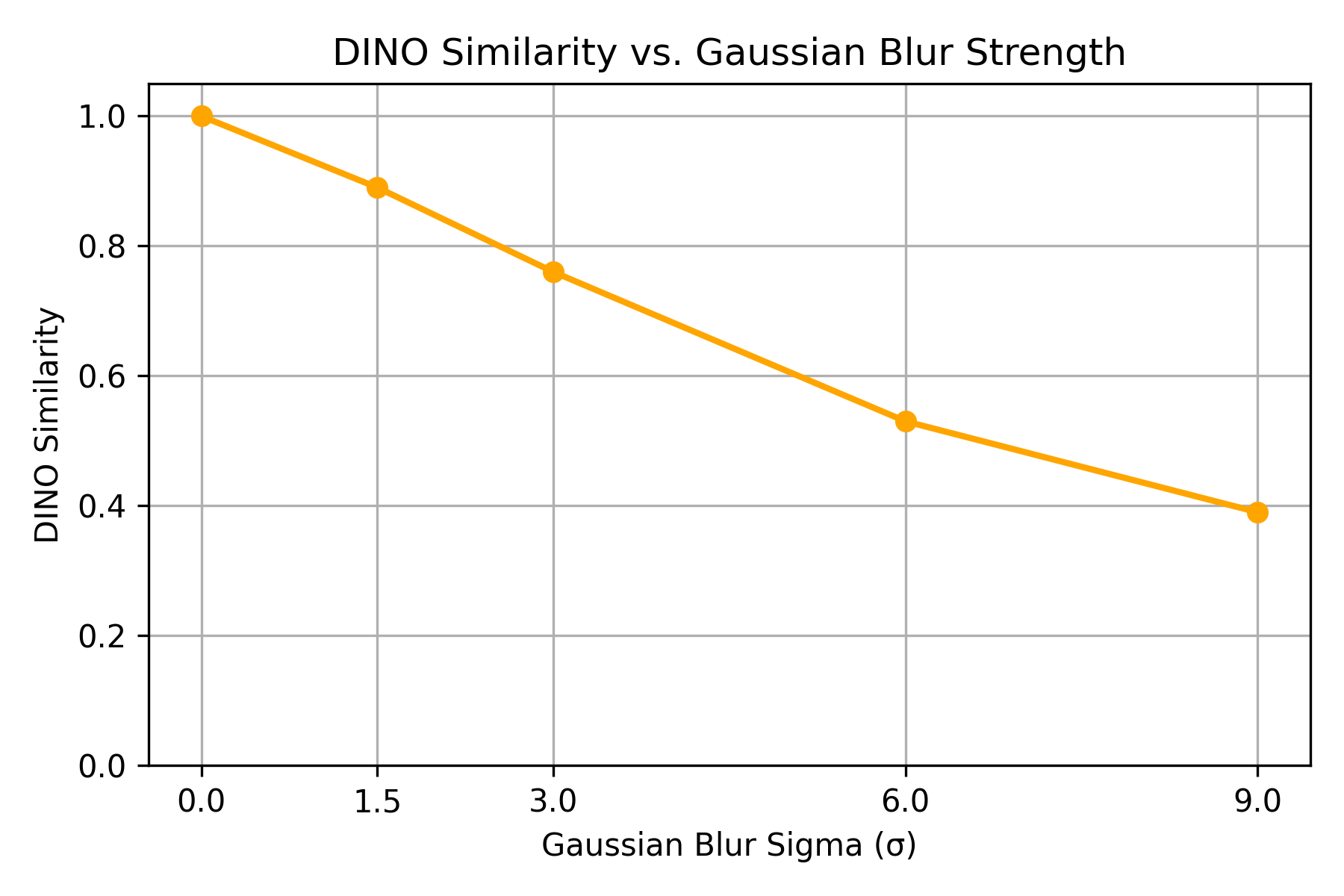}
        \caption{Gaussian Deblurring}
    \end{subfigure}

    \caption{Similarity of representations under increasing levels of corruption.}
    \label{fig:similarity_corruption_levels}
\end{figure}

\begin{figure}[H]
    \centering
    \begin{subfigure}[t]{0.24\textwidth}
        \includegraphics[width=\linewidth]{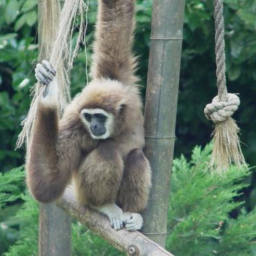}
        \caption{Ground Truth}
    \end{subfigure}
    \hfill
    \begin{subfigure}[t]{0.24\textwidth}
        \includegraphics[width=\linewidth]{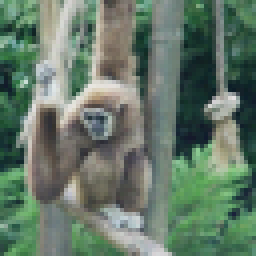}
        \caption{Super Resolution ×4}
    \end{subfigure}
    \hfill
    \begin{subfigure}[t]{0.24\textwidth}
        \includegraphics[width=\linewidth]{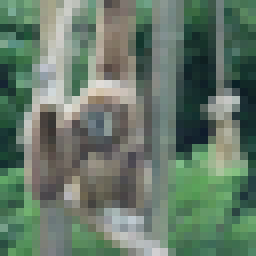}
        \caption{Super Resolution ×8}
    \end{subfigure}
    \hfill
    \begin{subfigure}[t]{0.24\textwidth}
        \includegraphics[width=\linewidth]{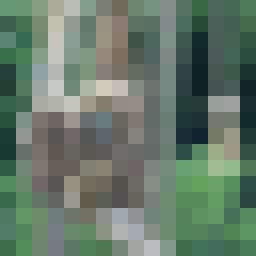}
        \caption{Super Resolution ×16}
    \end{subfigure}

    \vspace{0.2cm}
   
    \begin{subfigure}[t]{0.24\textwidth}
        \includegraphics[width=\linewidth]{media/00019.png}
        \caption{Ground Truth}
    \end{subfigure}
    \hfill
    \begin{subfigure}[t]{0.24\textwidth}
        \includegraphics[width=\linewidth]{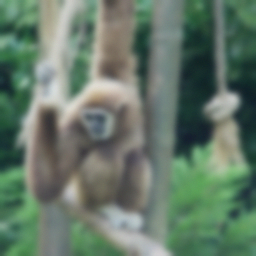}
        \caption{Gaussian deblur $\sigma = 3$}
    \end{subfigure}
    \hfill
    \begin{subfigure}[t]{0.24\textwidth}
        \includegraphics[width=\linewidth]{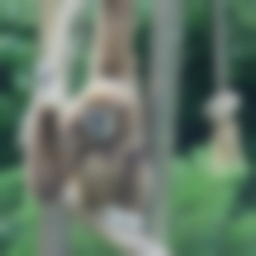}
        \caption{Gaussian deblur $\sigma = 6$}
    \end{subfigure}
    \hfill
    \begin{subfigure}[t]{0.24\textwidth}
        \includegraphics[width=\linewidth]{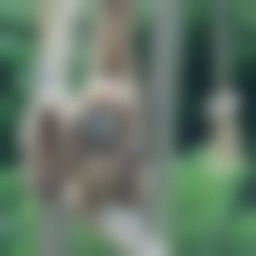}
        \caption{Gaussian deblur $\sigma = 9$}
    \end{subfigure}
    \vspace{2pt}
    \vspace{0.2cm}
    \caption{Visual comparison of a corrupted images for different corruption types and severity levels.}
    \label{fig:corruption_same_image}
\end{figure}

Figure~\ref{fig:corruption_same_image} first visualizes how the pixel-space appearance of a single image changes across different corruption levels. 
Figure~\ref{fig:similarity_corruption_levels} then shows how the DINOv2 representation similarity changes as the corruption severity increases for the two tasks. 
Interestingly, we observe that DINOv2 representations remain significantly more robust to both super-resolution and Gaussian deblurring. 
Despite substantial visual degradation in pixel space, DINOv2 features maintain a strong alignment with the original image features.

\subsection{Selection of Proxy Representation}
\label{sec:proxy_selection}

Since pretrained DINOv2 features are robust to a wide range of common image corruptions,
we use the DINOv2 encoding of the measurement as an initial proxy representation,
\(f_{\mathrm{DINOv2}}(y)\).
We find this choice to be effective across all considered inverse problems.
In addition, we introduce a cutoff timestep at which the proxy representation is updated
to depend on the model’s current denoised estimate, namely
\(f_{\mathrm{DINOv2}}(\mathcal{D}(\mathbb{E}[z_0 \mid z_t]))\).
\begin{itemize}
    \item \textbf{Box Inpainting.}
    We construct the proxy representation by combining the observed regions of
    the measurement with the model’s current denoised estimate:
    \begin{equation}
        c_{\text{proxy}}
        =
        \mathrm{mask} \odot f_{\mathrm{DINOv2}}(y)
        +
        (I - \mathrm{mask}) \odot
        f_{\mathrm{DINOv2}}\!\bigl(
        \mathcal{D}(\mathbb{E}[z_0 \mid z_t])
        \bigr),
    \end{equation}
    where \(\mathrm{mask}\) denotes the binary inpainting mask.
    \item \textbf{Super-resolution, Gaussian Deblurring, and Motion Deblurring.}
For these tasks, we use \(f_{\mathrm{DINOv2}}(y)\) as the proxy representation during
the early stages of sampling.
On the FFHQ dataset, we find it beneficial to switch to the reconstruction-based proxy
\(f_{\mathrm{DINOv2}}(\mathcal{D}(\mathbb{E}[z_0 \mid z_t]))\) after
\(80\%\) of the diffusion steps.
On ImageNet, this transition does not lead to consistent improvements, and we therefore use
the measurement-based proxy throughout. We include an ablation to justify these choices in \ref{subsec:ablations}. 
\end{itemize}

\subsection{Results Using a CLIP Model}\label{subseq:CLIP}
In this subsection, we compare the results obtained using a DINOv2 encoder with those obtained using a CLIP model. To this end, we train a model with representation alignment using CLIP representations instead of DINOv2 representations on the FFHQ dataset. Both models were trained for 100K iterations using a batch size of 256. Table~\ref{table:Dino_clip} presents quantitative results when applying our inference-time framework to both models. As we can see, the inclusion of our regularizer leads to consistent improvements across all tasks and metrics.

\begin{figure}[ht]
\centering
\includegraphics[width=0.4\textwidth]{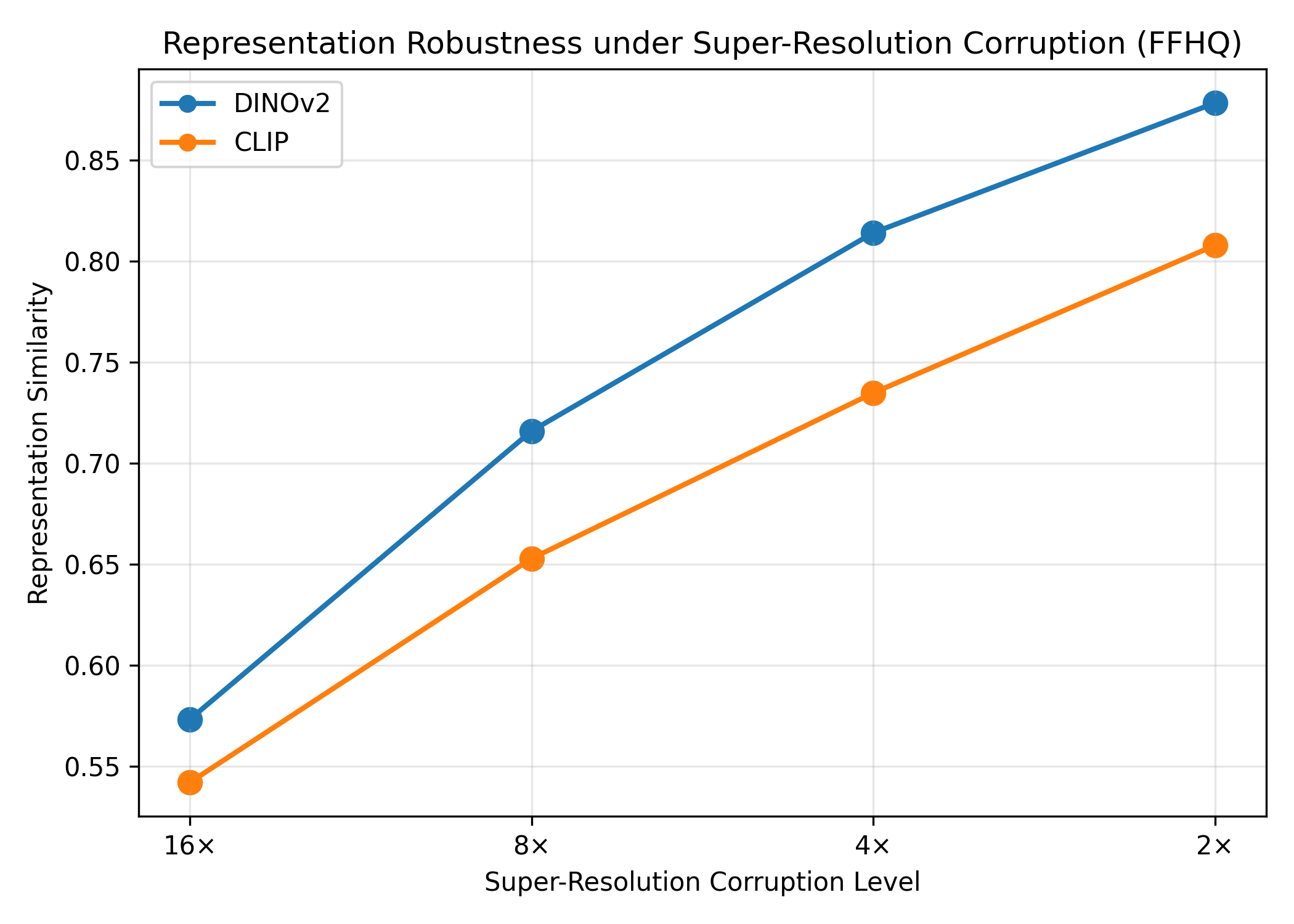}\hfill
\includegraphics[width=0.4\textwidth]{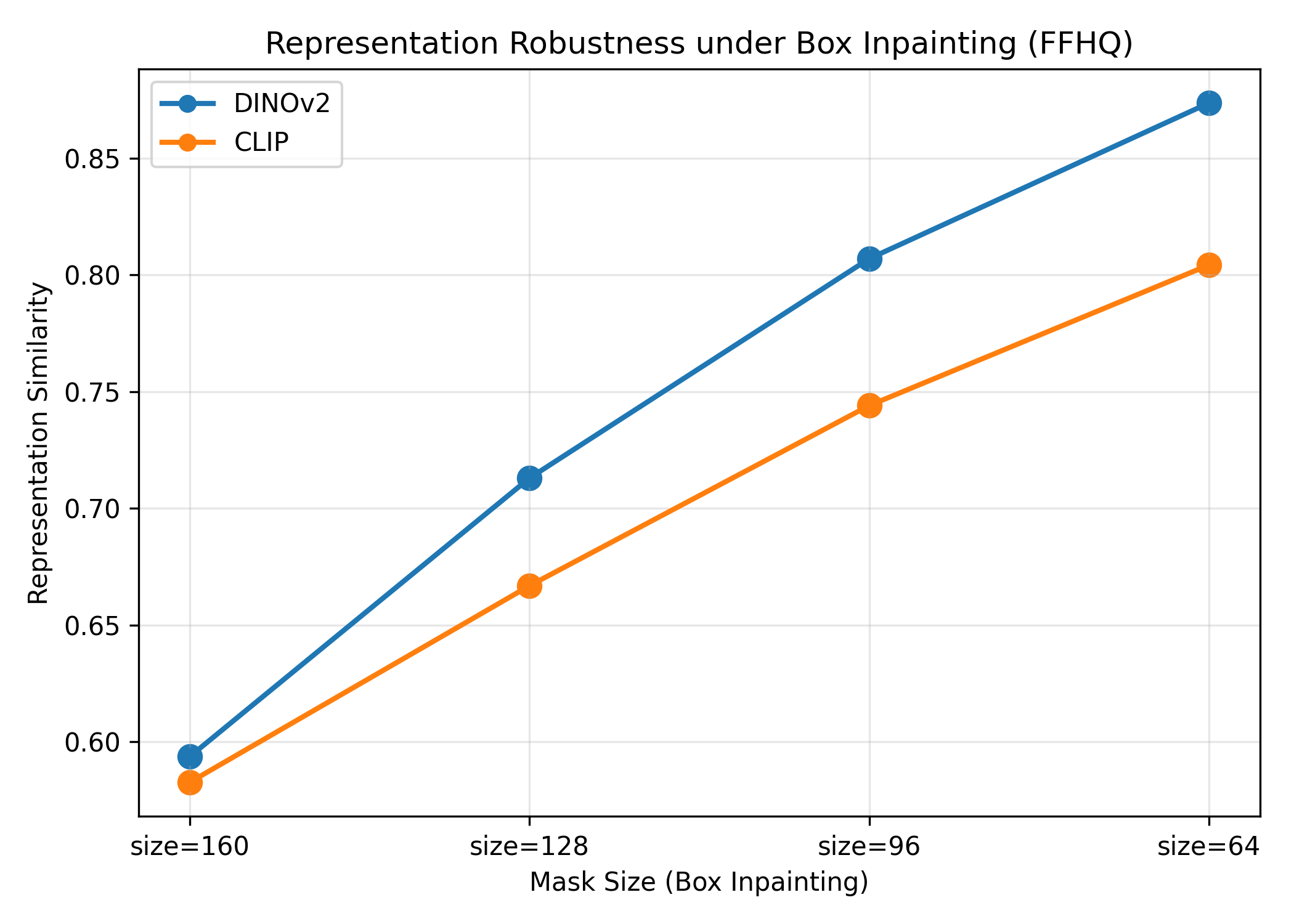}\hfill
\caption{Comparison of the DINOv2 and CLIP robustness to common degradations.}
\label{fig:comp_clip_dino}
\end{figure}

To compare the influence of DINOv2 and CLIP, we make the following observations. First, the baseline algorithm performs slightly better when the model has been trained using DINOv2 representations rather than CLIP. This result is consistent with the findings of \cite{yu2024repa}, where the best-performing model was also trained using DINOv2 representations, suggesting that DINOv2-based representation alignment leads to a stronger generative prior. Regarding the comparison of our test-time regularizer between DINOv2 and CLIP, we observe that DINOv2 representations appear to be more robust to the standard degradations used in inverse problem evaluation. We illustrate this behavior in Figure~\ref{fig:comp_clip_dino}.
\begin{table*}[ht]
\centering
\small
\caption{Performance comparison on FFHQ ($\sigma = 0.05$) for models trained with DINOv2 and CLIP representations.}
\begin{tabular}{llcccc}
\toprule
\textbf{Task} & \textbf{Method} & LPIPS↓ & FID↓ & PSNR↑ & SSIM↑ \\
\midrule

\multirow{4}{*}{$4\times$ SR}
& Latent DPS (DINOv2-trained) & 0.188 & 56.69 & 28.99 & 0.814 \\
& Latent DPS + REPA (DINOv2-trained) & \textbf{0.176} & \textbf{51.27} & \textbf{29.12} & \textbf{0.819} \\
& Latent DPS (CLIP-trained) & 0.190 & 63.63 & 28.95 & 0.811 \\
& Latent DPS + REPA (CLIP-trained) & 0.183 & 54.88 & 28.99 & 0.815 \\
\midrule
\multirow{4}{*}{Box Inpainting}
& Latent DPS (DINOv2 Trained)& 0.192 & 65.89 & 23.55 & \textbf{0.785} \\
& Latent DPS + REPA (DINOv2 Trained) & \textbf{0.178} & \textbf{57.04} & \textbf{23.83} & 0.784 \\
& Latent DPS (CLIP-trained) & 0.193 & 54.01 & 23.53 & 0.784 \\
& Latent DPS + REPA (CLIP-trained) & 0.185 & 52.45 & 23.65 & 0.784 \\
\midrule
\multirow{4}{*}{Gaussian Deblur}
& Latent DPS (DINOv2-trained) & 0.192 & 60.65 & 28.15 & 0.783 \\
& Latent DPS + REPA (DINOv2-trained) & \textbf{0.186} & \textbf{55.53} & \textbf{28.21} & \textbf{0.787} \\
& Latent DPS (CLIP-trained) & 0.193 & 60.72 & 28.02 & 0.782 \\
& Latent DPS + REPA (CLIP-trained) & 0.188 & 59.76 & 28.07 & 0.779 \\
\bottomrule
\end{tabular}
\label{table:Dino_clip}
\end{table*}


\subsection{Solving Inverse Problems under Severe Degradations}\label{subsec:severe_deg}
\begin{figure}[H]
\centering
\begin{minipage}{0.4\linewidth}
    \centering
    \includegraphics[width=\linewidth]{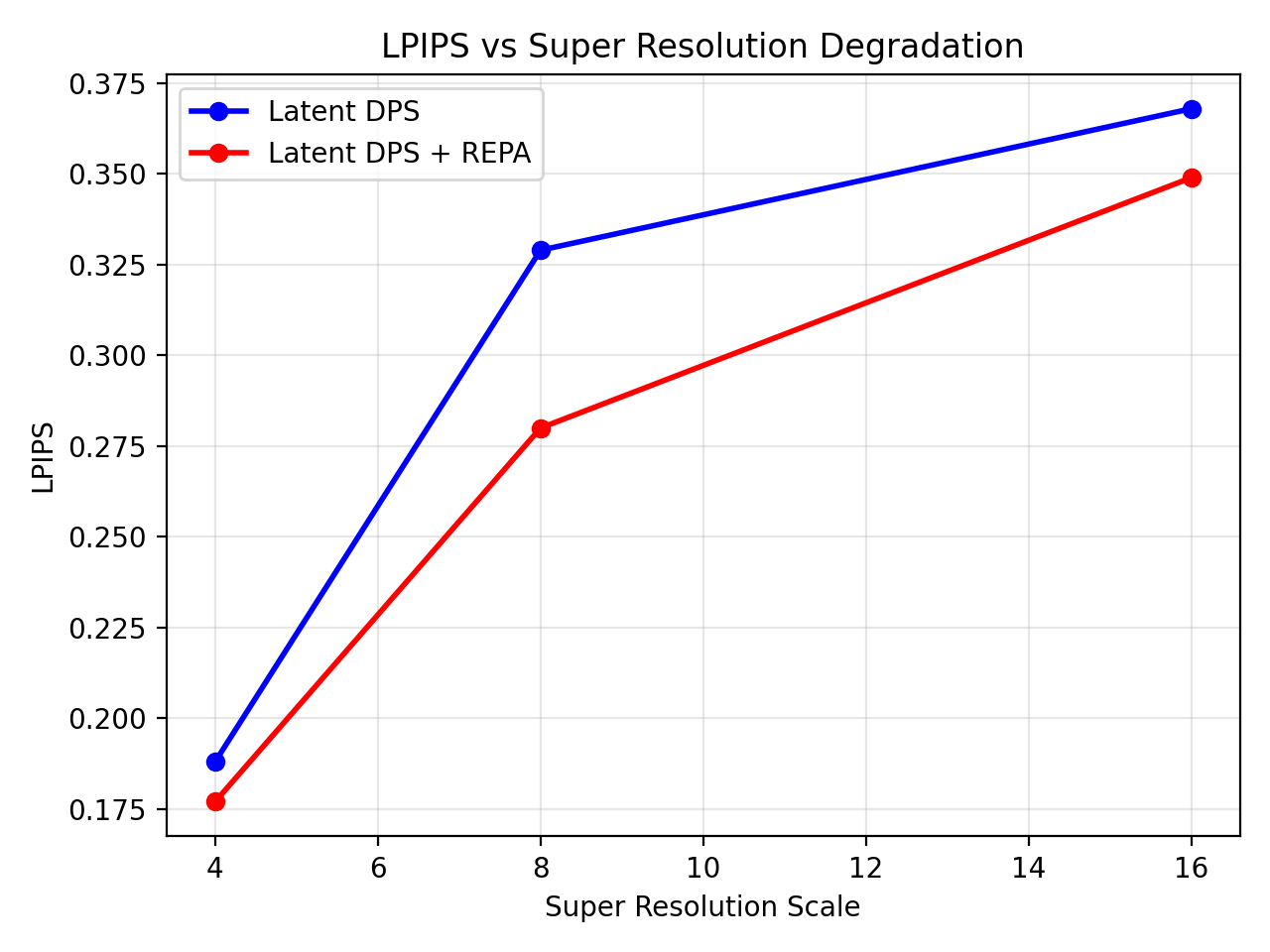}
    \caption*{LPIPS vs Super Resolution Degradation}
\end{minipage}
\hfill
\begin{minipage}{0.4\linewidth}
    \centering
    \includegraphics[width=\linewidth]{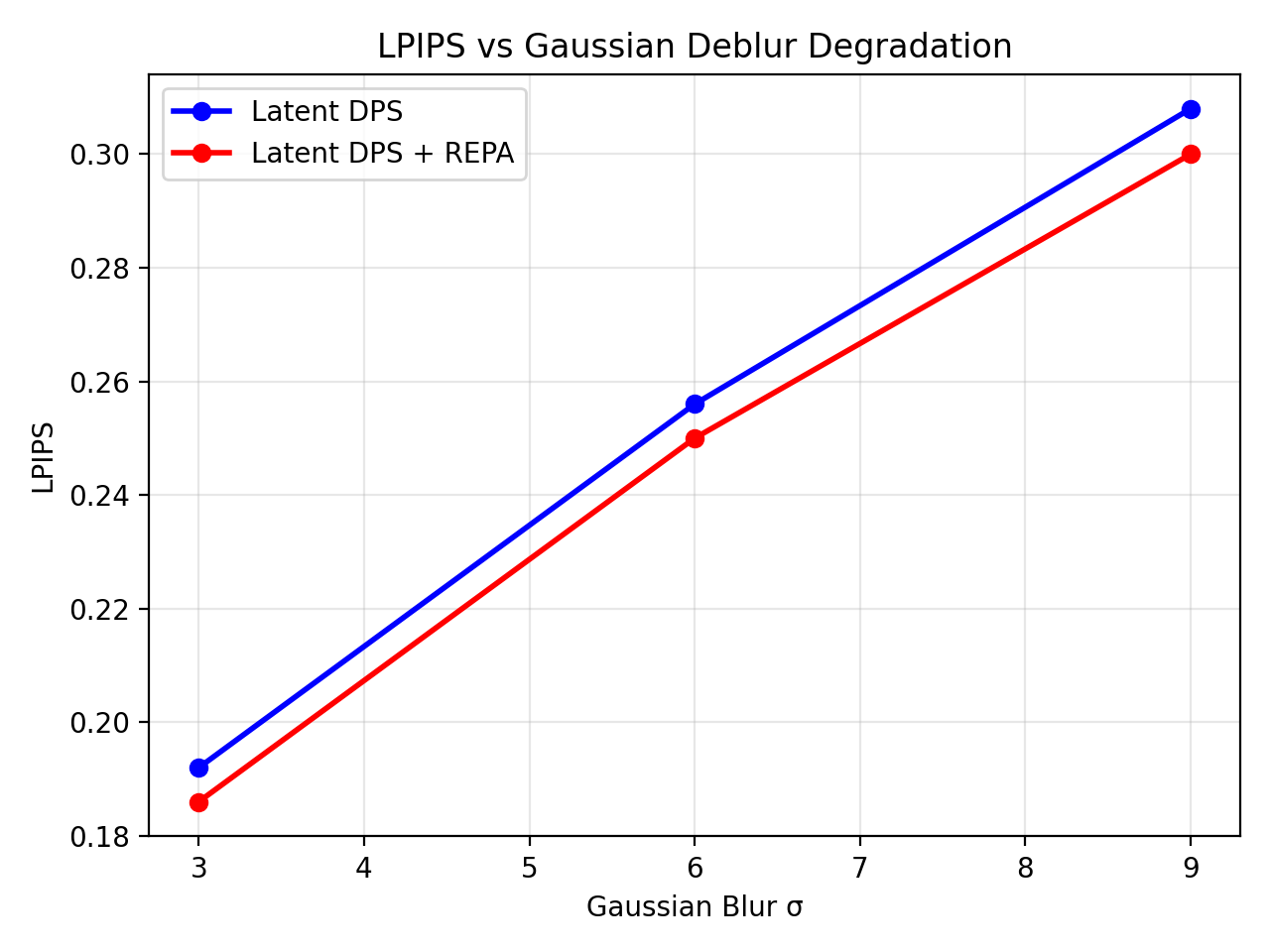}
    \caption*{LPIPS vs Gaussian Deblur Degradation}
\end{minipage}
\caption{LPIPS as a function of degradation severity for super-resolution (left) and Gaussian deblurring (right). REPA consistently improves perceptual quality across all degradation levels.}
\label{fig:lpips_severe}
\end{figure}

In this subsection, we study the effect of the $\textsc{REPA}$ regularizer under severe degradations. 
We evaluate the performance of our method using Latent DPS on the FFHQ dataset for both super-resolution and Gaussian deblurring. 
For super-resolution, we consider more challenging settings with $8\times$ and $16\times$ upscaling factors, while for Gaussian deblurring we consider stronger blur levels with $\sigma = 6$ and $\sigma = 9$. 
Figure~\ref{fig:lpips_severe} shows how LPIPS varies as the degradation severity increases. 
We also include the full quantitative results in Table~\ref{tab:severe_results}.   

\begin{table}[t]
\centering
\small
\caption{Performance under severe degradations on FFHQ using Latent DPS.}
\label{tab:severe_results}
\begin{tabular}{llcccc}
\toprule
\textbf{Task} & \textbf{Method} & LPIPS↓ & FID↓ & PSNR↑ & SSIM↑ \\
\midrule

\multirow{2}{*}{$8\times$ Super Resolution}
& Latent DPS & 0.329 & 92.18 & 22.85 & 0.593 \\
& Latent DPS + REPA & 0.280 & 72.70 & 23.24 & 0.617 \\

\midrule

\multirow{2}{*}{$16\times$ Super Resolution}
& Latent DPS & 0.368 & 109.87 & 20.11 & 0.499 \\
& Latent DPS + REPA & 0.349 & 95.87 & 20.41 & 0.523 \\
\midrule

\multirow{2}{*}{Gaussian Deblur ($\sigma = 6$)}
& Latent DPS & 0.256 & 77.82 & 24.35 & 0.658 \\
& Latent DPS + REPA & 0.250 & 72.43 & 24.42 & 0.665 \\

\midrule

\multirow{2}{*}{Gaussian Deblur ($\sigma = 9$)}
& Latent DPS & 0.308 & 86.58 & 21.82 & 0.566 \\
& Latent DPS + REPA & 0.300 & 81.84 & 21.97 & 0.582 \\

\bottomrule
\end{tabular}
\end{table}

\subsection{Extension to Non Linear Problems}\label{subsec:non_linear}
In this section, we extend our evaluation of the \textsc{REPA} regularizer to more complex inverse problems. 
Specifically, we evaluate our method on non-linear deblurring and high dynamic range reconstruction using the degradation model of \cite{zhang2025improving}. 
Table~\ref{tab:nonlinear_results} presents quantitative results for Latent DPS and Resample, indicating improved performance when incorporating the \textsc{REPA} regularizer. 

\begin{table*}[ht]
\centering
\small
\caption{Performance comparison on ImageNet for non-linear degradation tasks.}
\label{tab:nonlinear_results}
\begin{tabular}{llcccc}
\toprule
 &  & \multicolumn{4}{c}{\textbf{ImageNet}} \\
\textbf{Task} & \textbf{Method} & LPIPS↓ & FID↓ & PSNR↑ & SSIM↑ \\
\midrule
\multirow{4}{*}{Non-linear Deblurring}
& Latent DPS & 0.302 & 170.41 & 24.62 & 0.653 \\
& Latent DPS + REPA & 0.266 & 105.73 & 24.52 &  0.645\\
& Resample & 0.276 & 141.44 & 25.04 & 0.661 \\
& Resample + REPA & \textbf{0.248} & \textbf{103.96} & \textbf{25.13} & \textbf{0.668} \\
\midrule
\multirow{4}{*}{HDR Reconstruction}
& Latent DPS & 0.310 & 138.78 & 23.12 & 0.690 \\
& Latent DPS + REPA & \textbf{0.258} & \textbf{78.24} & \textbf{24.11} & 0.701 \\
& Resample & 0.295 & 125.00 & 23.61 & 0.701 \\
& Resample + REPA & 0.282 & 108.41 & 24.01 & \textbf{0.711}  \\
\bottomrule
\end{tabular}
\end{table*}

\subsection{Effect of the Choice of Transformer Block for Representation Alignment}\label{subsec:effect_block}

\begin{figure}[ht]
\centering
\includegraphics[width=0.4\linewidth]{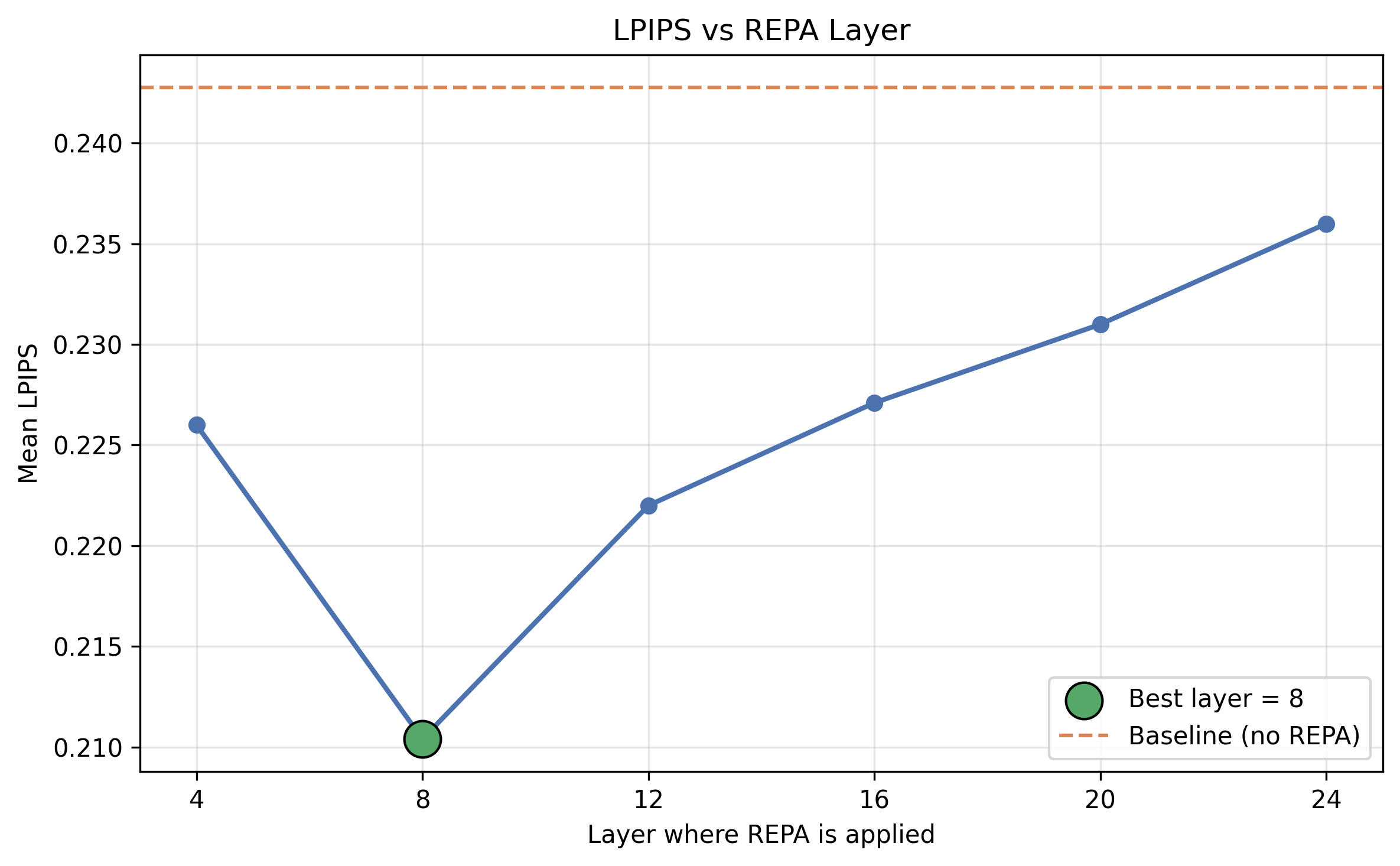}
\caption{Ablation study on the transformer block used for representation alignment.}
\label{fig:layer_ablation}
\end{figure}

We include an ablation study investigating the transformer block at which representation alignment is applied. 
We use the ImageNet model from \cite{yu2024repa} and study the effect of changing the alignment layer for the task of $4\times$ super-resolution. 
The mapping $g_{\phi}$ for the ImageNet model was trained using representations from the eighth transformer block. 
Figure~\ref{fig:layer_ablation} shows how performance varies when alignment is applied at different transformer blocks. 
The reported LPIPS values are averaged over 50 images. 
We observe that applying alignment at different blocks still leads to performance improvements; however, the largest improvements are obtained when alignment is applied at the same block on which the mapping $g_{\phi}$ was trained.

\subsection{Ablation Studies}\label{subsec:ablations}

\begin{figure}[t]
\centering

\begin{tabular}{ccc}
\textbf{t\_cutoff} & \textbf{Regularization Strength ($\lambda$)} & \textbf{Learning Rate (k)} \\

\includegraphics[width=0.3\textwidth]{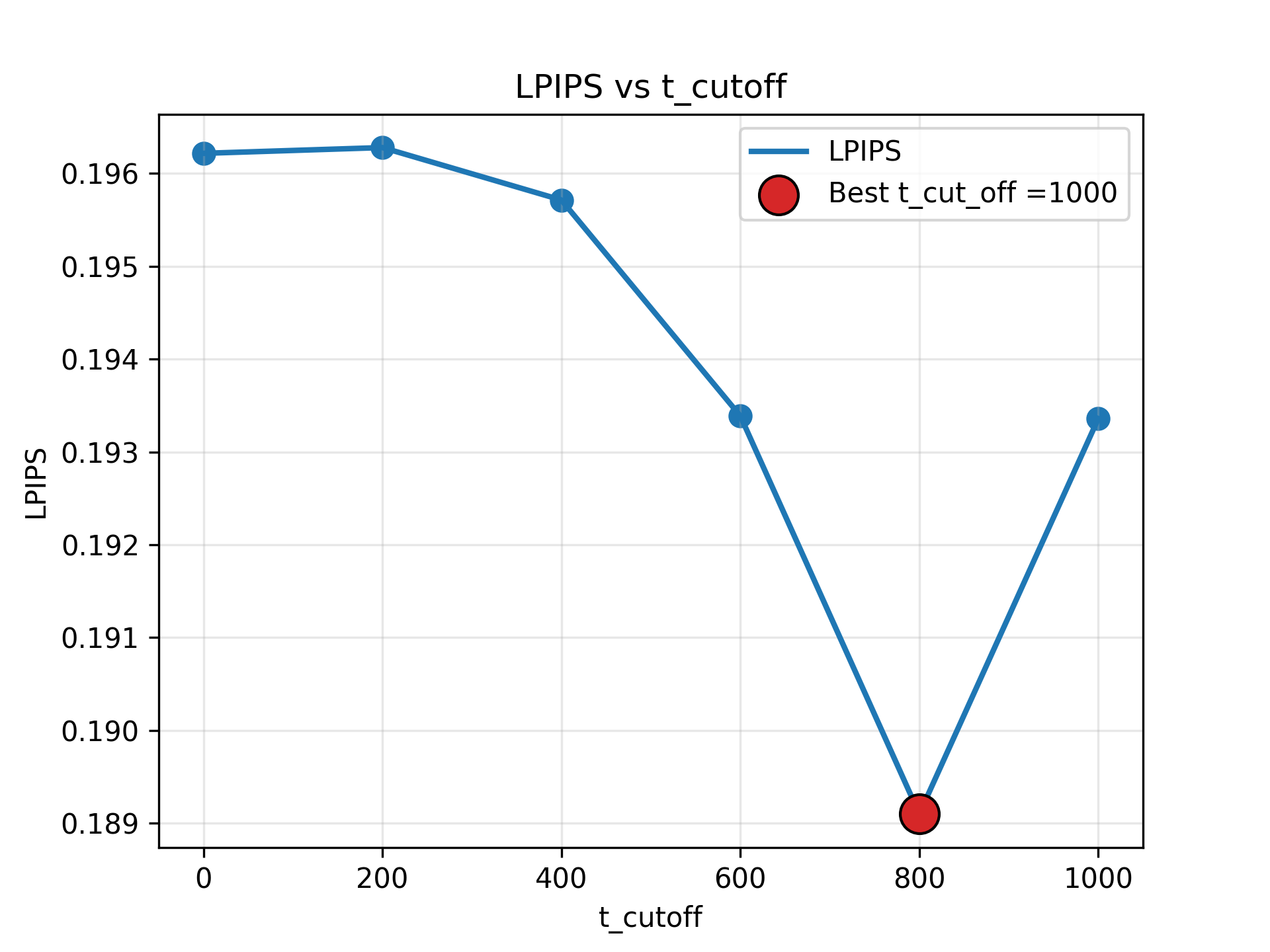} &
\includegraphics[width=0.3\textwidth]{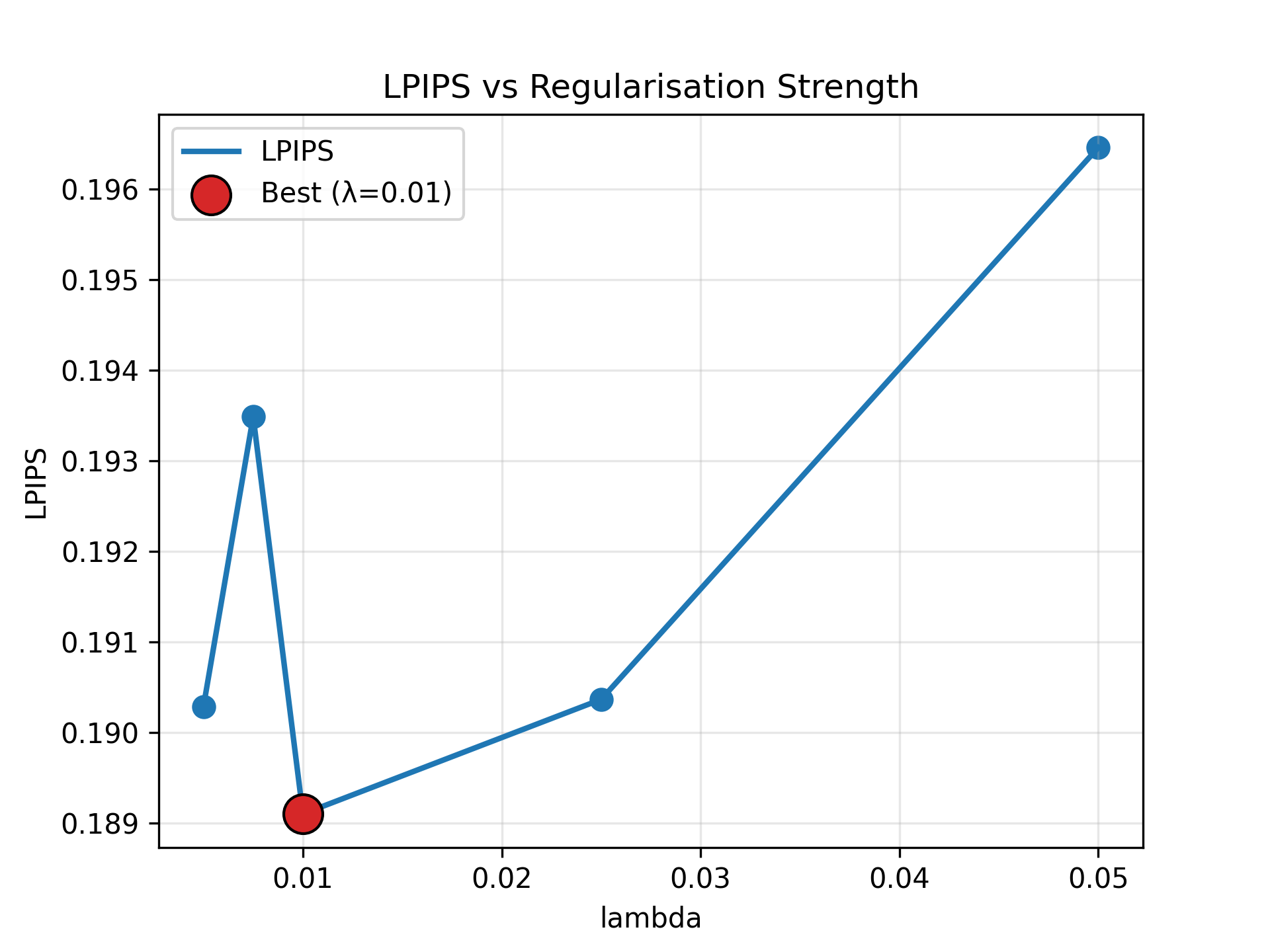} &
\includegraphics[width=0.3\textwidth]{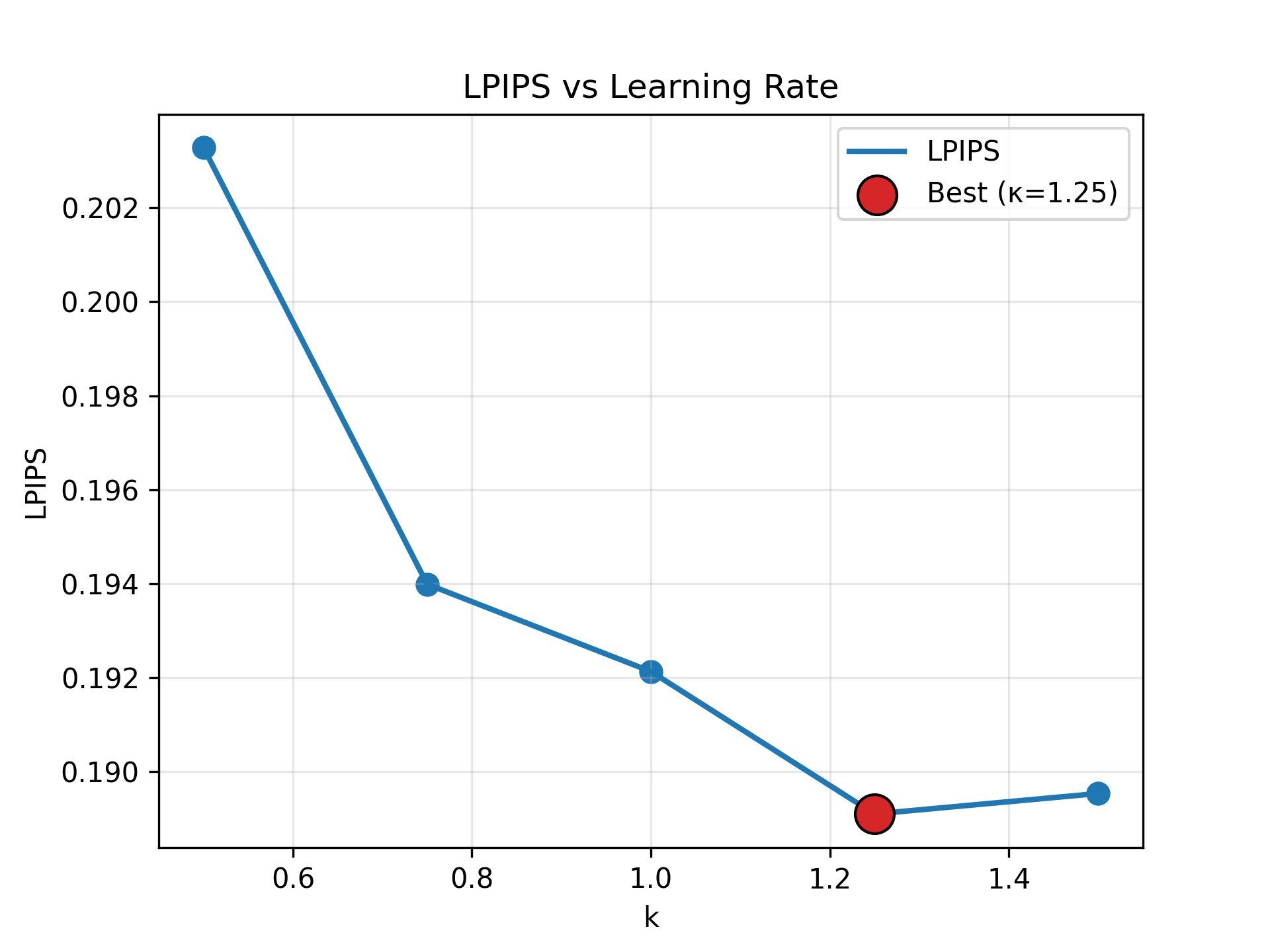} \\

\includegraphics[width=0.3\textwidth]{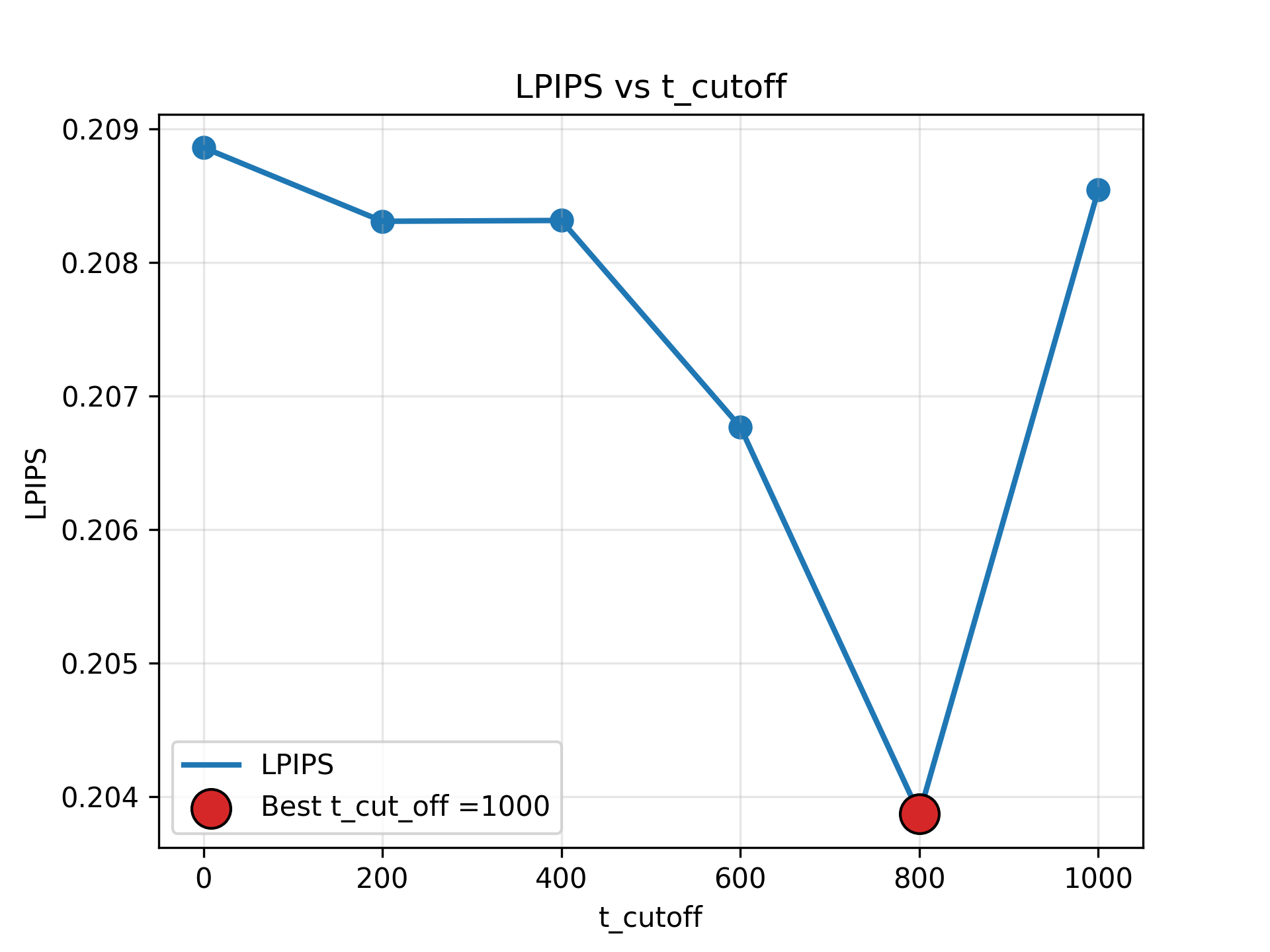} &
\includegraphics[width=0.3\textwidth]{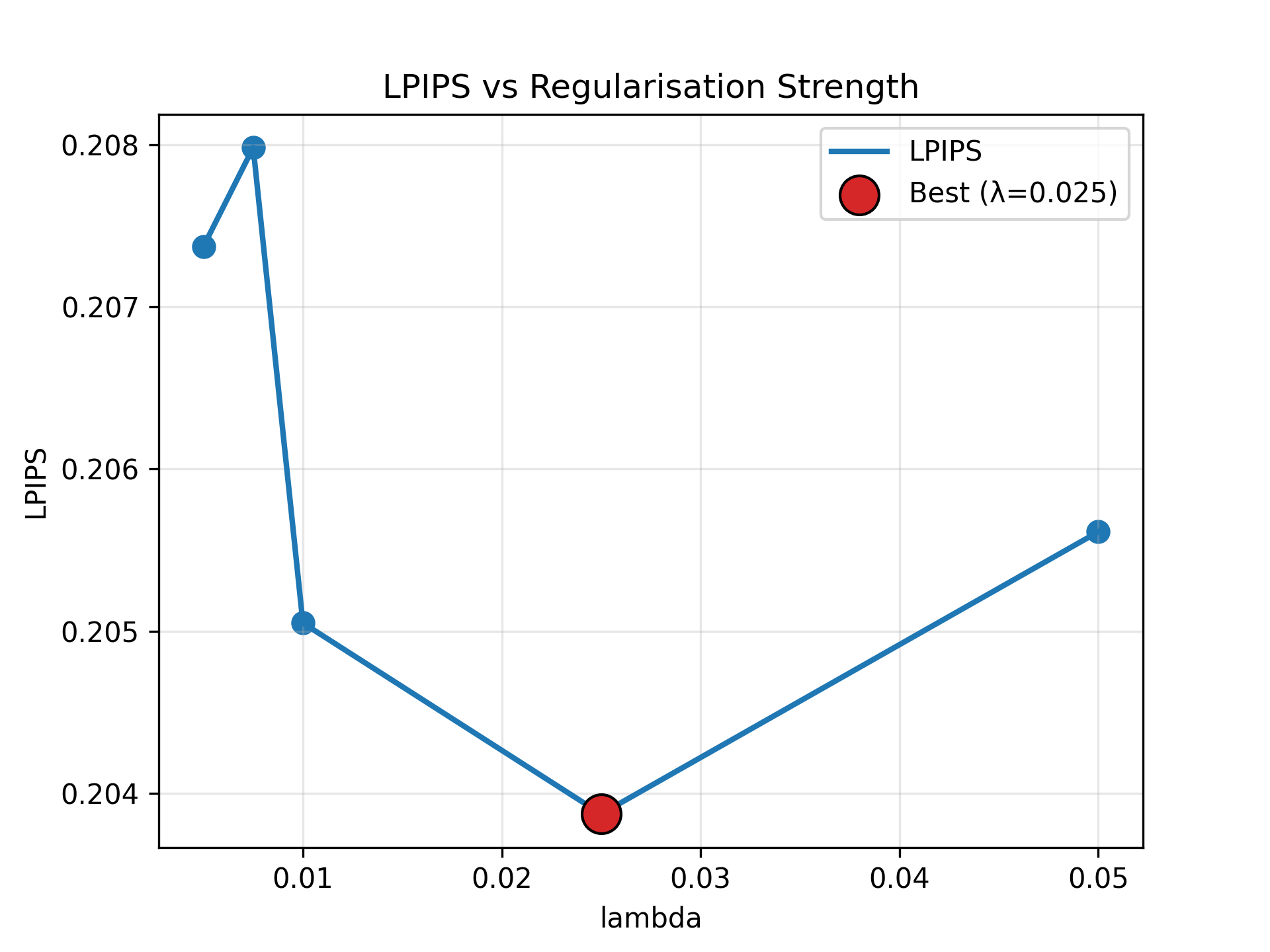} &
\includegraphics[width=0.3\textwidth]{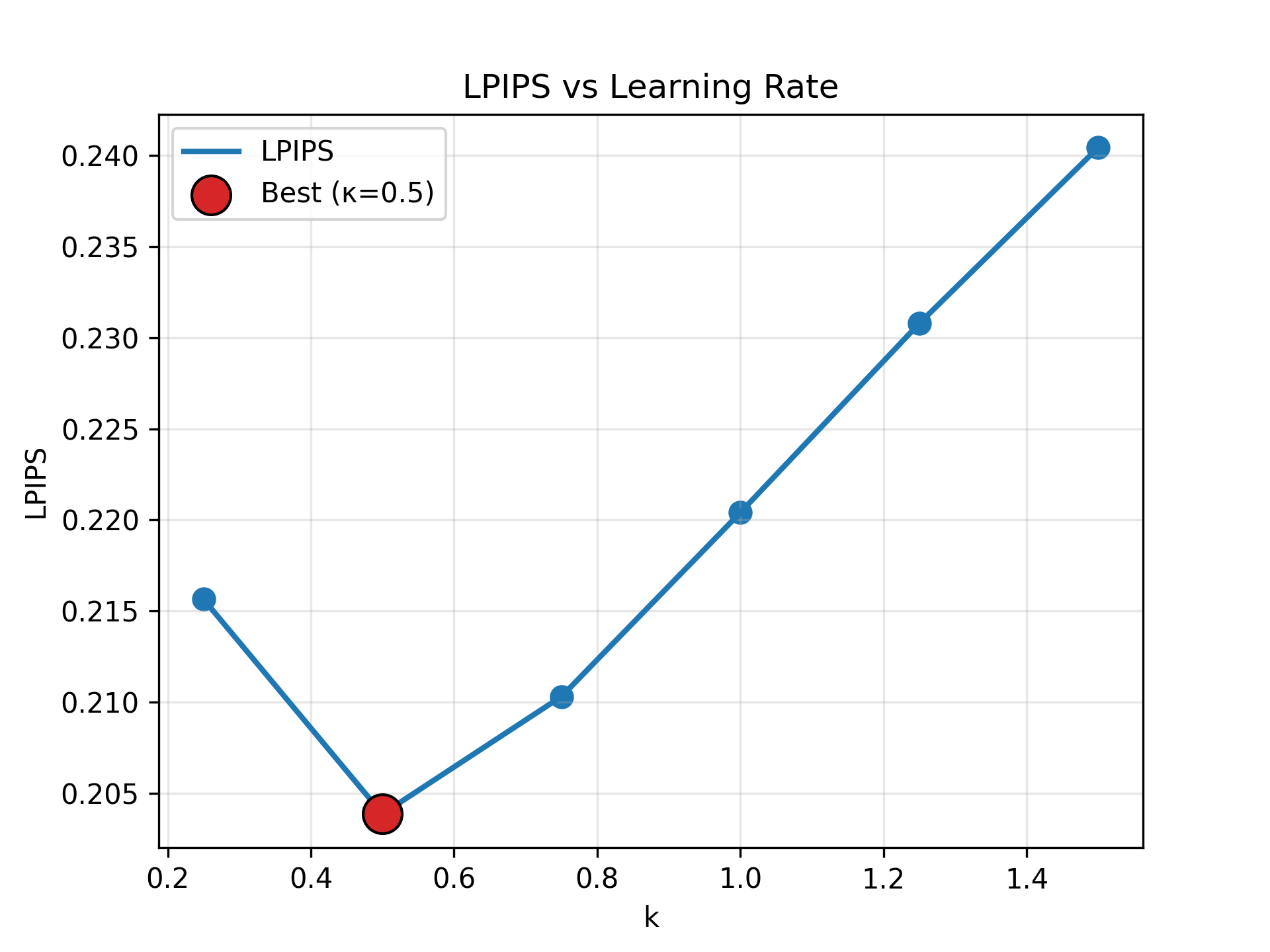} \\

\includegraphics[width=0.3\textwidth]{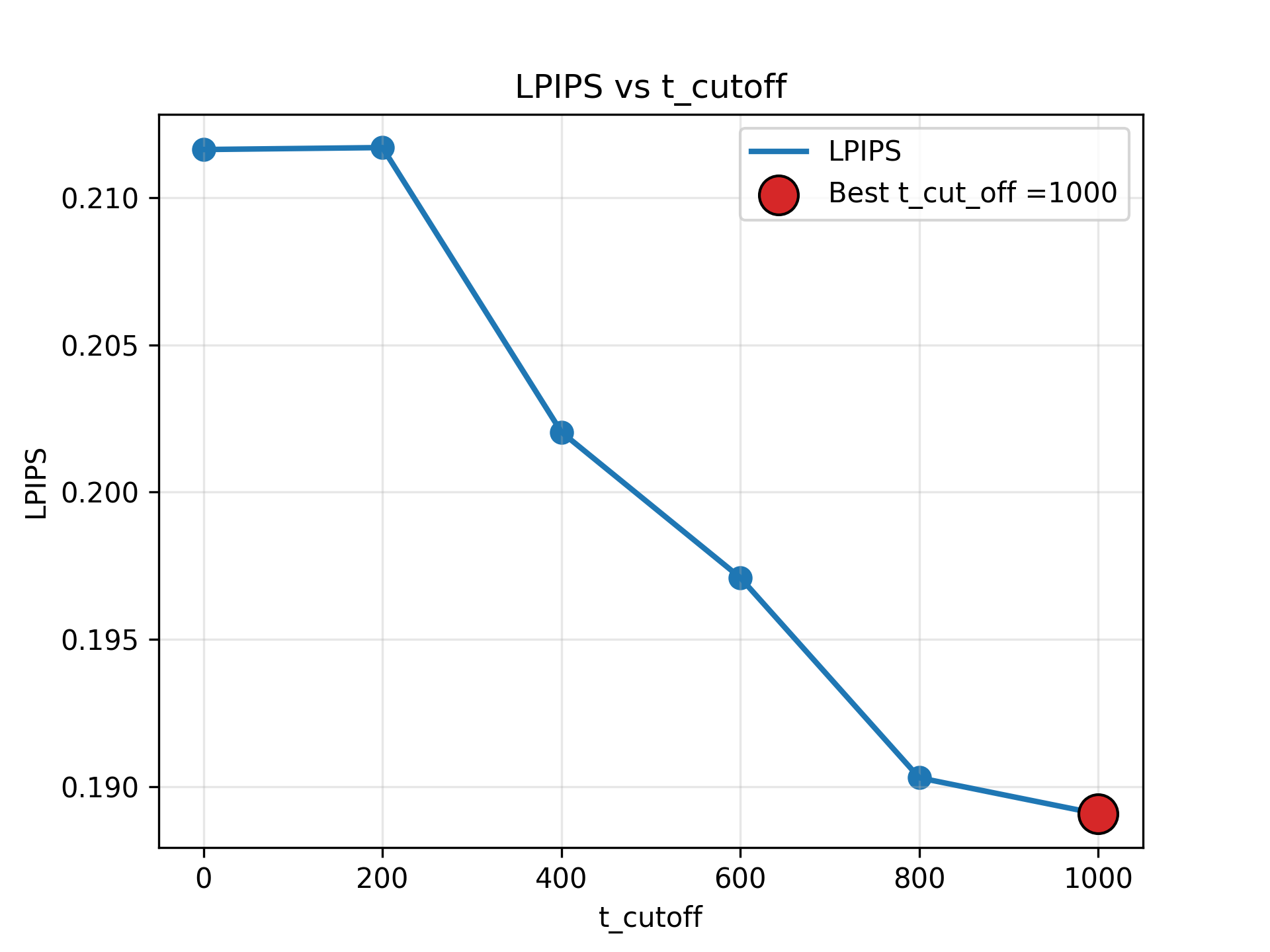} &
\includegraphics[width=0.3\textwidth]{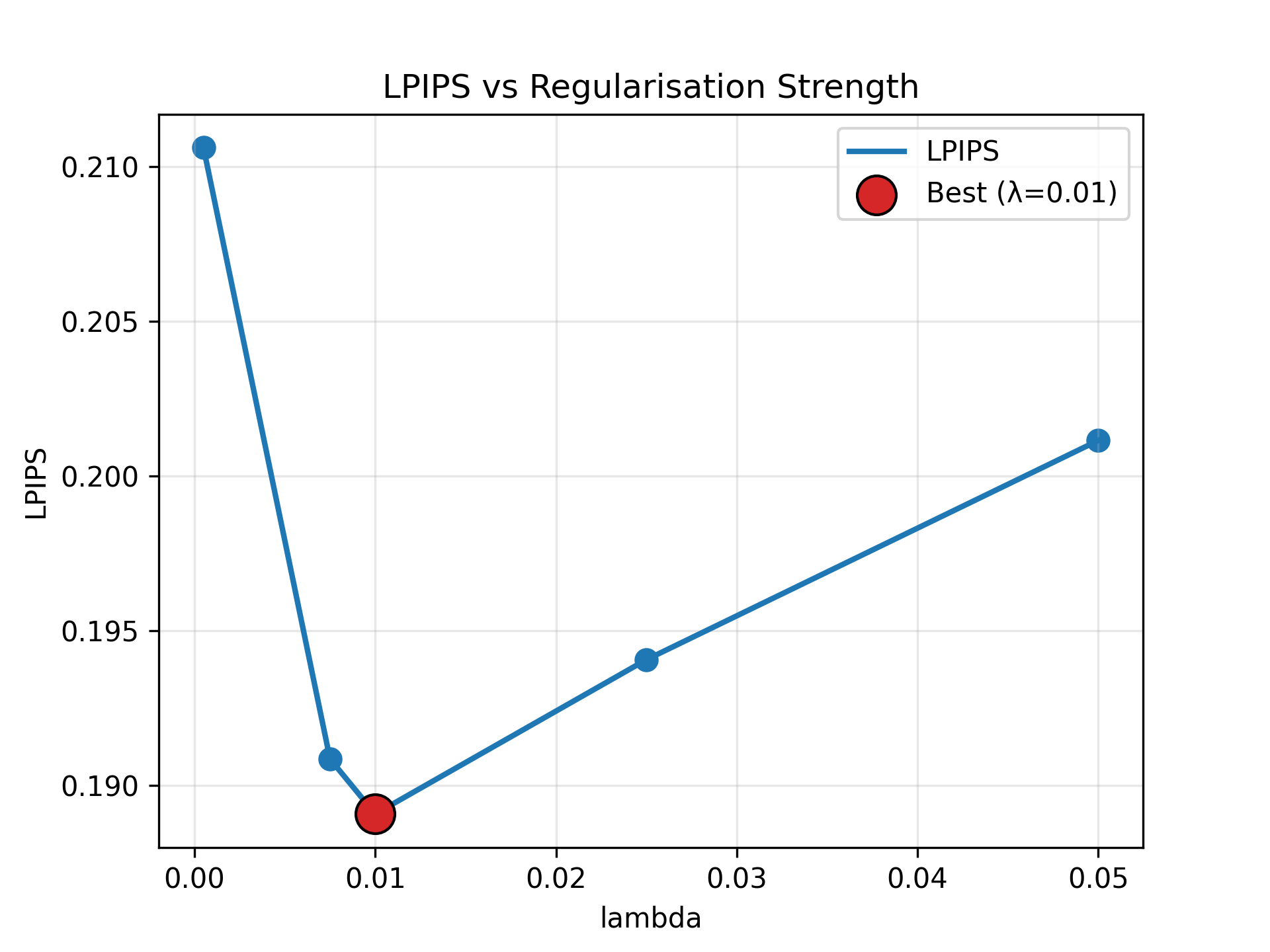} &
\includegraphics[width=0.3\textwidth]{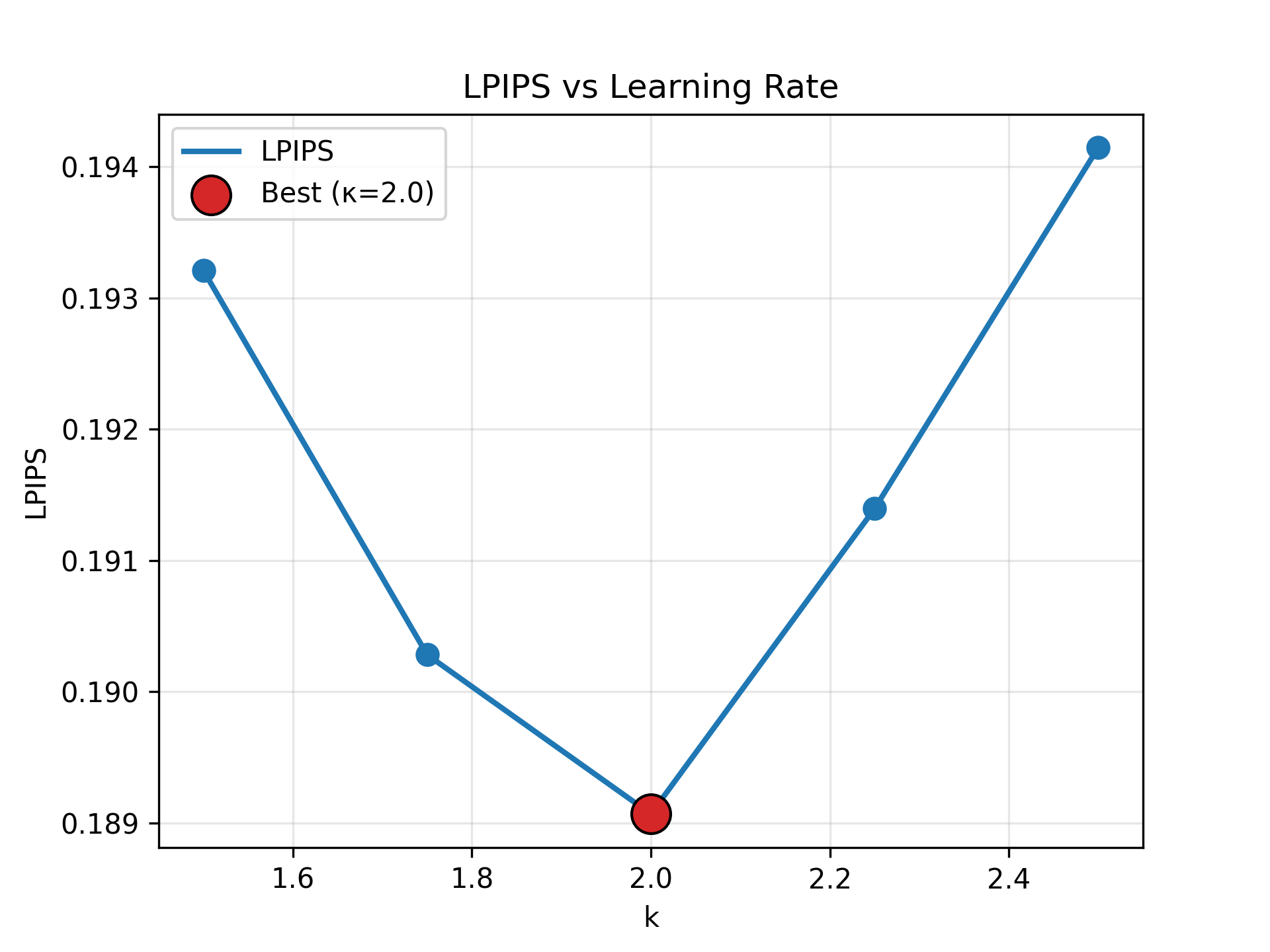} \\

\includegraphics[width=0.3\textwidth]{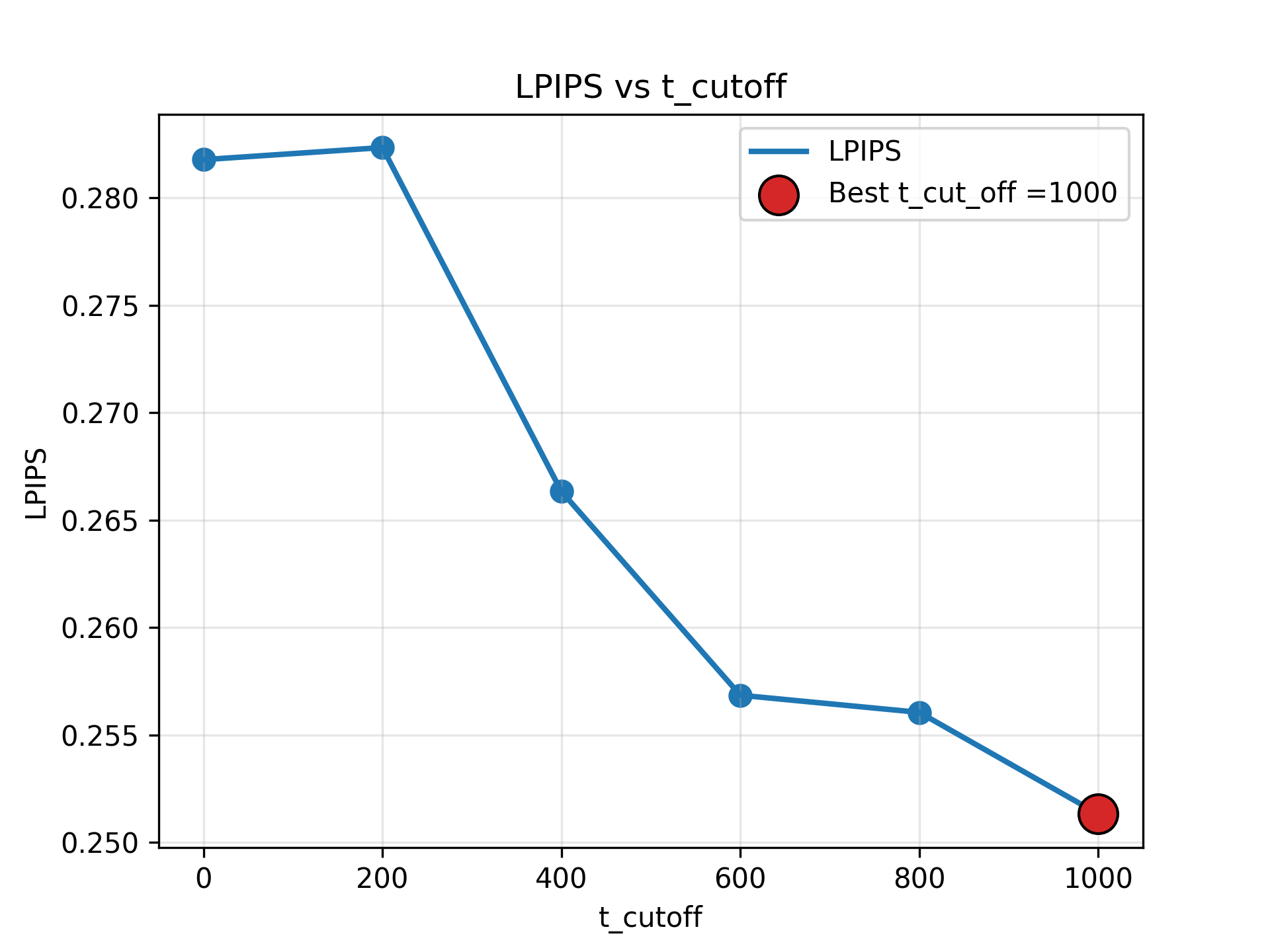} &
\includegraphics[width=0.3\textwidth]{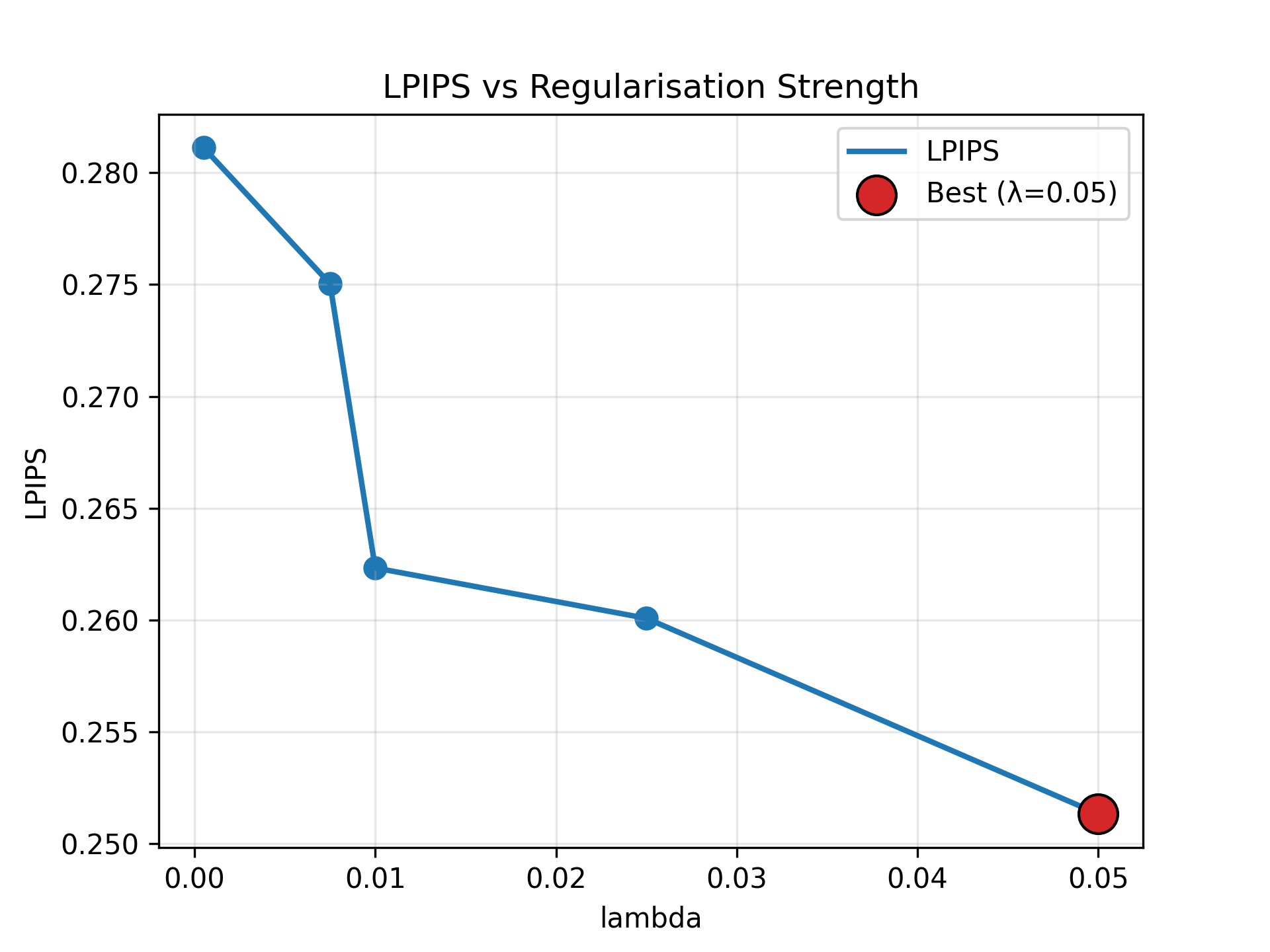} &
\includegraphics[width=0.3\textwidth]{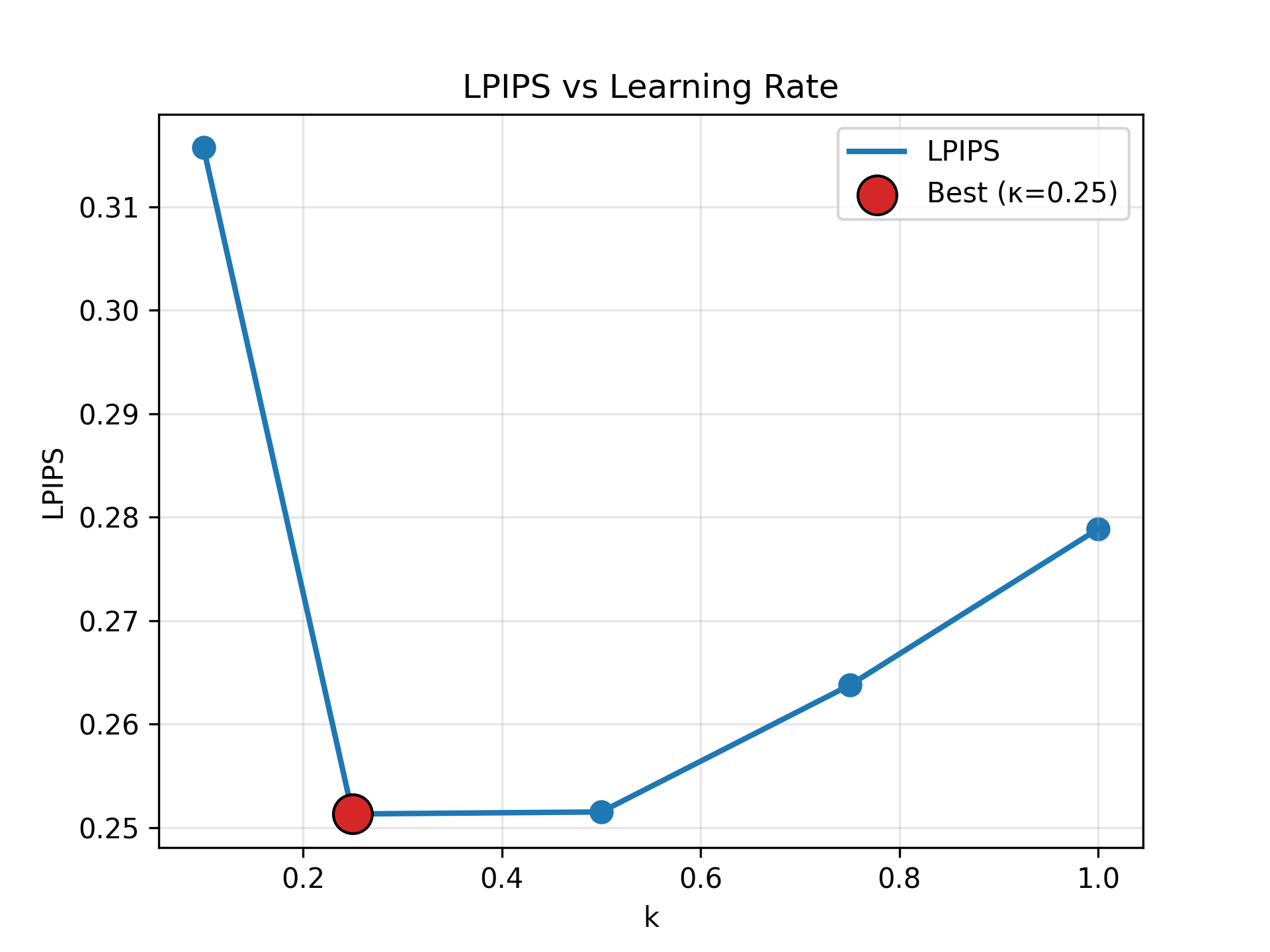} \\

\end{tabular}

\caption{
Ablation studies on the main hyperparameters of our framework. Rows correspond to different tasks and datasets: the first row shows FFHQ super-resolution, the second row FFHQ Gaussian deblurring, the third row ImageNet super-resolution, and the fourth row ImageNet Gaussian deblurring. Each plot shows the best LPIPS value achievable when the corresponding hyperparameter is fixed and the remaining hyperparameters are varied.
}
\label{fig:ablation}
\end{figure}

In this subsection, we present ablation studies on the main hyperparameters of our framework. 
The hyperparameters were selected through a small grid search on a validation set of 15 images. 
The results are summarized in Figure~\ref{fig:ablation}. 
Each row corresponds to a different task and dataset, while the columns correspond to the hyperparameters $t_{\text{cutoff}}$, the regularization strength $\lambda$, and the learning rate $k$. 
For each hyperparameter, we report the best LPIPS value obtained when fixing that parameter and varying the remaining hyperparameters. 

Regarding the selection of the cutoff timestep, we observe that switching to the denoised estimate at very early timesteps leads to significantly worse performance compared to switching later in the diffusion process. 
This can be explained by the fact that the denoised estimate is still very noisy during the early diffusion steps. 
Our framework leverages the robustness of the DINOv2 features and therefore benefits from using the denoised estimate only after a sufficiently large cutoff timestep. 
Interestingly, for ImageNet we found that the best performance is obtained without switching, indicating that the optimal cutoff strategy depends on the dataset and reconstruction difficulty.
\pagebreak

\subsection{Experimental Details}
We implement the inverse problem operators following the setup of
\cite{chung2023diffusion}.
For deblurring tasks, we follow \cite{zhang2025improving} and fix a single realization of
the degradation operator to ensure fair and consistent comparisons across methods.
All baseline results are obtained using the official implementations provided by
\cite{chung2023diffusion} and \cite{zhang2025improving}.  

For DPS, we use the dataset-specific hyperparameters reported in the original paper. For DAPS, we adopt the hyperparameters provided in the official codebase for latent
diffusion models. 

For PSNR and SSIM, we report results obtained by averaging the reconstructed images across
five independent runs of the solver and then computing the metrics on the averaged
reconstruction.

\subsection{Perception Distortion Tradeoff}\label{sec:perception_distortion}

We include qualitative examples for the task of $4\times$ super-resolution on the ImageNet dataset. 
In these examples, we observe that although the reconstructions obtained with \textsc{REPA} are perceptually more realistic and contain sharper details, the PSNR values do not necessarily increase. 
More generally, we observe that PSNR often favours smoother images, whereas perceptual metrics and visual inspection favour images with sharper details. 

\begin{figure*}[t]
    \centering
    \setlength{\tabcolsep}{2pt}
    \renewcommand{\arraystretch}{0.9}

    \begin{tabular}{cccc}
        \textbf{Measurement} & \textbf{Without REPA} & \textbf{With REPA} & \textbf{Ground Truth} \\

        \includegraphics[width=0.18\linewidth]{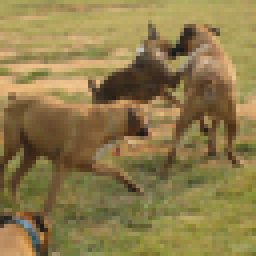} &
        \includegraphics[width=0.18\linewidth]{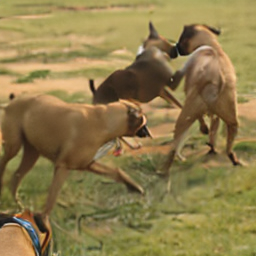} &
        \includegraphics[width=0.18\linewidth]{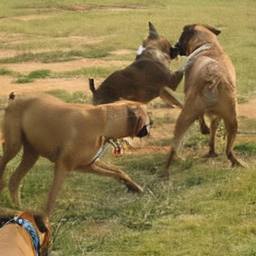} &
        \includegraphics[width=0.18\linewidth]{media/00040.png} \\

        & \footnotesize PSNR: 26.68 & \footnotesize PSNR: 26.01 & \\[4pt]

        \includegraphics[width=0.18\linewidth]{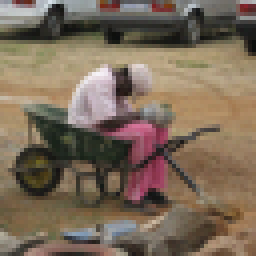} &
        \includegraphics[width=0.18\linewidth]{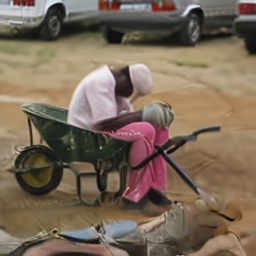} &
        \includegraphics[width=0.18\linewidth]{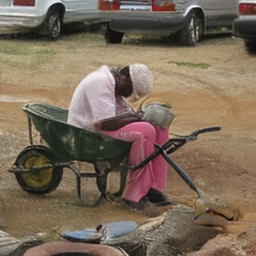} &
        \includegraphics[width=0.18\linewidth]{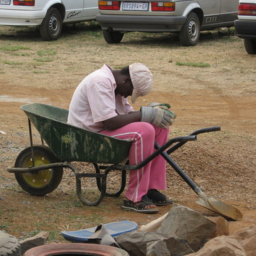} \\

        & \footnotesize PSNR: 26.03 & \footnotesize PSNR: 25.77 & \\

    \end{tabular}

    \caption{
Qualitative comparison for $4\times$ super-resolution on the ImageNet dataset.
While \textsc{REPA} produces sharper and more perceptually realistic reconstructions, these do not necessarily correspond to higher PSNR values, illustrating the perception--distortion trade-off.
}
    \label{fig:perception_distortion_sr}
\end{figure*}

\subsection{Enforcing Feature-Level DINOv2 Consistency without Representation Alignment}
\label{sec:feature_vs_repa}

In this ablation study, we compare REPA against a variant that enforces consistency directly in the DINOv2 feature-space, without involving the diffusion model representations. 
More specifically, instead of using the representation alignment loss in line~8 of Algorithm~\ref{alg:latent_dps_repa}, we replace it with the following update step:
\begin{equation}
    z_{t-1} \gets z_{t-1} + \lambda \nabla_{z_t} 
    \sum_{n=1}^{N} 
    \cos\!\left(
        c^{[n]}_{\text{proxy}},\,
        f_{\text{DINOv2}}^{[n]}\!\big(\mathcal{D}(\mathbb{E}[z_0 \mid z_t])\big)
    \right),
\end{equation}
Note that in this case we can only use $f_{\text{DINOv}2}(y)$ as the proxy term. While both objectives influence the sampling trajectory, the feature-only variant derives its guidance from the DINOv2 features of the decoded image. 
In contrast, REPA aligns the diffusion model’s internal representations with DINOv2 features directly at each timestep. As shown in Table~\ref{tab:feature_vs_repa}, the direct feature-space alignment yields consistently lower perceptual quality than the proposed representation-based formulation. 
for both experiments we have chosen the best hyperparameters using a small grid search as in \ref{sec:latent_dps} to ensure fairness in comparison. 







\begin{table}[h]
\centering
\caption{
Quantitative comparison between REPA and direct feature-space alignment.
}
\vspace{0.5em}
\begin{tabular}{lcc|cc}
\toprule
 & \multicolumn{2}{c|}{\textbf{Super-resolution ($\times 4$)}} & \multicolumn{2}{c}{\textbf{Gaussian deblurring}} \\
\textbf{Method} & \textbf{LPIPS} $\downarrow$ & \textbf{FID} $\downarrow$ & \textbf{LPIPS} $\downarrow$ & \textbf{FID} $\downarrow$ \\
\midrule
Feature-space alignment & 0.227 & 95.69 & 0.276 & 128.29 \\
Representation alignment (proposed) & 0.217 & 86.88 & 0.256 & 102.99 \\
\bottomrule
\end{tabular}
\label{tab:feature_vs_repa}
\end{table}

\pagebreak

\subsection{Representation Alignment on an Independently Pretrained Model}\label{app:mapping}

The effectiveness of \textsc{Repa} for inverse problems relies on the availability
of a mapping $g_{\phi}$ between the internal representations of the diffusion
model and the \text{DINOv2} feature space. Such a mapping arises naturally when
the diffusion model is pretrained with representation alignment. However, when
starting from an independently pretrained diffusion model, this mapping is not
readily available. We describe here how to obtain such a mapping in this setting.

A natural approach is to learn $g_{\phi}$ from pairs of diffusion and
\text{DINOv2} representations extracted from the same images. Specifically,
given a dataset $\{x_i\}_{i=1}^{n_0} \subset \mathbb{R}^k, \quad n_0 \in \mathbb{N}$ and a pretrained encoder
$f : \mathbb{R}^k \to \mathbb{R}^{N \times D_1}$, we train $g_{\phi}$ by maximizing
the \textsc{Repa} objective:
\begin{equation}
    \mathcal{L}_{\textsc{Repa}}(\phi)
    =
    \mathbb{E}_{x,\, t,\, \epsilon}
    \left[
    \frac{1}{N} \sum_{n=1}^{N}
    \cos\!\left(
    f^{[n]}(x),\,
    g^{[n]}_{\phi}\!\big(h_t\big)
    \right)
    \right],
\end{equation}
where the intermediate diffusion representation is given by
$
h_t = \textsc{DiffEnc}(x_t, t) \in \mathbb{R}^{N \times D_2}
$. Here, $f^{[n]}(x) \in \mathbb{R}^{D_1}$ denotes the corresponding patch-level
representation produced by the \text{DINOv2} encoder. Importantly, only the mapping $g_{\phi}$
is optimized, while both the diffusion model (through $\textsc{DiffEnc}$) and
the encoder $f$ remain frozen. As a result, learning $g_{\phi}$ is
computationally lightweight compared to full diffusion model training.

We consider two parameterizations of $g_{\phi}$. The first follows~\cite{yu2024repa}
and applies a token-wise MLP independently to each patch. The second parameterizes $g_{\phi}$ as a lightweight Transformer-style network
operating on patch-level representations. Given input tokens
$h_t \in \mathbb{R}^{N \times D_2}$, the model first applies a linear projection
to an intermediate dimension $D$. The resulting tokens are augmented with learned
positional embeddings and processed by a stack of $L$ Transformer encoder blocks.
Each block consists of a multi-head self-attention layer followed by a feed-forward
network, with residual connections. Finally, a linear projection maps the output tokens to the DINOv2 feature space,
yielding $g_{\phi}(h_t) \in \mathbb{R}^{N \times D_1}$. In our experiments, we use $D=768$, $L=2$, and 4 attention heads.

\begin{figure}[t]
    \centering
    \includegraphics[width=0.3\linewidth]{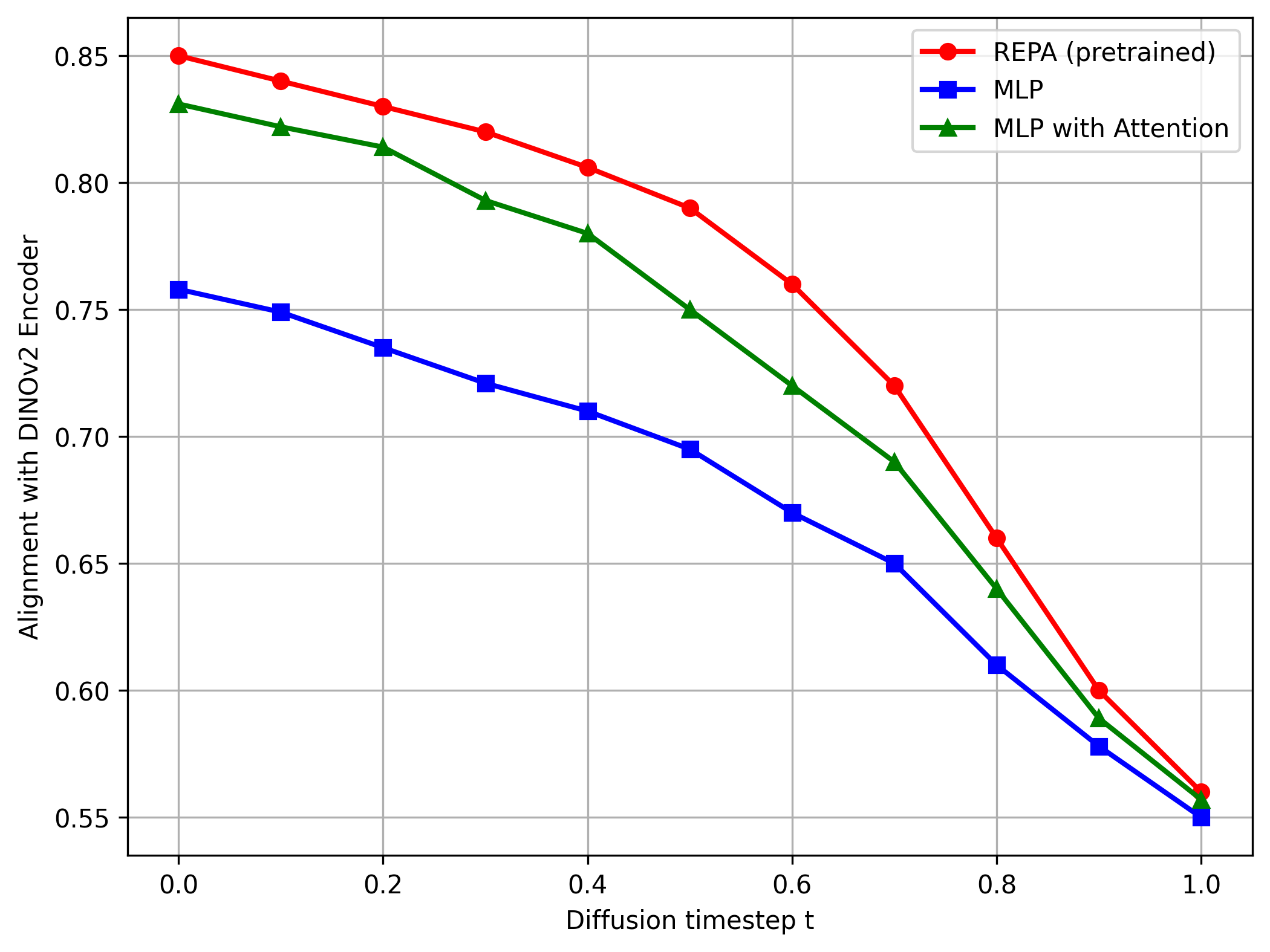}
    \caption{Alignment with the DINOv2 encoder across diffusion timesteps for
    different parameterizations of $g_{\phi}$. The red curve corresponds to a model
    pretrained with representation alignment, the blue curve to a token-wise MLP,
    and the green curve to the attention-based mapping.}
    \label{fig:alignment_curves}
\end{figure}

We train the mapping $g_{\phi}$ on the FFHQ dataset for 15k iterations using the
alignment objective described above. Training is performed with batch size 256 and
learning rate $10^{-4}$ using the AdamW optimizer. We present the resulting alignment curves in for different timesteps in Figure~\ref{fig:alignment_curves}. The corresponding training setup is summarized in Table~\ref{tab:mapping_cost}. We observe that the token-wise MLP is able to learn a meaningful mapping, but achieves lower alignment compared to both the attention-based parameterization and a model pretrained with representation alignment.

\begin{figure}[t]
    \centering
    \setlength{\tabcolsep}{2pt}
    \begin{tabular}{cccc}
        \includegraphics[width=0.2\linewidth]{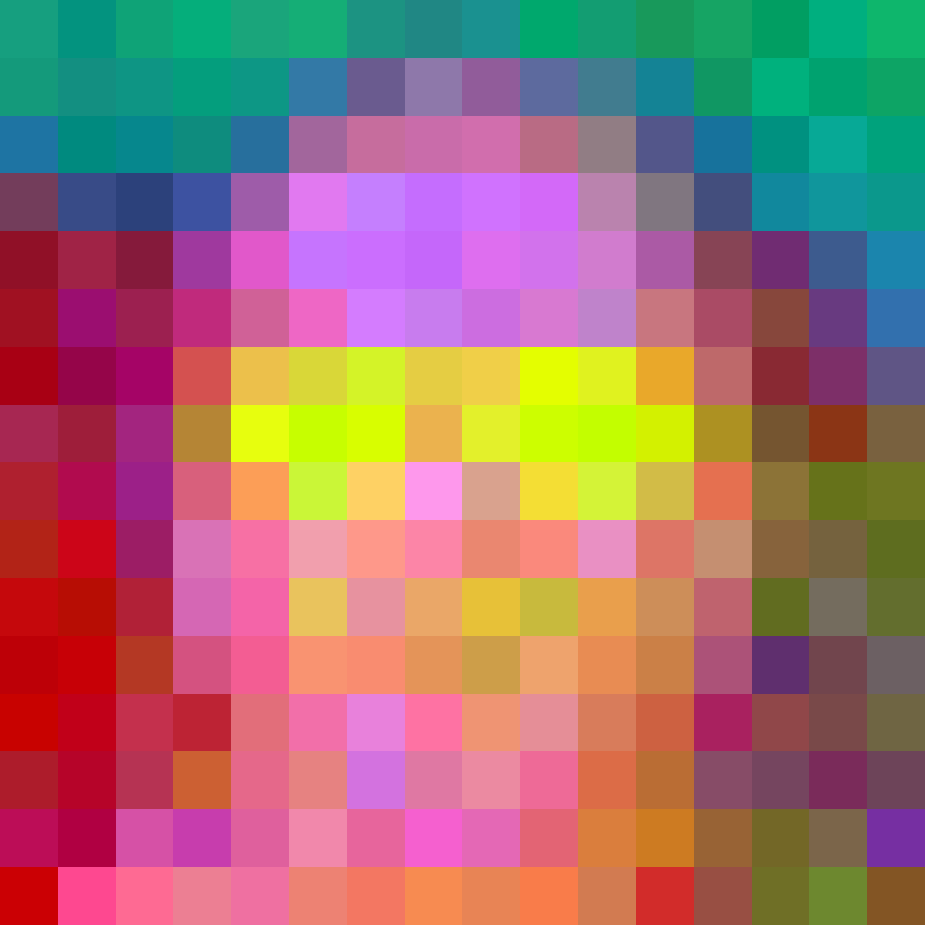} &
        \includegraphics[width=0.2\linewidth]{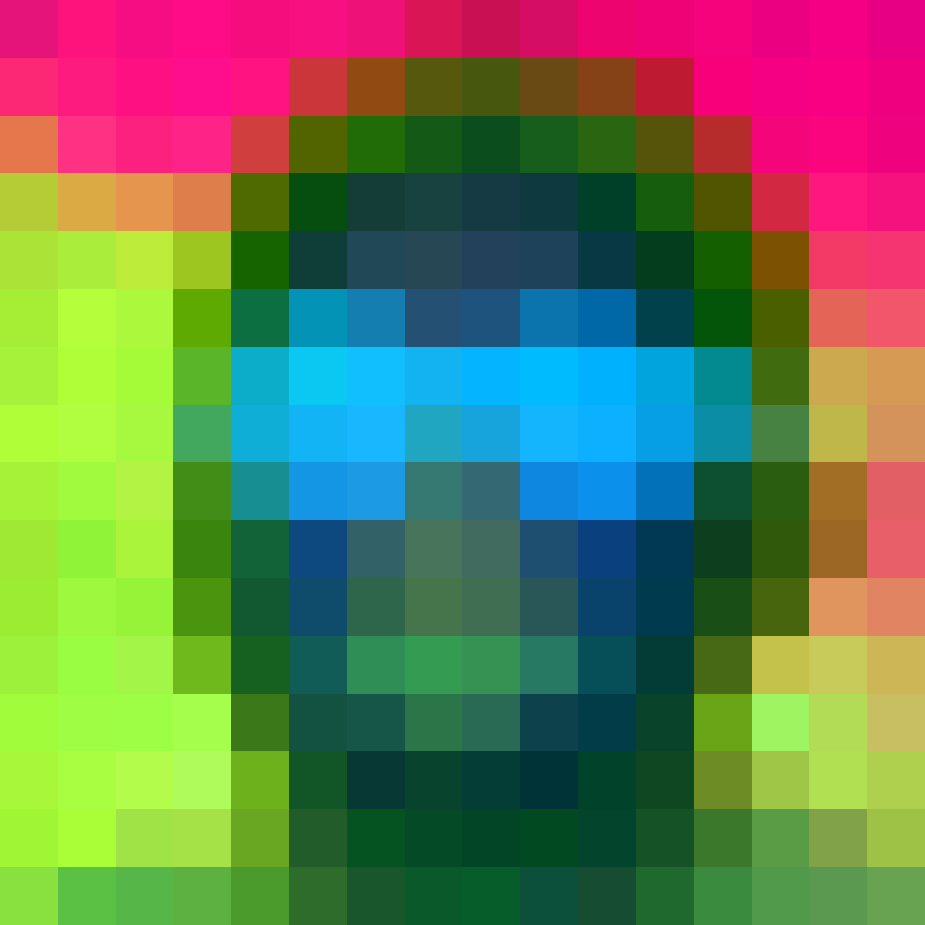} &
        \includegraphics[width=0.2\linewidth]{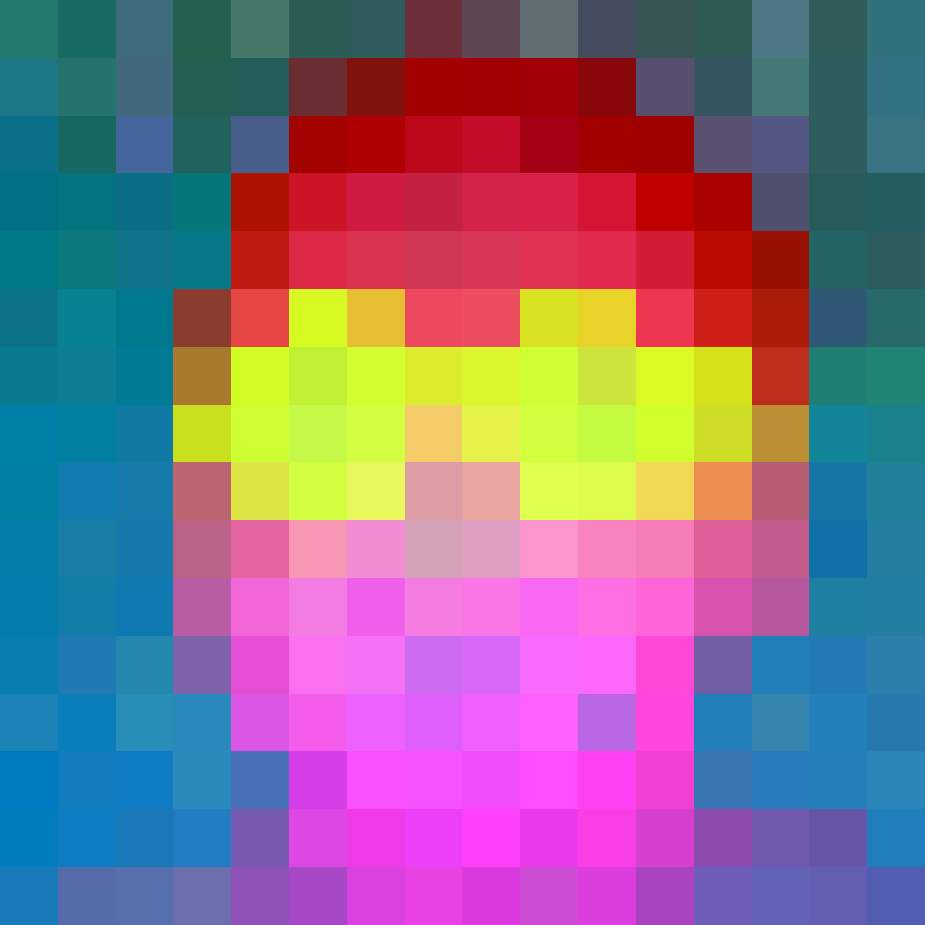} &
        \includegraphics[width=0.2\linewidth]{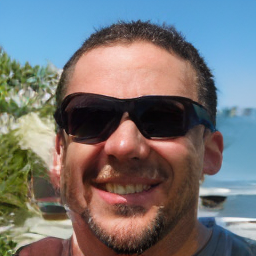} \\
    \end{tabular}
    \caption{
        PCA visualizations of patch-level representations.
        From left to right: representations from a standard diffusion model,
        representations from a diffusion model trained with \textsc{Repa}, \text{DINOv2} features, and the
        corresponding input image.
        }
        \label{fig:mapping_examples}
\end{figure}

To motivate the need for a more expressive mapping between independently
pretrained diffusion representations and the \text{DINOv2} feature space, we
visualize PCA feature maps similar to those in~\cite{yu2024repa}. As shown in
Figure~\ref{fig:mapping_examples}, representations extracted from a standard
diffusion model appear significantly noisier and less spatially coherent than
those produced by a \textsc{Repa}-pretrained model or \text{DINOv2}. The
observed structure suggests that recovering semantically aligned features
benefits from aggregating information across spatial locations.

\begin{table}[t]
\centering
\caption{Comparison of training cost across components.}
\begin{tabular}{lcccc}
\toprule
Model & Trainable parameters & Training steps & Batch size & Optimizer \\
\midrule
SiT full model & 136.9M & 100k & 256 & AdamW ($10^{-4}$) \\
MLP mapping $g_\phi$ & 7.3M & 15k & 256 & AdamW ($10^{-4}$) \\
Attention mapping $g_\phi$ & 15.6M & 15k & 256 & AdamW ($10^{-4}$) \\
\bottomrule
\end{tabular}
\label{tab:mapping_cost}
\end{table}



\pagebreak

\subsection{Results with an Independently Pretrained Model}\label{sec:independently_pretrained_results}
We apply our inference-time regularizer to inverse problems using the learned mapping.
Table~\ref{table:fint} reports the corresponding results, alongside those obtained
with a model pretrained with representation alignment. Across all inverse problems,
incorporating our regularizer consistently improves perceptual metrics (LPIPS and FID),
demonstrating that learning the mapping $g_{\phi}$ alone is sufficient to enable
effective use of \textsc{Repa} at inference time. More specifically, both learned mappings improve over the baseline Latent DPS (SiT),
with the attention-based parameterization consistently outperforming the token-wise
MLP across all tasks. While the MLP already yields gains, incorporating attention
further improves LPIPS and FID. These trends are
consistent with our theoretical analysis: performance is governed by the accuracy of
the learned mapping, as captured by the $\mathrm{MisREPA}$ error. In this view, the
attention-based mapping achieves lower $\mathrm{MisREPA}$, leading to improved alignment
and downstream performance, while models pretrained with representation alignment
correspond to the smallest mismatch and therefore achieve the strongest overall results.

\FloatBarrier
\begin{table}[H]
\centering
\small
\caption{Performance comparison on FFHQ ($\sigma = 0.05$).}
\setlength{\tabcolsep}{6pt}
\renewcommand{\arraystretch}{1.15}
\begin{tabular}{llcccc}
\toprule
\textbf{Task} & \textbf{Method} & \textbf{LPIPS}$\downarrow$ & \textbf{FID}$\downarrow$ & \textbf{PSNR}$\uparrow$ & \textbf{SSIM}$\uparrow$ \\
\midrule

\multirow{5}{*}{$4\times$ SR}
& Latent DPS (REPA-backbone) 
& 0.188 & 56.69 & 28.99 & 0.814 \\

& \hspace{2mm}$\hookrightarrow$ + REPA regularization 
& \textbf{0.176} & \textbf{51.27} & \textbf{29.12} & \textbf{0.819} \\

\cmidrule(lr){2-6}

& Latent DPS (SiT backbone) 
& 0.190 & 61.36 & 28.92 & 0.805 \\

& \hspace{2mm}$\hookrightarrow$ + REPA (MLP mapping) 
& 0.184 & 56.17 & 28.95 & 0.809 \\

& \hspace{2mm}$\hookrightarrow$ + REPA (Attention mapping) 
& \textbf{0.177} & \textbf{54.21} & \textbf{28.97} & \textbf{0.814} \\

\midrule

\multirow{5}{*}{Gaussian Deblur}
& Latent DPS (REPA-backbone) 
& 0.192 & 60.65 & 28.15 & 0.783 \\

& \hspace{2mm}$\hookrightarrow$ + REPA regularization 
& \textbf{0.186} & \textbf{55.53} & \textbf{28.21} & \textbf{0.787} \\

\cmidrule(lr){2-6}

& Latent DPS (SiT backbone) 
& 0.193 & 60.72 & 28.02 & 0.782 \\

& \hspace{2mm}$\hookrightarrow$ + REPA (MLP mapping) 
& 0.191 & 58.92 & 28.09 & 0.770 \\

& \hspace{2mm}$\hookrightarrow$ + REPA (Attention mapping) 
& \textbf{0.187} & \textbf{58.56} & \textbf{28.17} & \textbf{0.778} \\

\midrule

\multirow{5}{*}{Motion Deblur}
& Latent DPS (REPA-backbone) 
& 0.170 & 52.14 & \textbf{27.20} & \textbf{0.773} \\

& \hspace{2mm}$\hookrightarrow$ + REPA regularization 
& \textbf{0.165} & \textbf{47.18} & 27.16 & 0.772 \\

\cmidrule(lr){2-6}

& Latent DPS (SiT backbone) 
& 0.171 & 54.01 & \textbf{28.02} & \textbf{0.779} \\

& \hspace{2mm}$\hookrightarrow$ + REPA (MLP mapping) 
& 0.171 & 52.93 & 27.73 & 0.769 \\

& \hspace{2mm}$\hookrightarrow$ + REPA (Attention mapping) 
& \textbf{0.167} & \textbf{49.94} & 27.89 & 0.775 \\

\bottomrule
\end{tabular}
\label{table:fint}
\end{table}
\FloatBarrier




\subsection{Additional Qualitative Results}\label{sec:qualitative}

\begin{center}
\makebox[\textwidth][c]{%
\begin{minipage}{1.05\textwidth}
\centering
\setlength{\tabcolsep}{1pt}
\renewcommand{\arraystretch}{0.5}

\begin{tabular}{ccccc}
    \textbf{Measurement} & \textbf{DPS} & \textbf{Without REPA} & \textbf{With REPA} & \textbf{Ground Truth} \\

    \includegraphics[width=0.2\textwidth]{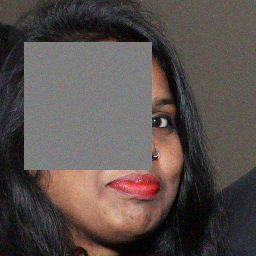} &
    \includegraphics[width=0.2\textwidth]{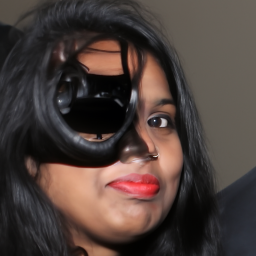} &
    \includegraphics[width=0.2\textwidth]{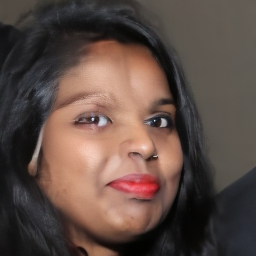} &
    \includegraphics[width=0.2\textwidth]{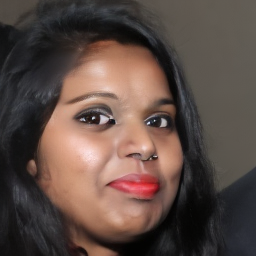} &
    \includegraphics[width=0.2\textwidth]{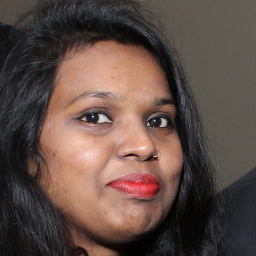} \\

    \includegraphics[width=0.2\textwidth]{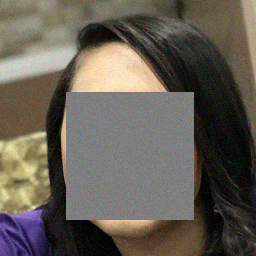} &
    \includegraphics[width=0.2\textwidth]{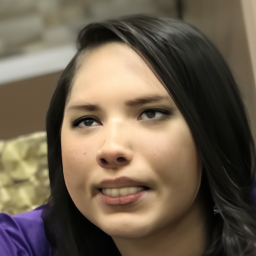} &
    \includegraphics[width=0.2\textwidth]{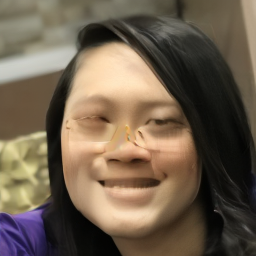} &
    \includegraphics[width=0.2\textwidth]{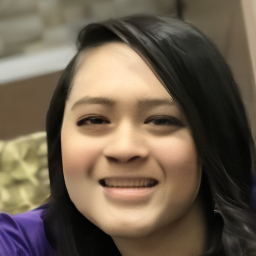} &
    \includegraphics[width=0.2\textwidth]{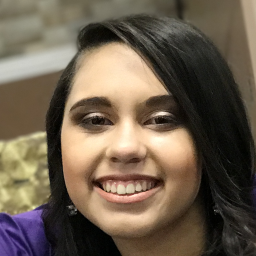} \\

    \includegraphics[width=0.2\textwidth]{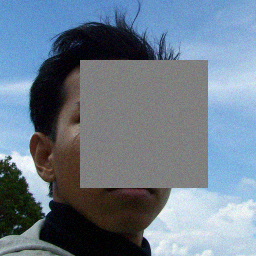} &
    \includegraphics[width=0.2\textwidth]{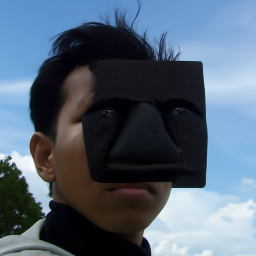} &
    \includegraphics[width=0.2\textwidth]{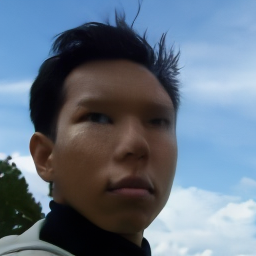} &
    \includegraphics[width=0.2\textwidth]{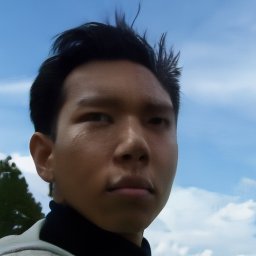} &
    \includegraphics[width=0.2\textwidth]{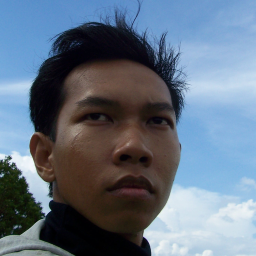} \\

    \includegraphics[width=0.2\textwidth]{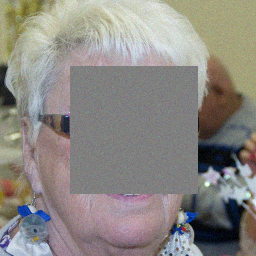} &
    \includegraphics[width=0.2\textwidth]{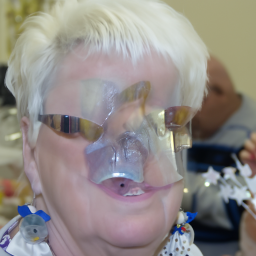} &
    \includegraphics[width=0.2\textwidth]{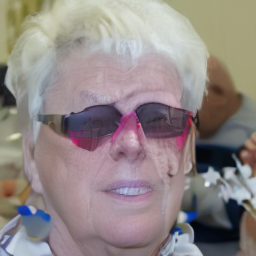} &
    \includegraphics[width=0.2\textwidth]{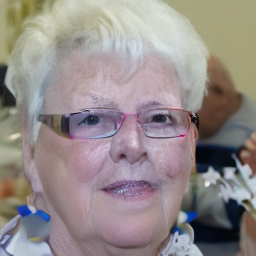} &
    \includegraphics[width=0.2\textwidth]{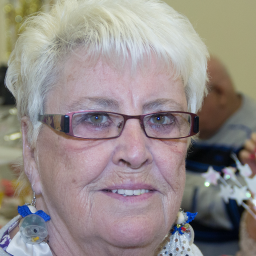} \\

    \includegraphics[width=0.2\textwidth]{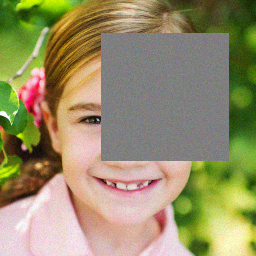} &
    \includegraphics[width=0.2\textwidth]{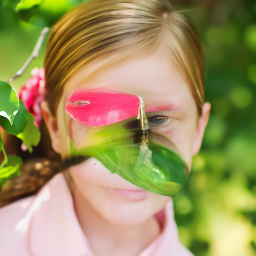} &
    \includegraphics[width=0.2\textwidth]{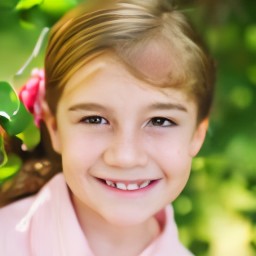} &
    \includegraphics[width=0.2\textwidth]{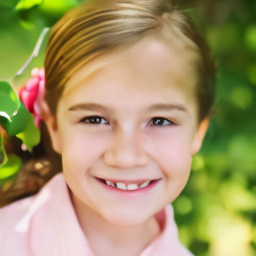} &
    \includegraphics[width=0.2\textwidth]{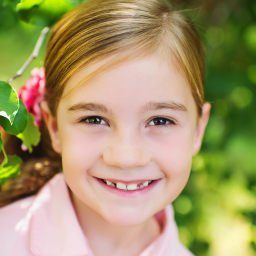} \\
\end{tabular}

\captionof{figure}{
Qualitative comparison for box inpainting on the FFHQ dataset.
Each row shows, from left to right: the measurement, DPS, the baseline latent solver,
its REPA-enhanced variant, and the ground truth.
}
\label{fig:deblur_grid}
\end{minipage}%
}
\end{center}

\begin{figure*}[t]
\makebox[\textwidth][l]{%
\hspace*{-0.06\textwidth}%
\begin{minipage}{1.12\textwidth}
\centering
\setlength{\tabcolsep}{1pt}
\renewcommand{\arraystretch}{0.5}

    \begin{tabular}{ccccc}
        \textbf{Measurement} & \textbf{Latent DAPS} & \textbf{Without REPA} & \textbf{With REPA} & \textbf{Ground Truth} \\

        \includegraphics[width=0.2\linewidth]{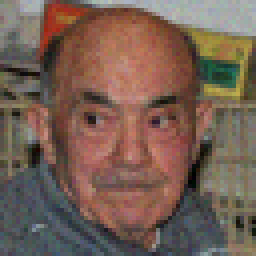} &
        \includegraphics[width=0.2\linewidth]{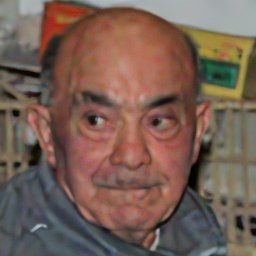} &
        \includegraphics[width=0.2\linewidth]{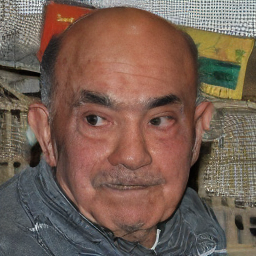} &
        \includegraphics[width=0.2\linewidth]{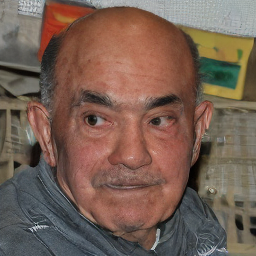} &
        \includegraphics[width=0.2\linewidth]{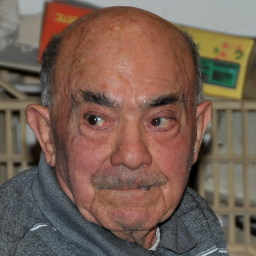} \\

        \includegraphics[width=0.2\linewidth]{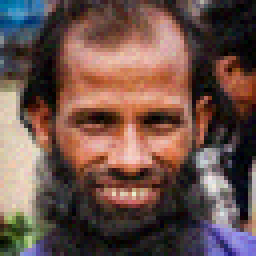} &
        \includegraphics[width=0.2\linewidth]{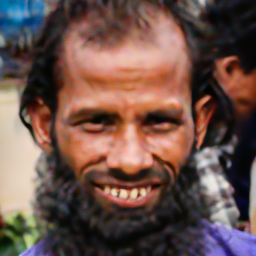} &
        \includegraphics[width=0.2\linewidth]{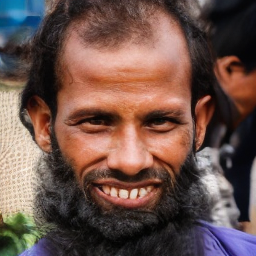} &
        \includegraphics[width=0.2\linewidth]{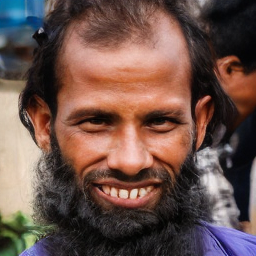} &
        \includegraphics[width=0.2\linewidth]{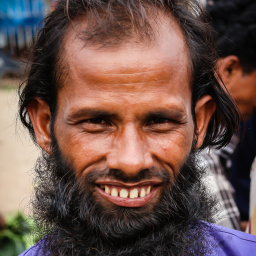} \\

        \includegraphics[width=0.2\linewidth]{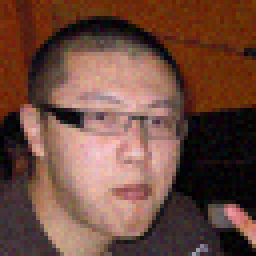} &
        \includegraphics[width=0.2\linewidth]{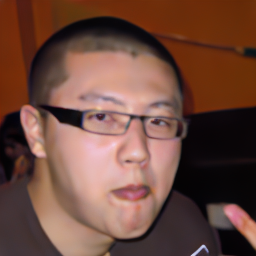} &
        \includegraphics[width=0.2\linewidth]{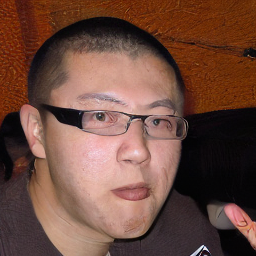} &
        \includegraphics[width=0.2\linewidth]{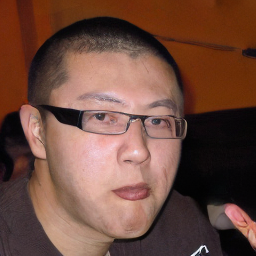} &
        \includegraphics[width=0.2\linewidth]{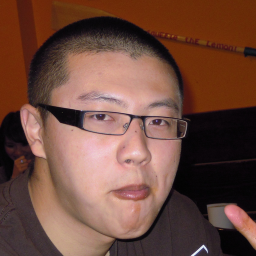} \\

        \includegraphics[width=0.2\linewidth]{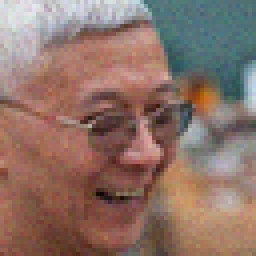} &
        \includegraphics[width=0.2\linewidth]{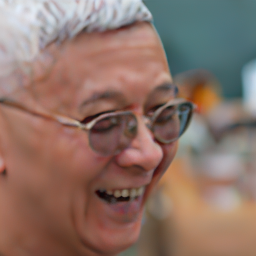} &
        \includegraphics[width=0.2\linewidth]{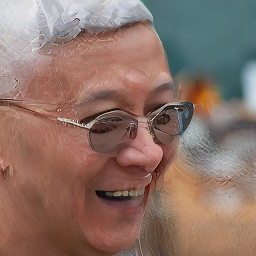} &
        \includegraphics[width=0.2\linewidth]{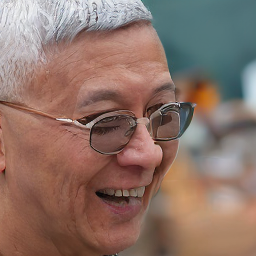} &
        \includegraphics[width=0.2\linewidth]{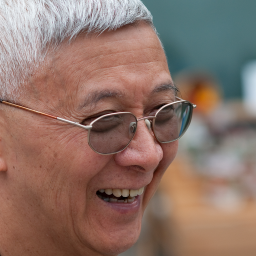} \\

        \includegraphics[width=0.2\linewidth]{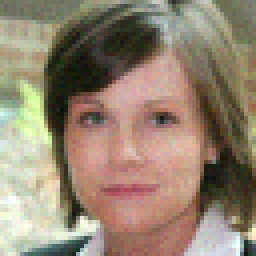} &
        \includegraphics[width=0.2\linewidth]{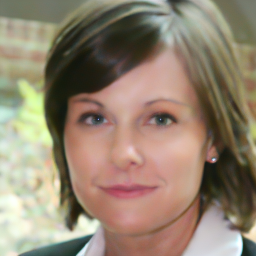} &
        \includegraphics[width=0.2\linewidth]{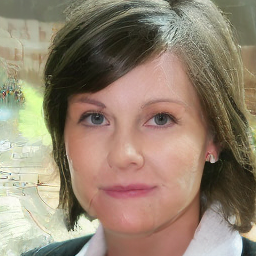} &
        \includegraphics[width=0.2\linewidth]{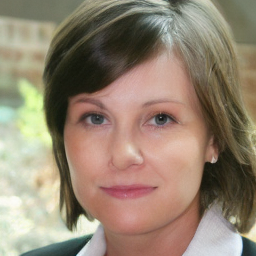} &
        \includegraphics[width=0.2\linewidth]{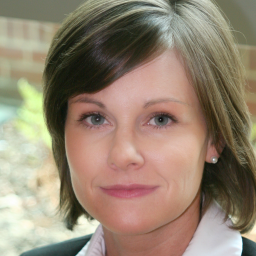} \\

    \end{tabular}

    \caption{
    Qualitative comparison for $4\times$ super-resolution on the FFHQ dataset.
    Each row shows (from left to right): the measurement, the baseline method (Latent DAPS), the baseline latent solver (Latent DPS or ReSample),
    its REPA-enhanced variant, and the ground truth.
    The first two rows correspond to ReSample, while the last three correspond to Latent DPS.
    }
    \label{fig:deblur_grid}
\end{minipage}
}
\end{figure*}

\begin{figure*}[t]
\makebox[\textwidth][l]{%
\hspace*{-0.06\textwidth}%
\begin{minipage}{1.12\textwidth}
\centering
\setlength{\tabcolsep}{1pt}
\renewcommand{\arraystretch}{0.5}

    \begin{tabular}{ccccc}
        \textbf{Measurement} & \textbf{DPS} & \textbf{Without REPA} & \textbf{With Repa} & \textbf{Ground Truth} \\

        \includegraphics[width=0.2\linewidth]{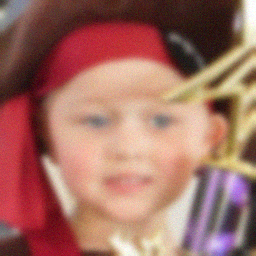} &
        \includegraphics[width=0.2\linewidth]{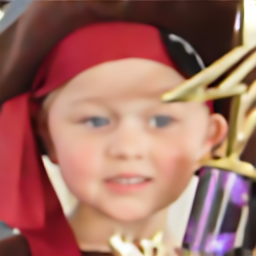} &
        \includegraphics[width=0.2\linewidth]{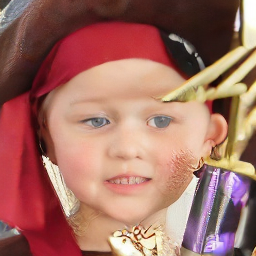} &
        \includegraphics[width=0.2\linewidth]{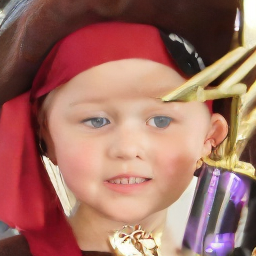} &
        \includegraphics[width=0.2\linewidth]{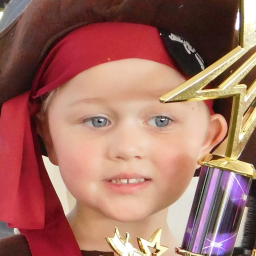} \\

        \includegraphics[width=0.2\linewidth]{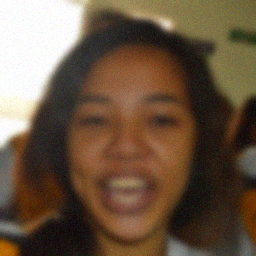} &
        \includegraphics[width=0.2\linewidth]{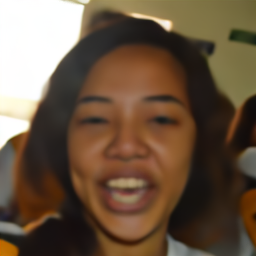} &
        \includegraphics[width=0.2\linewidth]{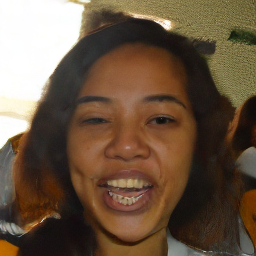} &
        \includegraphics[width=0.2\linewidth]{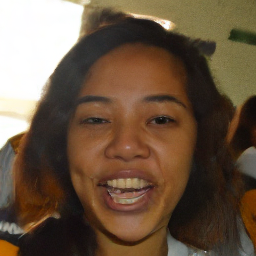} &
        \includegraphics[width=0.2\linewidth]{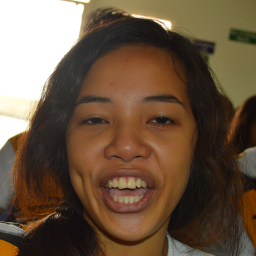} \\

        \includegraphics[width=0.2\linewidth]{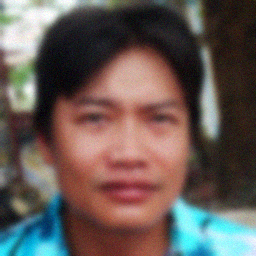} &
        \includegraphics[width=0.2\linewidth]{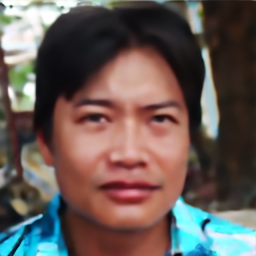} &
        \includegraphics[width=0.2\linewidth]{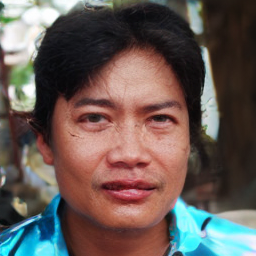} &
        \includegraphics[width=0.2\linewidth]{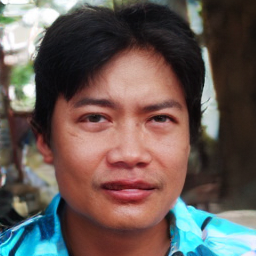} &
        \includegraphics[width=0.2\linewidth]{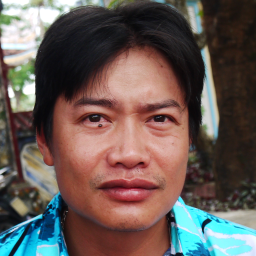} \\

        \includegraphics[width=0.2\linewidth]{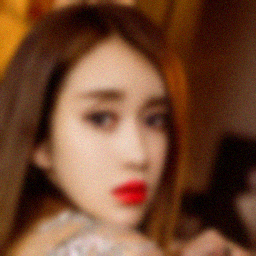} &
        \includegraphics[width=0.2\linewidth]{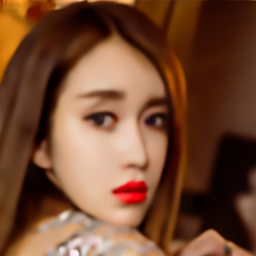} &
        \includegraphics[width=0.2\linewidth]{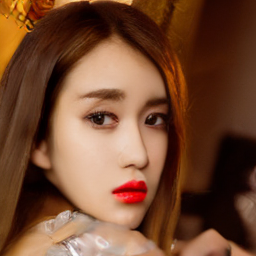} &
        \includegraphics[width=0.2\linewidth]{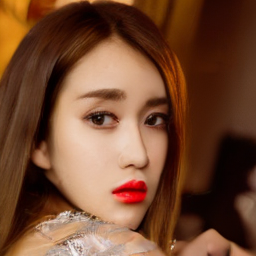} &
        \includegraphics[width=0.2\linewidth]{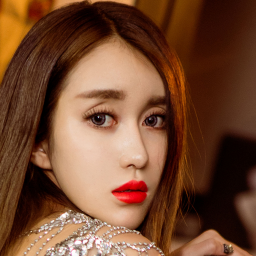} \\

        \includegraphics[width=0.2\linewidth]{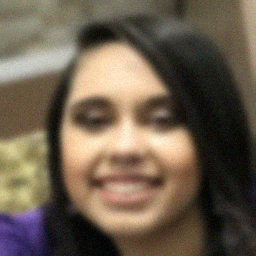} &
        \includegraphics[width=0.2\linewidth]{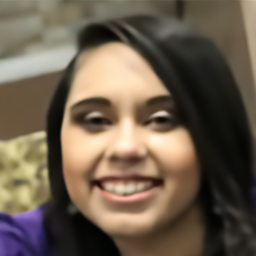} &
        \includegraphics[width=0.2\linewidth]{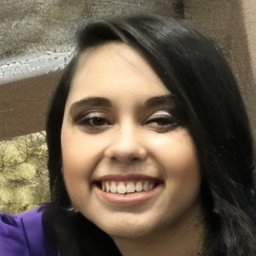} &
        \includegraphics[width=0.2\linewidth]{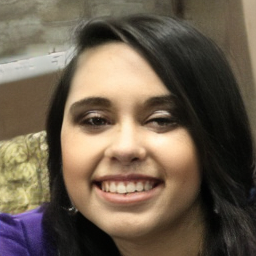} &
        \includegraphics[width=0.2\linewidth]{media/00076_ffhq.png} \\

    \end{tabular}

    \caption{
    Qualitative comparison for Gaussian Deblurring on the FFHQ dataset.
    Each row shows (from left to right): the measurement, the baseline method (Latent DAPS), the baseline latent solver (Latent DPS or ReSample),
    its REPA-enhanced variant, and the ground truth.
    The first two rows correspond to ReSample, while the last three correspond to Latent DPS.
    }
    \label{fig:deblur_grid}
\end{minipage}
}
\end{figure*}

\begin{figure*}[t]
\makebox[\textwidth][l]{%
\hspace*{-0.06\textwidth}%
\begin{minipage}{1.12\textwidth}
\centering
\setlength{\tabcolsep}{1pt}
\renewcommand{\arraystretch}{0.5}

    \begin{tabular}{ccccc}
        \textbf{Measurement} & \textbf{DPS} & \textbf{Without REPA} & \textbf{With REPA} & \textbf{Ground Truth} \\

        \includegraphics[width=0.2\linewidth]{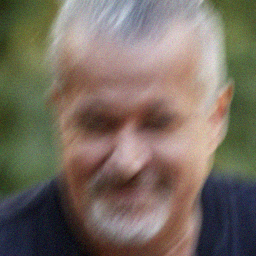} &
        \includegraphics[width=0.2\linewidth]{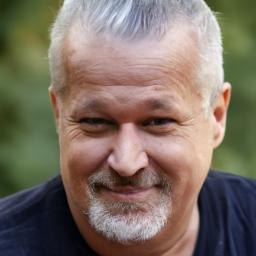} &
        \includegraphics[width=0.2\linewidth]{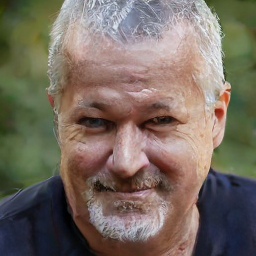} &
        \includegraphics[width=0.2\linewidth]{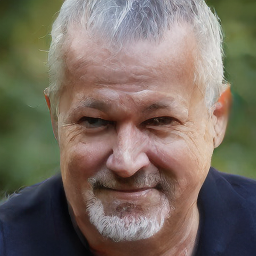} &
        \includegraphics[width=0.2\linewidth]{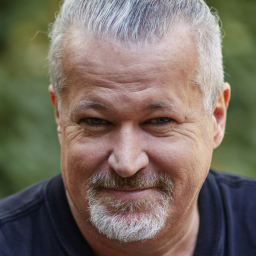} \\

        \includegraphics[width=0.2\linewidth]{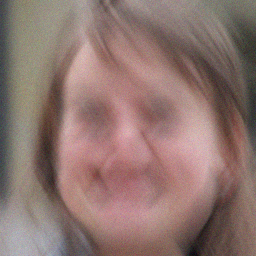} &
        \includegraphics[width=0.2\linewidth]{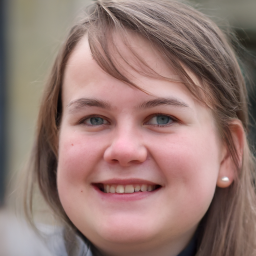} &
        \includegraphics[width=0.2\linewidth]{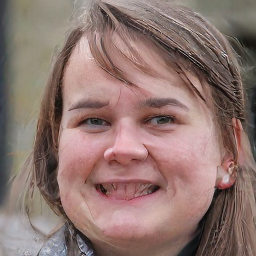} &
        \includegraphics[width=0.2\linewidth]{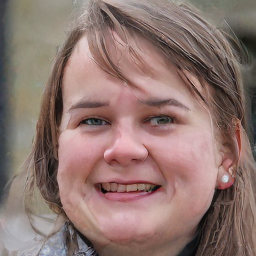} &
        \includegraphics[width=0.2\linewidth]{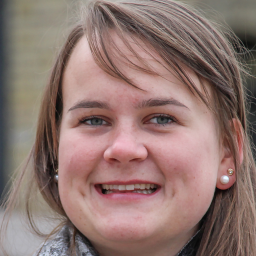} \\

        \includegraphics[width=0.2\linewidth]{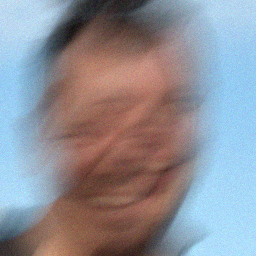} &
        \includegraphics[width=0.2\linewidth]{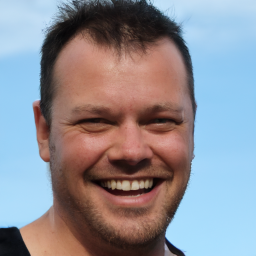} &
        \includegraphics[width=0.2\linewidth]{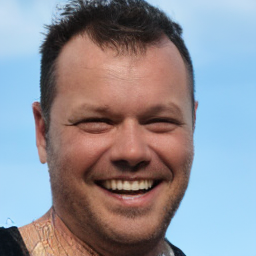} &
        \includegraphics[width=0.2\linewidth]{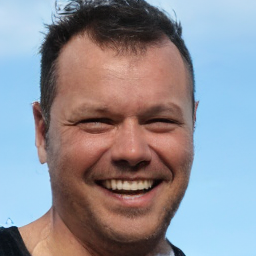} &
        \includegraphics[width=0.2\linewidth]{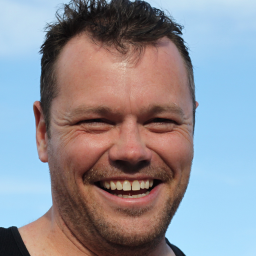} \\

        \includegraphics[width=0.2\linewidth]{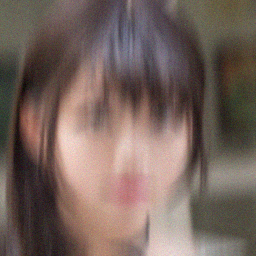} &
        \includegraphics[width=0.2\linewidth]{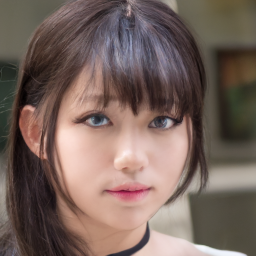} &
        \includegraphics[width=0.2\linewidth]{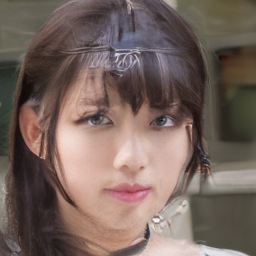} &
        \includegraphics[width=0.2\linewidth]{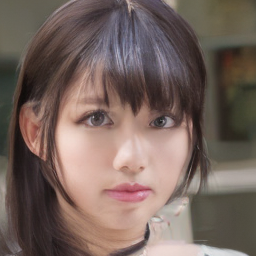} &
        \includegraphics[width=0.2\linewidth]{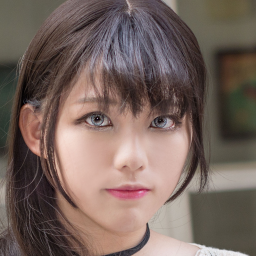} \\

        \includegraphics[width=0.2\linewidth]{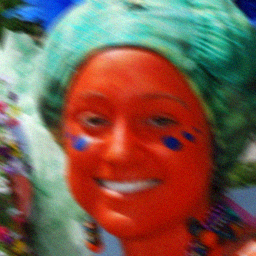} &
        \includegraphics[width=0.2\linewidth]{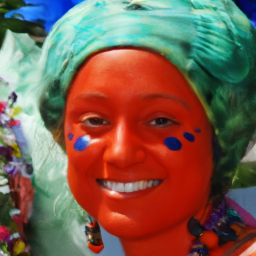} &
        \includegraphics[width=0.2\linewidth]{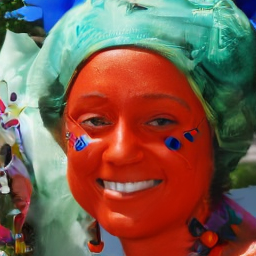} &
        \includegraphics[width=0.2\linewidth]{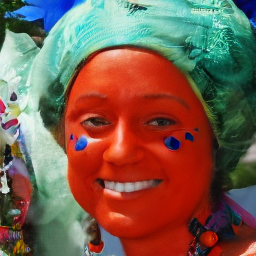} &
        \includegraphics[width=0.2\linewidth]{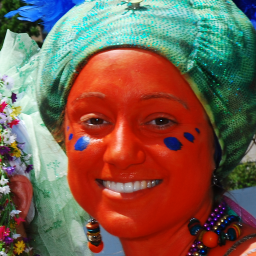} \\

    \end{tabular}

    \caption{
    Qualitative comparison for Motion Deblurring on the FFHQ dataset.
    Each row shows (from left to right): the measurement, the baseline method (DPS), the baseline latent solver (Latent DPS or ReSample),
    its REPA-enhanced variant, and the ground truth.
    The first two rows correspond to ReSample, while the last three correspond to Latent DPS.
    }
    \label{fig:deblur_grid}
\end{minipage}
}
\end{figure*}

\begin{figure*}[t]
\makebox[\textwidth][l]{%
\hspace*{-0.06\textwidth}%
\begin{minipage}{1.12\textwidth}
\centering
\setlength{\tabcolsep}{1pt}
\renewcommand{\arraystretch}{0.5}

    \begin{tabular}{ccccc}
        \textbf{Measurement} & \textbf{DPS} & \textbf{Without REPA} & \textbf{With REPA} & \textbf{Ground Truth} \\

        \includegraphics[width=0.2\linewidth]{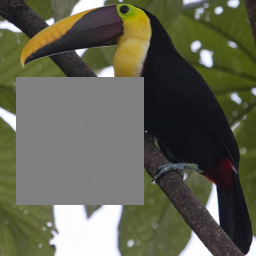} &
        \includegraphics[width=0.2\linewidth]{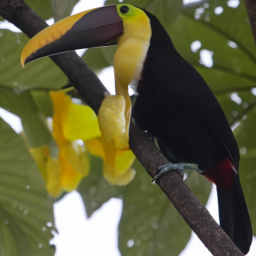} &
        \includegraphics[width=0.2\linewidth]{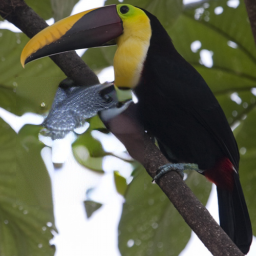} &
        \includegraphics[width=0.2\linewidth]{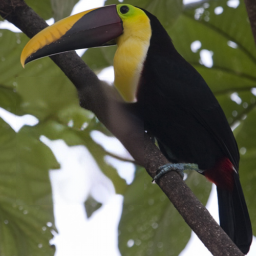} &
        \includegraphics[width=0.2\linewidth]{media/00093.png} \\

        \includegraphics[width=0.2\linewidth]{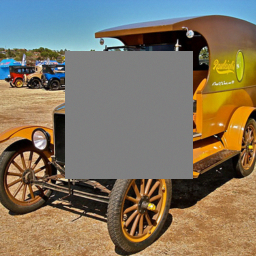} &
        \includegraphics[width=0.2\linewidth]{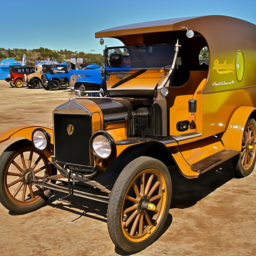} &
        \includegraphics[width=0.2\linewidth]{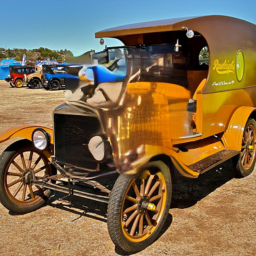} &
        \includegraphics[width=0.2\linewidth]{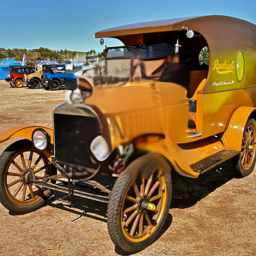} &
        \includegraphics[width=0.2\linewidth]{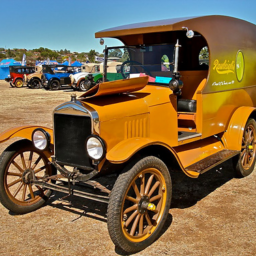} \\

        \includegraphics[width=0.2\linewidth]{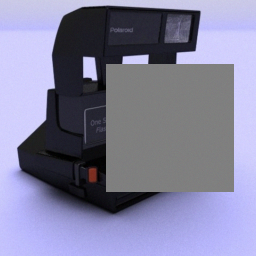} &
        \includegraphics[width=0.2\linewidth]{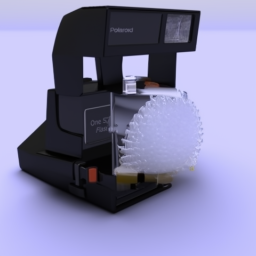} &
        \includegraphics[width=0.2\linewidth]{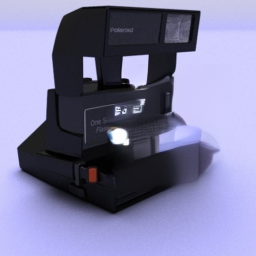} &
        \includegraphics[width=0.2\linewidth]{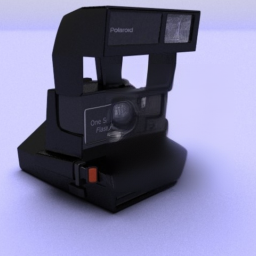} &
        \includegraphics[width=0.2\linewidth]{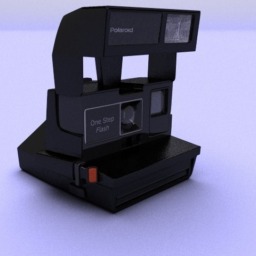} \\

        \includegraphics[width=0.2\linewidth]{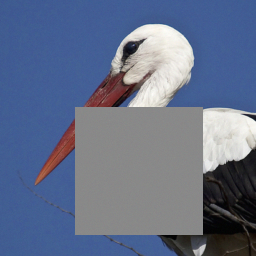} &
        \includegraphics[width=0.2\linewidth]{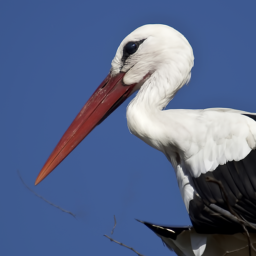} &
        \includegraphics[width=0.2\linewidth]{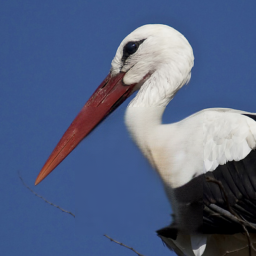} &
        \includegraphics[width=0.2\linewidth]{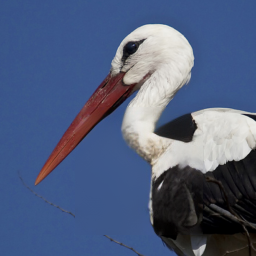} &
        \includegraphics[width=0.2\linewidth]{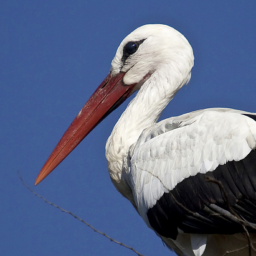} \\

        \includegraphics[width=0.2\linewidth]{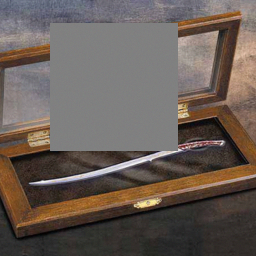} &
        \includegraphics[width=0.2\linewidth]{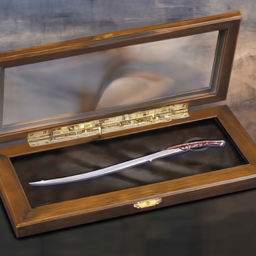} &
        \includegraphics[width=0.2\linewidth]{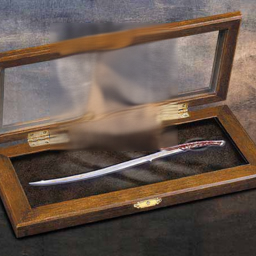} &
        \includegraphics[width=0.2\linewidth]{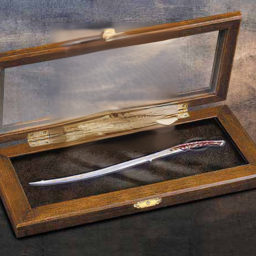} &
        \includegraphics[width=0.2\linewidth]{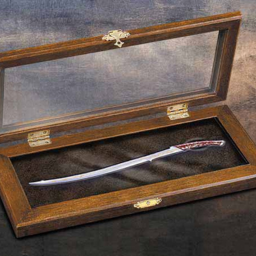} \\

    \end{tabular}

    \caption{
    Qualitative comparison for box inpainting on the ImageNet dataset.
    Each row shows (from left to right): the measurement, the baseline method (DPS), the baseline latent solver
    (Latent DPS or ReSample), its REPA-enhanced variant, and the ground truth.
    The first three rows correspond to ReSample, while the last two correspond to Latent DPS.
    }
    \label{fig:deblur_grid}
\end{minipage}
}
\end{figure*}

\begin{figure*}[t]
\makebox[\textwidth][l]{%
\hspace*{-0.06\textwidth}%
\begin{minipage}{1.12\textwidth}
\centering
\setlength{\tabcolsep}{1pt}
\renewcommand{\arraystretch}{0.5}

    \begin{tabular}{ccccc}
        \textbf{Measurement} & \textbf{Latent DAPS} & \textbf{Without REPA} & \textbf{With REPA} & \textbf{Ground Truth} \\

        \includegraphics[width=0.2\linewidth]{media/00040_sr.png} &
        \includegraphics[width=0.2\linewidth]{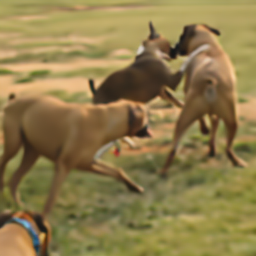} &
        \includegraphics[width=0.2\linewidth]{media/040_resample_sr.png} &
        \includegraphics[width=0.2\linewidth]{media/00040_resample_repa_sr.png} &
        \includegraphics[width=0.2\linewidth]{media/00040.png} \\

        \includegraphics[width=0.2\linewidth]{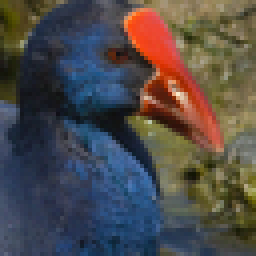} &
        \includegraphics[width=0.2\linewidth]{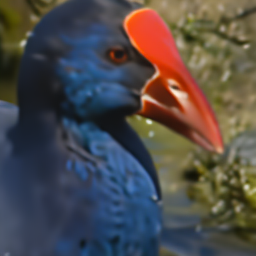} &
        \includegraphics[width=0.2\linewidth]{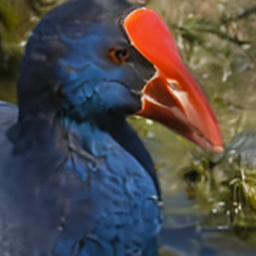} &
        \includegraphics[width=0.2\linewidth]{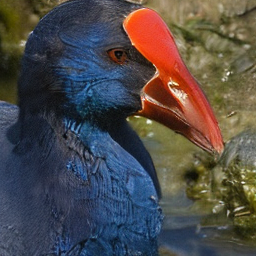} &
        \includegraphics[width=0.2\linewidth]{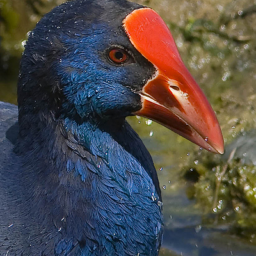} \\

        \includegraphics[width=0.2\linewidth]{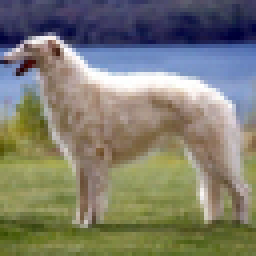} &
        \includegraphics[width=0.2\linewidth]{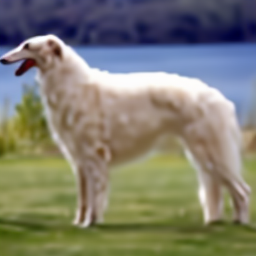} &
        \includegraphics[width=0.2\linewidth]{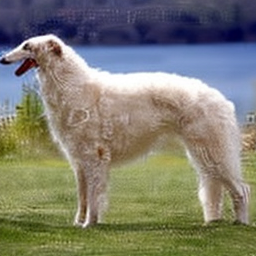} &
        \includegraphics[width=0.2\linewidth]{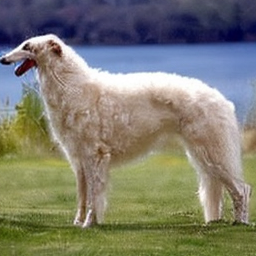} &
        \includegraphics[width=0.2\linewidth]{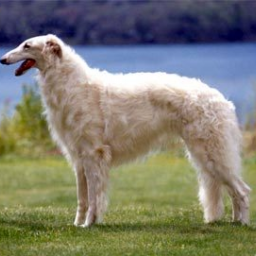} \\

        \includegraphics[width=0.2\linewidth]{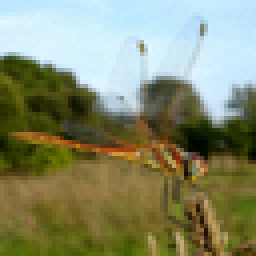} &
        \includegraphics[width=0.2\linewidth]{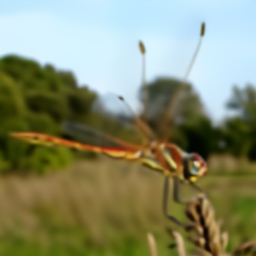} &
        \includegraphics[width=0.2\linewidth]{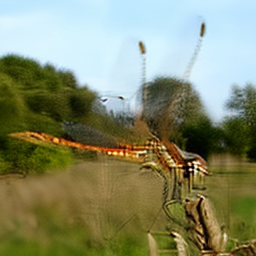} &
        \includegraphics[width=0.2\linewidth]{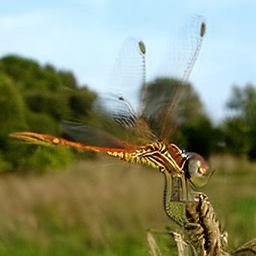} &
        \includegraphics[width=0.2\linewidth]{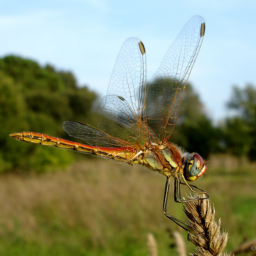} \\

        \includegraphics[width=0.2\linewidth]{media/00035_noisy.png} &
        \includegraphics[width=0.2\linewidth]{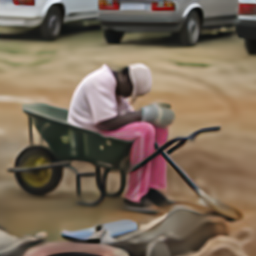} &
        \includegraphics[width=0.2\linewidth]{media/00035_snr.png} &
        \includegraphics[width=0.2\linewidth]{media/00035_snr_y.png} &
        \includegraphics[width=0.2\linewidth]{media/00035.png} \\

    \end{tabular}

    \caption{
    Qualitative comparison for Super resolution on the ImageNet dataset.
    Each row shows (from left to right): the measurement, the baseline method (Latent DAPS), the baseline latent solver
    (Latent DPS or ReSample), its REPA-enhanced variant, and the ground truth.
    The first three rows correspond to ReSample, while the last two correspond to Latent DPS.
    }
    \label{fig:deblur_grid}
\end{minipage}
}
\end{figure*}

\begin{figure*}[t]
\makebox[\textwidth][l]{%
\hspace*{-0.06\textwidth}%
\begin{minipage}{1.12\textwidth}
\centering
\setlength{\tabcolsep}{1pt}
\renewcommand{\arraystretch}{0.5}

    \begin{tabular}{ccccc}
        \textbf{Measurement} & \textbf{DPS} & \textbf{Without REPA} & \textbf{With REPA} & \textbf{Ground Truth} \\

        \includegraphics[width=0.2\linewidth]{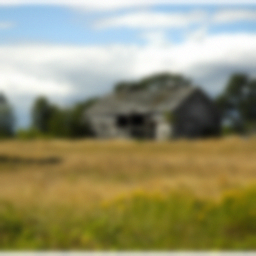} &
        \includegraphics[width=0.2\linewidth]{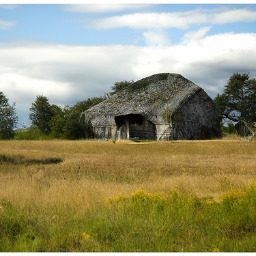} &
        \includegraphics[width=0.2\linewidth]{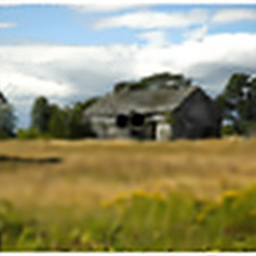} &
        \includegraphics[width=0.2\linewidth]{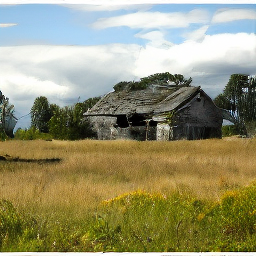} &
        \includegraphics[width=0.2\linewidth]{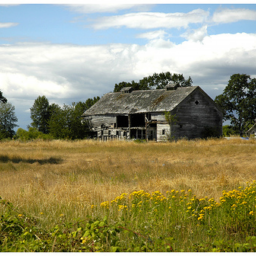} \\

        \includegraphics[width=0.2\linewidth]{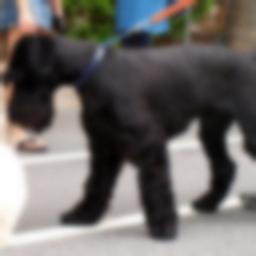} &
        \includegraphics[width=0.2\linewidth]{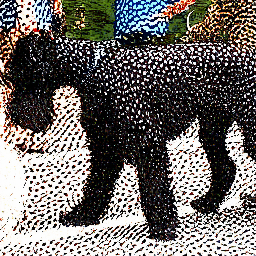} &
        \includegraphics[width=0.2\linewidth]{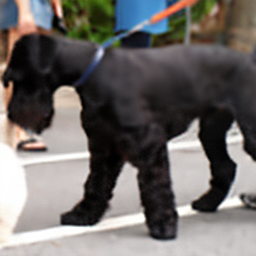} &
        \includegraphics[width=0.2\linewidth]{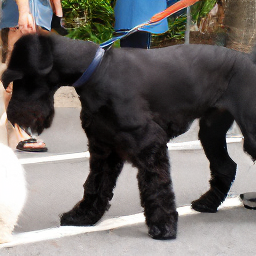} &
        \includegraphics[width=0.2\linewidth]{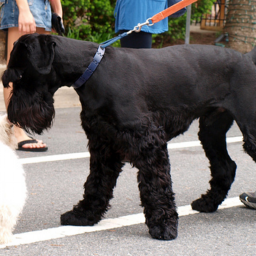} \\

        \includegraphics[width=0.2\linewidth]{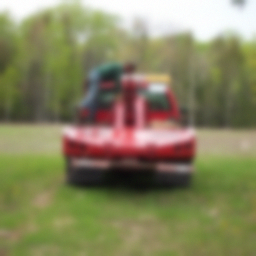} &
        \includegraphics[width=0.2\linewidth]{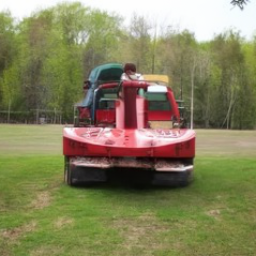} &
        \includegraphics[width=0.2\linewidth]{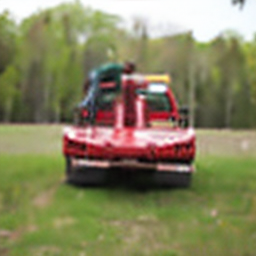} &
        \includegraphics[width=0.2\linewidth]{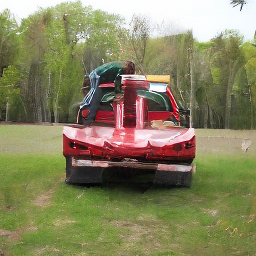} &
        \includegraphics[width=0.2\linewidth]{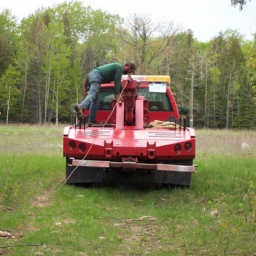} \\

        \includegraphics[width=0.2\linewidth]{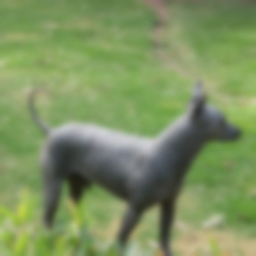} &
        \includegraphics[width=0.2\linewidth]{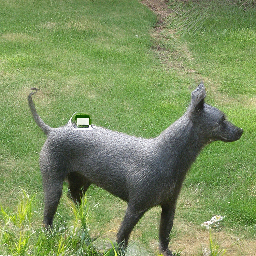} &
        \includegraphics[width=0.2\linewidth]{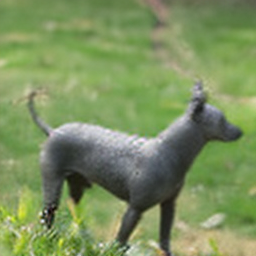} &
        \includegraphics[width=0.2\linewidth]{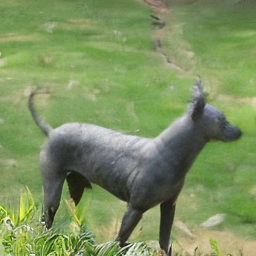} &
        \includegraphics[width=0.2\linewidth]{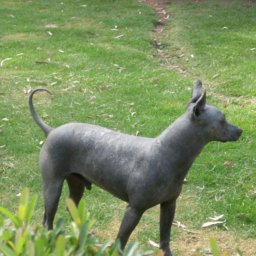} \\

        \includegraphics[width=0.2\linewidth]{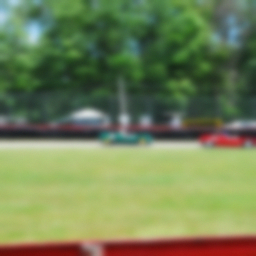} &
        \includegraphics[width=0.2\linewidth]{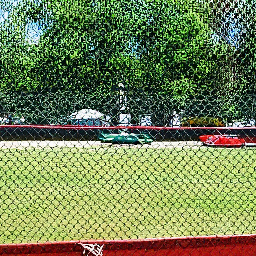} &
        \includegraphics[width=0.2\linewidth]{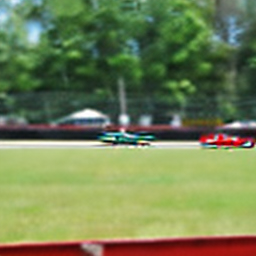} &
        \includegraphics[width=0.2\linewidth]{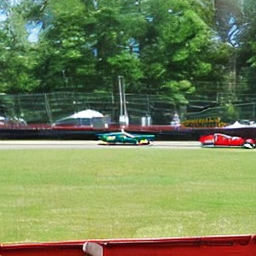} &
        \includegraphics[width=0.2\linewidth]{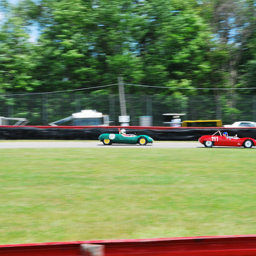} \\

    \end{tabular}

    \caption{
    Qualitative comparison for Gaussian Deblurring on the ImageNet dataset.
    Each row shows (from left to right): the measurement, the baseline method (DPS), the baseline latent solver
    (Latent DPS or ReSample), its REPA-enhanced variant, and the ground truth.
    The first three rows correspond to ReSample, while the last two correspond to Latent DPS.
    }
    \label{fig:deblur_grid}
\end{minipage}
}
\end{figure*}

\begin{figure*}[t]
\makebox[\textwidth][l]{%
\hspace*{-0.06\textwidth}%
\begin{minipage}{1.12\textwidth}
\centering
\setlength{\tabcolsep}{1pt}
\renewcommand{\arraystretch}{0.5}

    \begin{tabular}{ccccc}
        \textbf{Measurement}  & \textbf{DPS}  & \textbf{Without REPA} & \textbf{With REPA} & \textbf{Ground Truth} \\

        \includegraphics[width=0.20\linewidth]{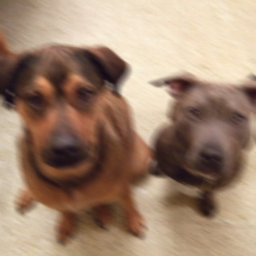} &
        \includegraphics[width=0.20\linewidth]{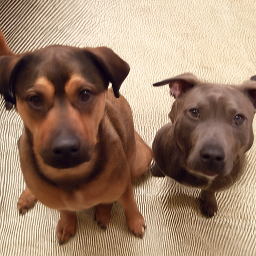} &
        \includegraphics[width=0.20\linewidth]{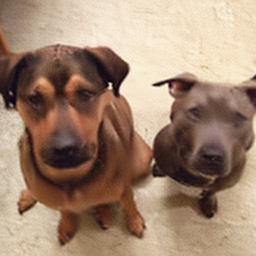} &
        \includegraphics[width=0.20\linewidth]{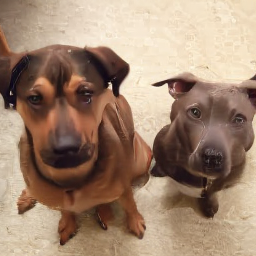} &
        \includegraphics[width=0.20\linewidth]{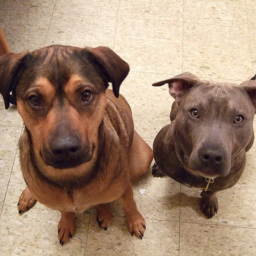} \\

        \includegraphics[width=0.20\linewidth]{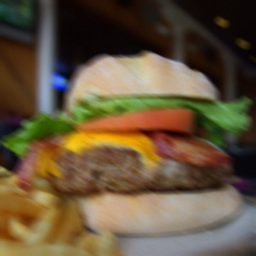} &
        \includegraphics[width=0.20\linewidth]{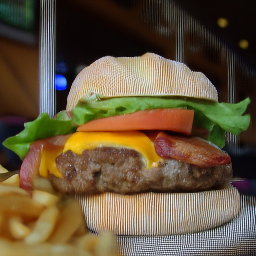} &
        \includegraphics[width=0.20\linewidth]{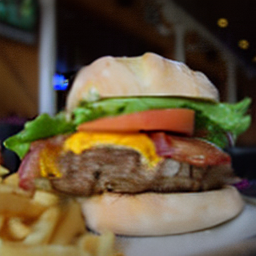} &
        \includegraphics[width=0.20\linewidth]{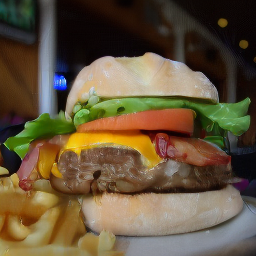} &
        \includegraphics[width=0.20\linewidth]{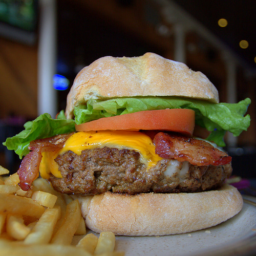} \\

        \includegraphics[width=0.20\linewidth]{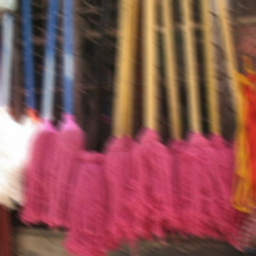} &
        \includegraphics[width=0.20\linewidth]{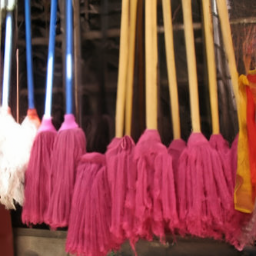} &
        \includegraphics[width=0.20\linewidth]{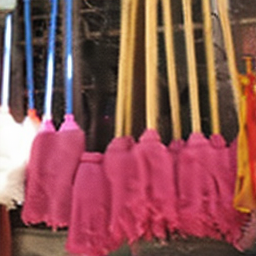} &
        \includegraphics[width=0.20\linewidth]{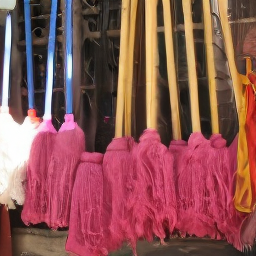} &
        \includegraphics[width=0.20\linewidth]{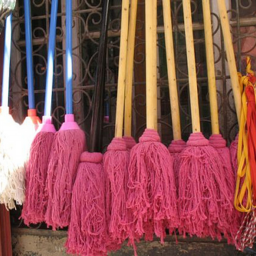} \\

        \includegraphics[width=0.20\linewidth]{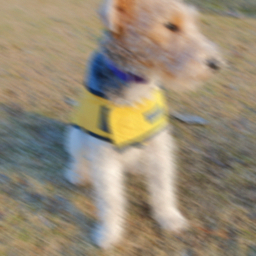} &
        \includegraphics[width=0.20\linewidth]{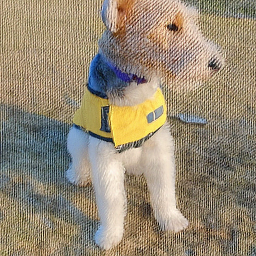} &
        \includegraphics[width=0.20\linewidth]{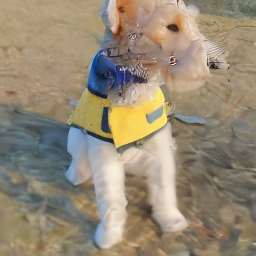} &
        \includegraphics[width=0.20\linewidth]{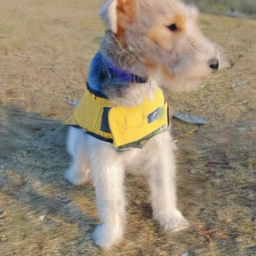} &
        \includegraphics[width=0.20\linewidth]{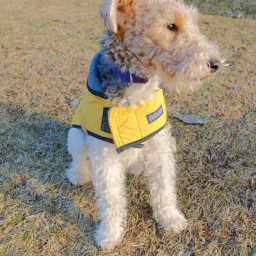} \\

        \includegraphics[width=0.20\linewidth]{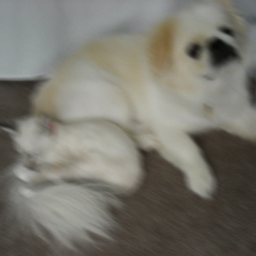} &
        \includegraphics[width=0.20\linewidth]{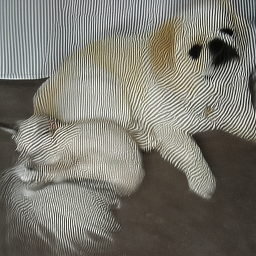} &
        \includegraphics[width=0.20\linewidth]{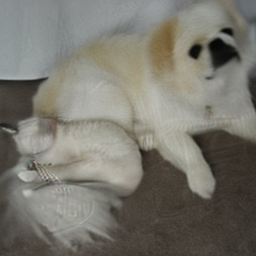} &
        \includegraphics[width=0.20\linewidth]{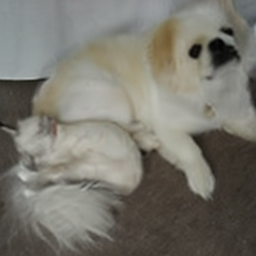} &
        \includegraphics[width=0.20\linewidth]{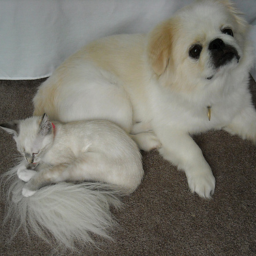} \\

    \end{tabular}

    \caption{
    Qualitative comparison for Motion Deblurring on the ImageNet dataset.
    Each row shows (from left to right): the measurement, the baseline method (DPS), the baseline latent solver
    (Latent DPS or ReSample), its REPA-enhanced variant, and the ground truth.
    The first three rows correspond to ReSample, while the last two correspond to Latent DPS.
    }
    \label{fig:deblur_grid}
\end{minipage}
}
\end{figure*}

\FloatBarrier
\onecolumn
\newpage

\end{document}